\pdfoutput=1

\documentclass[12pt,twoside]{article}

\usepackage{amsmath}
\usepackage{amsthm}
\usepackage{amssymb}
\usepackage{amsfonts}
\usepackage{algorithm}
\usepackage{algpseudocode}
\usepackage[top = 1 in, bottom = 1.2 in, left = 1 in, right = 1 in]{geometry}
\usepackage{color}
\usepackage{mathrsfs}
\usepackage[normalem]{ulem}
\usepackage{longtable}
\usepackage{hyperref}
\usepackage{cleveref}
\usepackage{natbib}
\usepackage{dsfont}
\usepackage{graphicx}
\usepackage{caption}
\usepackage{subcaption}
\usepackage{float}
\usepackage{mathtools}
\usepackage{etoolbox}
\usepackage{thmtools}
\usepackage{enumitem}
\usepackage[textsize=tiny]{todonotes}
\usepackage{multirow}

\usepackage{esvect}

\usepackage[utf8]{inputenc}
\usepackage{comment}

\newtheorem{claim}{Claim}
\newtheorem{theorem}{Theorem}
\newtheorem{lemma}{Lemma}

\newtheorem{assumption}{Assumption}
\declaretheoremstyle[headfont=\bf,bodyfont=\normalfont]{ex}
\declaretheorem[style=ex]{example}
\declaretheoremstyle[bodyfont=\normalfont]{rm}
\declaretheorem[numbered=no,style=rm]{remark}

\newcommand{\normmm}{{\vert\kern-0.25ex\vert\kern-0.25ex\vert}}
\newcommand{\bignormmm}{{\big\vert\kern-0.25ex\big\vert\kern-0.25ex\big\vert}}
\newcommand{\Bignormmm}{{\Big\vert\kern-0.25ex\Big\vert\kern-0.25ex\Big\vert}}

\newcommand{\defn}{\ensuremath{:\,=}}

\newcommand{\trace}{\ensuremath{{\rm Tr}}}

\usepackage{lipsum}

\newcommand\blfootnote[1]{%
	\begingroup
	\renewcommand\thefootnote{}\footnotetext{#1}%
	\endgroup
}

\makeatletter
\long\def\@makecaption#1#2{
	\vskip 0.8ex
	\setbox\@tempboxa\hbox{\small {\bf #1:} #2}
	\parindent 1.5em  
	\dimen0=\hsize
	\advance\dimen0 by -3em
	\ifdim \wd\@tempboxa >\dimen0
	\hbox to \hsize{
		\parindent 0em
		\hfil 
		\parbox{\dimen0}{\def\baselinestretch{0.96}\small
			{\bf #1.} {#2}
		} 
		\hfil}
	\else \hbox to \hsize{\hfil \box\@tempboxa \hfil}
	\fi
}
\makeatother

\usepackage{hyperref}

\newcommand{\yaqidone}{}

\usepackage{stackrel}

\newcommand{\Prob}{\ensuremath{\mathds{P}}}
\newcommand{\Exp}{\ensuremath{\mathds{E}}}

\newcommand{\Real}{\ensuremath{\mathds{R}}}
\newcommand{\Realpos}{\ensuremath{\Real_+}}

\newcommand{\Sym}{\ensuremath{\mathds{S}}}
\newcommand{\Sphere}{\ensuremath{\mathds{U}}}

\newcommand{\plaincon}{\ensuremath{c}}
\newcommand{\PlainCon}{\ensuremath{C}}
\newcommand{\const}[1]{\ensuremath{\plaincon_{#1}}}
\newcommand{\Const}[1]{\ensuremath{\PlainCon_{#1}}}

\newcommand{\rank}{\ensuremath{{\rm rank}}}
\newcommand{\clip}[1]{\ensuremath{{\rm clip}[\,#1\,]}}

\newcommand{\Dim}{\ensuremath{d}}

\newcommand{\one}{\ensuremath{\mathds{1}}}

\newcommand{\RKHS}{\ensuremath{\mathds{H}}}

\newcommand{\by}{\ensuremath{\boldsymbol{y}}}

\newcommand{\bX}{\ensuremath{\boldsymbol{X}}}

\newcommand{\IdMt}{\ensuremath{\boldsymbol{I}}}

\newcommand{\Data}{\ensuremath{\mathcal{H}}}

\newcommand{\nfed}{\ensuremath{= \, :}}

\newcommand{\numobs}{\ensuremath{n}}

\usepackage{upgreek} \newcommand{\distr}{\ensuremath{\upmu}}

\newcommand{\distrdata}{\ensuremath{\bar{\distr}}}

\newcommand{\Term}{\ensuremath{T}}
\newcommand{\Termone}{\ensuremath{\Term_1}}
\newcommand{\Termtwo}{\ensuremath{\Term_2}}

\newcommand{\bx}{\ensuremath{\boldsymbol{x}}}

\newcommand{\bB}{\ensuremath{\boldsymbol{B}}}

\DeclarePairedDelimiterX{\anglep}[1]{(}{)}{#1}
\makeatletter
\newcommand{\@spanstar}[1]{{\rm span}\anglep*{#1}}
\newcommand{\@spannostar}[2][]{{\rm span}\anglep[#1]{#2}}
\newcommand{\Span}{\@ifstar\@spanstar\@spannostar}
\makeatother

\DeclarePairedDelimiterX{\dfun}[2]{(}{)}{#1 \;\delimsize\|\; #2}
\makeatletter
\newcommand{\@trunstar}[2]{$\chi$^2\dfun*{#1}{#2}}
\newcommand{\@trunnostar}[3][]{$\chi$^2\dfun[#1]{#2}{#3}}
\newcommand{\chisq}{\@ifstar\@trunstar\@trunnostar}
\makeatother

\DeclarePairedDelimiterX{\inprod}[2]{\langle}{\rangle}{#1, \, #2}

\DeclarePairedDelimiterX{\kulldiv}[2]{(}{)}{#1\;\delimsize\|\;#2}
\makeatletter
\newcommand{\@kullstar}[2]{D_{\text{KL}}\kulldiv*{#1}{#2}}
\newcommand{\@kullnostar}[3][]{D_{\text{KL}}\kulldiv[#1]{#2}{#3}}
\newcommand{\kull}{\@ifstar\@kullstar\@kullnostar}
\makeatother

\makeatletter
\newcommand{\@hilinstar}[2]{\inprod*{#1}{#2}_{\RKHS}}
\newcommand{\@hilinnostar}[3][]{\inprod[#1]{#2}{#3}_{\RKHS}}
\newcommand{\hilin}{\@ifstar\@hilinstar\@hilinnostar}
\makeatother

\makeatletter
\newcommand{\@mudatainstar}[2]{\inprod*{#1}{#2}_{\distrdata}}
\newcommand{\@mudatainnostar}[3][]{\inprod[#1]{#2}{#3}_{\distrdata}}
\newcommand{\mudatain}{\@ifstar\@mudatainstar\@mudatainnostar}
\makeatother

\makeatletter
\newcommand{\@mudatahinstar}[3]{\inprod*{#2}{#3}_{\distrdata}}
\newcommand{\@mudatahinnostar}[4][]{\inprod[#1]{#3}{#4}_{\distrdata}}
\newcommand{\mudatahin}{\@ifstar\@mudatahinstar\@mudatahinnostar}
\makeatother

\DeclarePairedDelimiterX{\defabs}[1]{|}{|}{#1}
\makeatletter
\newcommand{\@absstar}[1]{\defabs*{#1}}
\newcommand{\@absnostar}[2][]{\defabs[#1]{#2}}
\newcommand{\abs}{\@ifstar\@absstar\@absnostar}
\makeatother

\DeclarePairedDelimiterX{\defceil}[1]{\lceil}{\rceil}{#1}
\makeatletter
\newcommand{\@ceilstar}[1]{\defceil*{#1}}
\newcommand{\@ceilnostar}[2][]{\defceil[#1]{#2}}
\newcommand{\ceil}{\@ifstar\@ceilstar\@ceilnostar}
\makeatother

\DeclarePairedDelimiterX{\norm}[1]{\|}{\|}{#1}
\makeatletter
\newcommand{\@normstar}[1]{\norm*{#1}_{\RKHS}}
\newcommand{\@normnostar}[2][]{\norm[#1]{#2}_{\RKHS}}
\newcommand{\hilnorm}{\@ifstar\@normstar\@normnostar}
\makeatother

\DeclareFontFamily{U}{matha}{\hyphenchar\font45}
\DeclareFontShape{U}{matha}{m}{n}{
	<-6> matha5 <6-7> matha6 <7-8> matha7
	<8-9> matha8 <9-10> matha9
	<10-12> matha10 <12-> matha12
}{}
\DeclareSymbolFont{matha}{U}{matha}{m}{n}

\DeclareFontFamily{U}{mathx}{\hyphenchar\font45}
\DeclareFontShape{U}{mathx}{m}{n}{
	<-6> mathx5 <6-7> mathx6 <7-8> mathx7
	<8-9> mathx8 <9-10> mathx9
	<10-12> mathx10 <12-> mathx12
}{}
\DeclareSymbolFont{mathx}{U}{mathx}{m}{n}

\DeclareMathDelimiter{\vvvert} {0}{matha}{"7E}{mathx}{"17}%

\DeclarePairedDelimiterX{\opnorm}[1]{\vvvert}{\vvvert}{#1}

\makeatletter
\newcommand{\@hilopnormstar}[1]{\opnorm*{#1}_{\RKHS}}
\newcommand{\@hilopnormnostar}[2][]{\opnorm[#1]{#2}_{\RKHS}}
\newcommand{\hilopnorm}{\@ifstar\@hilopnormstar\@hilopnormnostar}
\makeatother

\makeatletter
\newcommand{\@muopnormstar}[1]{\opnorm*{#1}_{\distr}}
\newcommand{\@muopnormnostar}[2][]{\opnorm[#1]{#2}_{\distr}}
\newcommand{\muopnorm}{\@ifstar\@muopnormstar\@muopnormnostar}
\makeatother

\makeatletter
\newcommand{\@mudataopnormstar}[1]{\opnorm*{#1}_{\distrdata}}
\newcommand{\@mudataopnormnostar}[2][]{\opnorm[#1]{#2}_{\distrdata}}
\newcommand{\mudataopnorm}{\@ifstar\@mudataopnormstar\@mudataopnormnostar}
\makeatother

\makeatletter
\newcommand{\@supnormstar}[1]{\norm*{#1}_{\infty}}
\newcommand{\@supnormnostar}[2][]{\norm[#1]{#2}_{\infty}}
\newcommand{\supnorm}{\@ifstar\@supnormstar\@supnormnostar}
\makeatother

\makeatletter
\newcommand{\@munormstar}[1]{\norm*{#1}_{\distr}}
\newcommand{\@munormnostar}[2][]{\norm[#1]{#2}_{\distr}}
\newcommand{\munorm}{\@ifstar\@munormstar\@munormnostar}
\makeatother

\makeatletter
\newcommand{\@mudatanormstar}[1]{\norm*{#1}_{\distrdata}}
\newcommand{\@mudatanormnostar}[2][]{\norm[#1]{#2}_{\distrdata}}
\newcommand{\mudatanorm}{\@ifstar\@mudatanormstar\@mudatanormnostar}
\makeatother

\makeatletter
\newcommand{\@xidatanormstar}[1]{\norm*{#1}_{\sdistrdata}}
\newcommand{\@xidatanormnostar}[2][]{\norm[#1]{#2}_{\sdistrdata}}
\newcommand{\xidatanorm}{\@ifstar\@xidatanormstar\@xidatanormnostar}
\makeatother

\makeatletter
\newcommand{\@distrnormstar}[2]{\norm*{#1}_{#2}}
\newcommand{\@distrnormnostar}[3][]{\norm[#1]{#2}_{#3}}
\newcommand{\distrnorm}{\@ifstar\@distrnormstar\@distrnormnostar}
\makeatother

\makeatletter
\newcommand{\@psinormstar}[2]{\norm*{#2}_{\psi_{#1}}}
\newcommand{\@psinormnostar}[3][]{\norm[#1]{#3}_{\psi_{#2}}}
\newcommand{\psinorm}{\@ifstar\@psinormstar\@psinormnostar}
\makeatother

\newcommand{\featureh}[1]{\ensuremath{\phi}}

\newcommand{\Lip}{\ensuremath{L}}

\newcommand{\policy}{\ensuremath{{\boldsymbol{\pi}}}}

\newcommand{\AMt}{\ensuremath{\boldsymbol{A}}}

\newcommand{\AMth}[1]{\ensuremath{A_{#1}}}

\newcommand{\stderr}{\ensuremath{\sigma}}

\newcommand{\indicator}[1]{\ensuremath{\one_{#1}}}

\newcommand{\Lipf}[1]{\ensuremath{\Lip_{f}}}

\newcommand{\bigO}{\ensuremath{{O}}}
\newcommand{\bigOtil}{\ensuremath{\widetilde{\bigO}}}
\newcommand{\bigTheta}{\ensuremath{{\Theta}}}
\newcommand{\bigThetatil}{\ensuremath{{\widetilde{\Theta}}}}

\newenvironment{carlist}
{\begin{list}{$\bullet$}
		{\setlength{\topsep}{0.1in} \setlength{\partopsep}{0in}
			\setlength{\parsep}{0.1in} \setlength{\itemsep}{\parskip}
			\setlength{\leftmargin}{0.15in} \setlength{\rightmargin}{0.08in}
			\setlength{\listparindent}{0in} \setlength{\labelwidth}{0.08in}
			\setlength{\labelsep}{0.1in} \setlength{\itemindent}{0in}}}
	{\end{list}}

\newcommand{\bcar}{\begin{carlist}}
	\newcommand{\ecar}{\end{carlist}}

\newcommand{\Loss}{\ensuremath{\mathcal{L}}}

\newcommand{\Omegatil}{\ensuremath{\widetilde{\Omega}}}

\newcommand{\demand}{\ensuremath{d}}
\newcommand{\Demand}{\ensuremath{D}}

\newcommand{\feat}{\ensuremath{\boldsymbol{x}}}
\newcommand{\feati}[1]{\ensuremath{x_{#1}}}

\newcommand{\featt}[1]{\ensuremath{\feat_{#1}}}
\newcommand{\FeatSp}{\ensuremath{\mathcal{X}}}
\newcommand{\price}{\ensuremath{p}}
\newcommand{\pricet}[1]{\ensuremath{\price_{#1}}}
\newcommand{\pricestar}{\ensuremath{\price^{\star}}}

\newcommand{\noise}{\ensuremath{\varepsilon}}

\newcommand{\return}{\ensuremath{r}}

\newcommand{\returnstar}{\ensuremath{\return^{\star}}}

\newcommand{\pricelb}{\ensuremath{\ell}}
\newcommand{\priceub}{\ensuremath{u}}

\newcommand{\para}{\ensuremath{\btheta}}
\newcommand{\paraint}{\ensuremath{\balpha}}
\newcommand{\paraslp}{\ensuremath{\bbeta}}
\newcommand{\parastar}{\ensuremath{\para^{\star}}}
\newcommand{\paraintstar}{\ensuremath{\paraint^{\star}}}
\newcommand{\paraslpstar}{\ensuremath{\paraslp^{\star}}}

\newcommand{\paraone}{\ensuremath{\widetilde{\para}}}
\newcommand{\paraintone}{\ensuremath{\widetilde{\paraint}}}
\newcommand{\paraslpone}{\ensuremath{\widetilde{\paraslp}}}
\newcommand{\priceone}{\ensuremath{\widetilde{\price}}}

\newcommand{\paratwo}{\ensuremath{\widehat{\para}}}
\newcommand{\parainttwo}{\ensuremath{\widehat{\paraint}}}
\newcommand{\paraslptwo}{\ensuremath{\widehat{\paraslp}}}
\newcommand{\pricetwo}{\ensuremath{\widehat{\price}}}

\newcommand{\Reg}{\ensuremath{\mathcal{R}}}

\newcommand{\perturb}{\ensuremath{\eta}}
\newcommand{\Rad}{\ensuremath{\xi}}

\newcommand{\bfm}[1]{\ensuremath{\mathbf{#1}}}

   \def\bB{\bfm B}

     \def\EE{\mathbb{E}}

   \def\bI{\bfm I}

     \def\RR{\mathbb{R}}

\def\bx{\bfm x}   \def\bX{\bfm X}  
\def\by{\bfm y}

\def\calI{{\cal  I}}

\def\calN{{\cal  N}} 
\def\calO{{\cal  O}}

\def\calU{{\cal  U}}

\newcommand{\bfsym}[1]{\ensuremath{\boldsymbol{#1}}}

 \def\balpha{\bfsym \alpha}
 \def\bbeta{\bfsym \beta}

 \def\btheta{\bfsym {\theta}}           
           \def\bepsilon{\bfsym \varepsilon}
              \def\bSigma{\bfsym \Sigma}

\newcommand{\GoodEvent}{\ensuremath{\mathcal{E}}}
\newcommand{\GoodEventcomp}{\ensuremath{\GoodEvent^c}}

\newcommand{\StatDim}{\ensuremath{K}}

\newcommand{\featparalb}{\ensuremath{b_1}}
\newcommand{\featparaub}{\ensuremath{b_2}}

\newcommand{\Covpara}{\ensuremath{\boldsymbol{S}}}
\newcommand{\Covparaone}{\ensuremath{\widetilde{\Covpara}\!\,}}
\newcommand{\Covparaonerealpop}{\ensuremath{\Covpara_0}}

\newcommand{\Covparatwo}{\ensuremath{\widehat{\Covpara}\!\,}}
\newcommand{\Covparatworealpop}{\ensuremath{{\Covpara^{\star}}}}
\newcommand{\Covparatwopop}{\ensuremath{{\Covparatwo^{\dagger}}}}
\newcommand{\Covparatwopopstar}{\ensuremath{{\Covparatwo^{\star}}}}

\newcommand{\Covfeat}{\ensuremath{\boldsymbol{\Sigma}}}

\newcommand{\Covfeattwo}{\ensuremath{\widehat{\Covfeat}\!\,}}

\newcommand{\bSigmahat}{\ensuremath{\widehat{\bSigma}\!\,}}
\newcommand{\bbetahat}{\ensuremath{\widehat{\bbeta}}}
\newcommand{\bnoise}{\ensuremath{\boldsymbol{\noise}}}

\newcommand{\vecu}{\ensuremath{\boldsymbol{u}}}
\newcommand{\vecv}{\ensuremath{\boldsymbol{v}}}
\newcommand{\vece}{\ensuremath{\boldsymbol{e}}}

\newcommand{\constcon}{\ensuremath{\const{{\sf ac}}}}

\newcommand{\eig}{\ensuremath{\lambda}}
\newcommand{\eigpop}{\ensuremath{\eig}}

\newcommand{\eigpopstar}{\ensuremath{\widehat{\eig}^{\star}}}
\newcommand{\eigtwo}{\ensuremath{\widehat{\eig}}}
\newcommand{\eigmin}{\ensuremath{\eig_{\min}}}
\newcommand{\eigmax}{\ensuremath{\eig_{\max}}}
\newcommand{\eigSp}{\ensuremath{\mathds{U}}}

\newcommand{\stdfeat}{\ensuremath{\stderr_{\feat}}}

\newcommand{\constCovfeat}{\ensuremath{\const{\Covfeat}}}
\newcommand{\ConstCovfeat}{\ensuremath{\Const{\Covfeat}}}

\newcommand{\constnew}{\ensuremath{\frac{\constCovfeat}{2}}}
\newcommand{\Constnew}{\ensuremath{2 \, \ConstCovfeat}}

\newcommand{\EffDim}{\ensuremath{\widetilde{\Dim}}}

\newcommand{\ConstCI}{\ensuremath{\kappa}}

\newcommand{\perturbstar}{\ensuremath{\perturb^{\star}}}

\newcommand{\SNR}{\ensuremath{\mbox{SNR}}}
\newcommand{\Singular}{\ensuremath{\mathcal{S}}}

\numberwithin{equation}{section}
\begin{document}
	\begin{center}
		{\bf \Large Localized Exploration in Contextual Dynamic Pricing Achieves Dimension-Free Regret}				\blfootnote{Author names are sorted alphabetically.  
		Duan's research is supported by an NSF grant \mbox{DMS-2413812}.
		Fan's research is supported by NSF grants  DMS-2210833 and DMS-2053832 and ONR grant \mbox{N00014-22-1-2340}.
		Wang's research is supported by an NSF grant DMS-2210907 and a startup grant at Columbia University.
		} \\
		
		\vspace{2em}
		{\large{
				\begin{tabular}{cccccccc}
					Jinhang Chai$^\dagger$ && Yaqi Duan$^\diamond$ &&
					Jianqing Fan$^\dagger$ && Kaizheng Wang$^{*}$
				\end{tabular}
		}}
		\vspace{1em}
		
		\medskip
		
		\begin{tabular}{c}
			Department of Operations Research and Financial Engineering \\
			Princeton University$^\dagger$
		\end{tabular}
		
		\medskip 
		\begin{tabular}{c}
			Department of Technology, Operations and Statistics \\
			Stern School of Business,  New York University$^{\diamond}$
		\end{tabular}
		
		\medskip 
		\begin{tabular}{c}
			Department of Industrial Engineering and Operations Research \\
			Columbia University$^{*}$
		\end{tabular}
		
		\vspace{1.6em}
		\today
	\end{center}

	\begin{center}
		{\bf Abstract} \\ \vspace{.6em}
		\begin{minipage}{0.9\linewidth}
	{\small ~~~~
   We study the problem of contextual dynamic pricing with a linear demand model. We propose a novel localized exploration-then-commit (LetC) algorithm which starts with a pure exploration stage, followed by a refinement stage that explores near the learned optimal pricing policy, and finally enters a pure exploitation stage.   
   The algorithm is shown to achieve a minimax optimal, dimension-free 
   regret bound when the time horizon exceeds a polynomial of the covariate dimension. 
   Furthermore, we provide a general theoretical framework that encompasses the entire time spectrum, demonstrating how to balance exploration and exploitation when the horizon is limited. 
   The analysis is powered by a novel critical inequality that depicts the exploration-exploitation trade-off in dynamic pricing, mirroring its existing counterpart for the bias-variance trade-off in regularized regression. 
   Our theoretical results are validated by extensive experiments on synthetic and real-world data.   
    }
		\end{minipage}
	\end{center}

	
	

\section{Introduction}

In recent years, the rise of online sales platforms has transformed how sellers acquire and utilize customer data, such as demographics, purchasing history, and social media interactions. At the same time, market insights---such as industry trends, competitor behavior, and economic cycles---provide a broader understanding of the forces shaping the marketplace. By integrating these data streams and employing personalized pricing strategies, retailers can significantly enhance their profitability. This evolution has fueled a growing interest in contextual dynamic pricing, which sets optimal prices in real time based on sales data and various covariates.

In this work, we study contextual dynamic pricing under a linear demand model. To tackle the celebrated exploration-exploitation dilemma in decision-making, we propose a \emph{localized Explore-then-Commit} (LetC) algorithm with three stages. The first one is a burn-in stage, where we perform pure exploration by alternating between two prices to obtain a pilot estimate of the demand model. This stage gains substantially statistical information but at the expenses of big revenue losses.  The second stage conducts localized exploration, where we  explore only near the learned optimal pricing policy. This new stage gains the statistical information at small losses of revenue.  With a proper choice of localization, the benefit of statistical precision is shown to exceed the cost of revenue.  Eventually, the parameters are learnt so precisely that we can fully deploy the optimal stragegy.  The third and final stage simply deploys the optimal policy estimated from the collected data.
Compared to popular Explore-then-Commit (ETC) pricing strategies \citep{javanmard2019dynamic,ban2021personalized,fan2022policy}, our additional localized exploration stage refines the initial policy at a low cost and helps achieve a sharp regret bound.

\paragraph{Our contributions}
Our main contributions are summarized as follows.

\begin{enumerate}
\item We develop a novel three-stage algorithm for contextual dynamic pricing. Under a linear demand model, we offer clear guidelines for determining the lengths of each stage and the size of price perturbations for localized exploration.

\item We prove a dimension-free regret bound for the algorithm with long horizons. Surprisingly, when the time horizon $\Term$ and the covariate dimension $\Dim$ satisfy $\Term > \Dim^4$, the regret is at most of order $\sqrt{\Term}$. We also derive a matching minimax lower bound. The above statements only hide polylogarithmic factors in $\Term$.

\item We extend our theoretical framework to cover the entire spectrum of time, enabling a seamless transition from short to long horizons. To capture the exploration-exploitation trade-off in dynamic pricing, we introduce a critical inequality analogous to the existing one for bias-variance trade-off in penalized regression. The inequality identifies the optimal size of localized exploration.
\end{enumerate}

\paragraph{Related works} Here, we give a selective review of related work in the literature. 
Dynamic pricing without covariates has been extensively studied over the past two decades \citep{broder2012dynamic,besbes2015surprising,Den15,wang2021multimodal}. Recently, there has a growing attention on how to incorporate contextual information in pricing. \cite{nambiar2019dynamic,chen2021nonparametric,bu2022context} studied demand models that involve general nonparametric functions of the covariate $\feat \in \Real^d$. The generality comes with the curse of dimension in regret. On the other hand, parametric and semi-parametric demand models can deal with high-dimensional covariates. A number of works in this direction assumed that the demand depends on the price $\price$ and the covariate $\feat$ through a linear link function $\balpha^{\top} \feat + \gamma \price$, where $\alpha \in \Real^d$ and $\gamma \in \Real$ are unknown coefficients. For instance, \cite{qiang2016dynamic} studied a linear demand model in a well-separated regime; \cite{COP22} considered a logit choice model in the offline setting; \cite{javanmard2019dynamic} worked on parametric binary choice models; \cite{xu2022towards,luo2024distribution,fan2022policy} investigated semi-parametric demand models.

Since the coefficient $\gamma$ of the price $\price$ does not depend on the covariate $\feat$, the above models do not capture the heterogeneous price sensitivity in the population. To handle such challenge, \cite{ban2021personalized} considered parametric demand models of the form $g \{ \balpha^{\top} \feat + (\bbeta^{\top} \feat) \price \}$, where $g:~\Real \to \Real$ is a known function and $\bbeta \in \Real^d$ is a coefficient vector governing the influence of covariates on the price sensitivity. They considered the high-dimensional setting where $d$ is large, but $\balpha$ and $\bbeta$ only have a small number $s$ of non-zero entries. They proposed a two-stage Explore-then-Commit algorithm that first alternates between two candidate prices to collect data and then deploys the optimal pricing policy based on the learned demand model. We consider the scenario where $g$ is the identity function, but the coefficients can be non-sparse. The result in \cite{ban2021personalized} implies a regret bound of order $\Dim \sqrt{\Term}$, up to logarithmic factors. Our theory shows that under mild assumptions, one can achieve a dimension-free $\sqrt{\Term}$ regret for large $\Term$. Compared to the Explore-then-Commit algorithm, the significant improvement of our regret is due to an additional localized exploration stage that conducts low-cost experiments to refine the initial estimates.

Localized random exploration is a popular method in control \citep{moore1990efficient,amin2021survey}, where an agent chooses from a perturbed set of best known actions. Our localized exploration stage uses similar ideas. 
In the dynamic pricing literature, price perturbation has been adopted for different purposes, such as addressing endogeneity \citep{nambiar2019dynamic,bu2022context} and handling reference effects \citep{den2022dynamic}. Those works only considered pricing without contextual information.


	
\paragraph{Paper organization}
 The rest of the paper is organized as follows. Section \ref{sec:setup} provides the background and problem setup. Section \ref{sec:alg} outlines our algorithmic framework. Section \ref{sec:main} presents the main theoretical results. Section \ref{sec:exp} demonstrates the practical performance of our approach through extensive numerical simulations and real-data experiments. Finally, Section \ref{sec:con} concludes the paper. For conciseness, all proofs are deferred to the appendix.
	
	\paragraph{Notation}
	The constants $c,c_0,C,\ldots$ may vary from line to line.
    We use boldface Greek letters to denote vectors, boldface capital letters to denote matrices, and plain letters to denote scalars. For nonnegative sequences $a_n$ and $b_n$, we write $a_n=\bigO(b_n)$ or $a_n\lesssim b_n$ if there exists some universal constant $C$ such that $a_n\le Cb_n$ for sufficiently large $n$, and write 
    $a_n=\Omega(b_n)$ or $a_n\gtrsim$ if there exists some universal constant $C$ such that $a_n\ge Cb_n$ for sufficiently large $n$. Furthermore, we write $a_n=\bigTheta(b_n)$ or $a_n\asymp b_n$ if both $a_n\lesssim b_n$ and $a_n\gtrsim b_n$ hold. Similarly, we use $\bigOtil$, $\Omegatil$, and $\bigThetatil$ for similar meanings with exceptions up to some logarithmic factors.
    For random event $\GoodEvent$, $\indicator\GoodEvent$ stands for its indicator function. We use $\Sym^{\Dim}$ to denote the set of $d\times d$ symmetric matrices, and $\mathds{U}^{d-1}$ to denote the unit sphere with ambient dimension $d$. For a random variable $X$ , the Orlicz norm $\|X\|_{\psi_q}$ is defined as $\inf\{\lambda>0\ |\ \EE[\psi_q(|X|/\lambda)]\le 1\}$ where $\psi_q(t):=\exp(t^q)-1$; and for a random vector $\feat\in \Real^d$, we denote $\|\feat\|_{\psi_q}=\sup_{\boldsymbol{u} \in \mathds{U}^{d-1}}\|\feat^{\top} \boldsymbol{u}\|_{\psi_q}$.
    In most cases, we omit the subscript when the dimension is clear from the context, e.g., 
    the identity matrix is denoted as $\IdMt$.

	
	\section{Problem set-up}
	\label{sec:setup}
	We consider a dynamic pricing problem where a seller aims to maximize its revenue by adjusting the price of a single product over time. The demand for the product is modeled by a linear function that depends on contextual information. Below, we outline the key components of the problem.
	
	
	\paragraph{Contextual linear demand model:}
	
	Suppose the demand for a product is influenced by contextual factors, such as customer demographics, purchase history, or market conditions. 
	At each time period $t$, the seller observes a $\Dim$-dimensional \emph{context vector} $\feat_t \in \FeatSp \subseteq \Real^{\Dim}$, which encodes the relevant information about the customer or market environment. 
	
	Given the context $\feat_t$, the seller selects a price $\price_t$ from the feasible interval $[\pricelb, \priceub] \subseteq \Realpos$. The customer or market responds to the selected price $\price_t$, conditioned on the current context~$\feat_t$, by generating a random demand $\Demand_t \in \Real$, modeled as:
	\begin{align}
		\label{eq:demand}
		\Demand_t \; = \; \feat_t^{\top} \paraintstar + \price_t \, (\feat_t^{\top} \paraslpstar) + \noise_t \, .
	\end{align}
	Here $\noise_t$ represents random demand shocks, which are assumed to be sub-Gaussian with zero mean and a variance proxy of $\stderr^2 > 0$;
	vectors $\paraintstar, \paraslpstar \in \Real^{\Dim}$ are unknown model parameters.
	For notational convenience, we define $\parastar \! \defn (\paraintstar{}^\top, \paraslpstar{}^\top)^\top \in \Real^{2\Dim}$ as the concatenation of vectors $\paraintstar$ and $\paraslpstar$.
	This contextual linear demand model~\eqref{eq:demand} was introduced by~\citet{ban2021personalized}.
	
	The demand model~\eqref{eq:demand} is called linear for two reasons:
	(i) The demand function \mbox{$\demand^{\star}(\feat_t, \price_t) $} $\defn \Exp [ \Demand_t \mid \feat_t, \price_t]$ is linear in the price $\price_t$, with the intercept $\feat_t^{\top} \paraintstar \geq 0$ and the slope $\feat_t^{\top} \paraslpstar \leq 0$.
	(ii) Both the intercept and slope are linear in the context vector $\feat_t \in \Real^{\Dim}$.
	
	We adopt this model for its analytical simplicity, though it can be extended to account for nonlinear and higher-order effects of context and price on demand.  As in the traditional linear regression models, the linear models in the intercepts and slopes here accommodate the transformed features, their interactions, and tensors.  Despite its straightforward structure, the model unveils several interesting phenomena.
	

	\paragraph{Optimal pricing policy:}
	
	Recall the seller’s goal is to maximize the revenue $\return_t$ at each stage by selecting an appropriate price $\price_t$. The (expected) revenue $\return_t$ is defined as
	\begin{align}
		\label{eq:revenue}
		\return_t \; \equiv \; \return(\feat_t, \price_t; \parastar) \; \defn \; \price_t \, \Exp [ \Demand_t \mid \feat_t, \price_t] \; = \; \price_t \, \big\{ \feat_t^{\top} \paraintstar + \price_t \, (\feat_t^{\top} \paraslpstar) \big\} ,
	\end{align}
	which is quadratic with respect to the price $\price_t$.
	To maximize revenue, the optimal price $\price_t$ is set as
	\footnote{We omit the effect of truncation to the interval $[\pricelb, \priceub]$ for now.}
	\begin{align}
		\label{eq:opt_price}
		\price_t = \pricestar(\feat_t) \equiv \price(\feat_t; \parastar) \defn - \frac{\feat_t^{\top} \paraintstar}{2 \, \feat_t^{\top} \paraslpstar} \, .
	\end{align}
	At this price, the resulting revenue is
	\begin{align}
    \label{eq:max-revenue}
		\return_t \; = \; \returnstar(\feat_t) \equiv \returnstar(\feat_t; \parastar) \defn - \frac{(\feat_t^{\top} \paraintstar)^2}{4 \, \feat_t^{\top} \paraslpstar} \, .
	\end{align}
	
	Therefore, if the true parameters $\parastar = (\paraintstar, \paraslpstar)$ were known, the optimal price $\price_t$ could be determined analytically from the observed context $\feat_t$ using equation~\eqref{eq:opt_price}. However, since the model parameters $\parastar$ are not directly observable, we must estimate them from the available data.

	\paragraph{Online learning process:}
	
	The objective of this paper is to derive a statistically efficient method for learning the optimal pricing strategy in an online setting.
	
	In the online learning process, at each time $t$, the seller observes a context $\feat_t$, which is independently drawn from a population distribution $\distr$ over the feature space $\FeatSp \subseteq \Real^{\Dim}$. Based on the observed history,
	\begin{align*}
		\Data_{1 : t-1} = (\feat_1, \price_1, \Demand_1, \feat_2, \price_2, \Demand_2, \ldots, \feat_{t-1}, \price_{t-1}, \Demand_{t-1}) \, .
	\end{align*}
	the seller then selects a price $\price_t$.
	Next, the seller observes the demand $\Demand_t$ based on equation~\eqref{eq:demand}, where the noise terms $\{\noise_t\}_{t \geq 1}$ are independent across time steps. 
	The realized revenue at time $t$, denoted by $\return_t$, is given by equation~\eqref{eq:revenue}.
	
	\begin{subequations}
	\label{eq:regret}
	For any online learning algorithm that helps the seller choose the price $\price_t$, we assess its effectiveness by considering the cumulative regret at a (pre-specified) terminal stage $\Term$: \footnote{It can be easily extended, through a doubling trick, to anytime-valid results that do not require pre-specifying the terminal stage.}
	\begin{align}
		\label{eq:regret_def}
		\Reg(\Term) \; \equiv \; \Reg(\Term;\parastar)\; \defn \; \sum_{t=1}^{\Term} \; \{ \returnstar(\feat_t) - \return_t \} \, .
	\end{align}
	At each stage, the regret measures the difference between the expected revenue achieved by the algorithm and the optimal revenue that could be achieved with the best price.
	After some algebra, we can express the regret $\Reg(\Term)$ as:
	\begin{align}
		\label{eq:regret_ub}
		\Reg(\Term) \; \leq \; \sum_{t=1}^{\Term} \; (-\feat_t^{\top} \paraslpstar) \big\{ \price_t - \pricestar(\feat_t) \big\}^2
		\; = \; \sum_{t=1}^{\Term} \; (-\feat_t^{\top} \paraslpstar) \Big( \price_t + \frac{\feat_t^{\top} \paraintstar}{2 \, \feat_t^{\top} \paraslpstar} \Big)^2 \, .
	\end{align}
	This inequality shows how the closeness between the implemented price $\price_t$ and the optimal price $\pricestar(\feat_t)$ governs the overall regret in terms of revenue.
	\end{subequations}
	
	In the following analysis, we focus on an efficient method to minimize the expected regret $\Exp[\Reg(\Term)]$, where the expectation is taken over both the randomness in the data generation process and the randomness of the learning algorithm.


	
	\section{A localized explore-then-commit algorithm \yaqidone}
	\label{sec:alg}
	
	In this section, we introduce a three-stage LetC pricing procedure featuring a novel localized exploration strategy. The complete algorithm is provided in \Cref{alg:localized-explore-then-commit}. Below, we elaborate on each stage in detail. We will use a clipping operator $\clip{\price}:=\max(\min(\price,\priceub),\pricelb)$ to project any proposed price to its feasible value in the interval $[\pricelb, \priceub]$.	
	\paragraph{Stage 1: Burn-in exploration.}
	
	In this stage, the main goal is to collect observations about the system without concern for revenue loss due to suboptimal pricing. To ensure effective exploration of the feature space, we alternate between the two extreme prices within the feasible range. 
	Specifically, 
	the prices $\price_t$ are independently drawn from a binary distribution, taking $\pricelb$ or $\priceub$, each with probability $\frac{1}{2}$.
	The binary design is optimal for learning parameters $\btheta^\star$ in the linear demand model.
	This pure exploration strategy has also been used in previous algorithms in the literature \citep{broder2012dynamic,ban2021personalized,bastani2022meta}.
	
	\begin{subequations}
		This initial phase proceeds for $\Termone$ steps, where $\Termone$ is determined by the choice of the total termination time $\Term$ and other problem-specific structures, with the optimal value to be detailed later in \Cref{sec:ub}.
		After collecting data from these $\Termone$ steps, we form an estimate of the parameter $\parastar$ using the historical data $\Data_{1 : \Termone} = \{ (\feat_t, \price_t, \Demand_t) \}_{t = 1}^{\Termone}$.
		Specifically, we employ ordinary linear squares (OLS) regression:
		Define an empirical quadratic loss function as \vspace{-.7em}
		\begin{align}
			\label{eq:Loss}
			\Loss(\para; \Data_{1:\Termone}) 
			\; \defn \; \frac{1}{\Termone} \sum_{t=1}^{\Termone}
			\big\{ \feat_t^{\top} \paraint + \price_t \, (\feat_t^{\top} \paraslp) \; - \; \Demand_t \big\}^2 \, .
		\end{align}~\vspace{-1.2em} \\
		The estimate $\paraone$ is obtained by minimizing the loss, i.e.
		\vspace{-.5em}
		\begin{align}
			\label{eq:paraone}
			\paraone \; \defn \; \arg \min_{\para \in \Real^{2\Dim}} \Loss(\para; \Data_{1:\Termone}) \, .
		\end{align}~\vspace{-1.4em} \\
		The optimal pricing strategy(before truncation), based on the estimate $\paraone$, is then given by
		\vspace{-.5em}
		\begin{align}
			\label{eq:priceone}
			\priceone(\feat) \defn \price(\feat; \paraone) = - \frac{\feat^{\top} \paraintone}{2 \, \feat^{\top} \paraslpone} \, ,
		\end{align}~\vspace{-1.2em} \\
		which aligns with the definition of the optimal price in equation~\eqref{eq:opt_price}. Of course, we will round $\priceone(\feat)$ to a feasible value.
	\end{subequations}
	
	We comment that a pure exploration phase is quite practical in real-world scenarios. When introducing a new product to the market, it is often beneficial to experiment with extreme pricing before settling on an optimal point. For instance, offering substantial discounts or coupons to encourage initial purchases simulates the effect of low prices. On the other hand, launching a premium or exclusive version of the product at a higher price helps gauge customer sensitivity and willingness to pay, providing insight into price elasticity.
	\vspace{-.5em}

	\paragraph{Stage 2: Localized exploration.}
	
	The localized exploration in Stage 2 is a novel contribution of our work, distinguishing our method from those proposed in \cite{ban2021personalized, fan2022policy, chen2021fairness, zhao2023high}. 
	Recall that following the first stage, we have obtained an initial pricing policy $\priceone$, though the estimate of the parameters in the demand function may still lack sufficient precision due to the cost of exploration. In the second phase, we aim to refine it by using $\priceone$ as a baseline while sampling around it to further fine-tune the parameters.  This stage aims at gaining statistical information at low cost of revenue loss.  With a proper choice of localization parameter, the statistical information gain will exceed the revenue lost.

	Specifically, at each time for $t = \Termone + 1, \Termone + 2, \ldots, \Termtwo$, we flip a coin. If heads, we set the price as $\price_t \defn \clip{\priceone(\feat_t) + \perturb}$; if tails, we set $\price_t \defn \clip{\priceone(\feat_t) - \perturb}$. Here, $\Termone + \Termtwo$ and $\perturb$ are parameters that will be determined later in \Cref{sec:ub}. We express the sampling scheme as 
	\begin{align}
		\label{eq:pricetwo}
		\price_t \defn \clip{\priceone(\feat_t)+ \perturb \cdot \Rad_t} \, ,
	\end{align}~\vspace{-1em} \\
	where $\{ \Rad_t \}_{t=\Termone + 1}^{\Termone + \Termtwo}$ are independent Rademacher random variables.

	
		At the end of this stage, we update the parameter estimate using the newly collected data from the second phase, incorporating it into the loss function $\Loss$ as presented in equation~\eqref{eq:Loss}. Specifically, we define the updated parameter estimate as
		\begin{align}
			\paratwo 
			\; \defn \; & \arg \min_{\para \in \Real^{2\Dim}} \; \Loss(\para; \Data_{\Termone + 1 : \Termone + \Termtwo})  
			\; = \; \arg \min_{\para \in \Real^{2\Dim}} \; \frac{1}{\Termtwo} \sum_{t = \Termone+1}^{\Termone + \Termtwo} 
			\big\{ \feat_t^{\top} \paraint + \price_t \, (\feat_t^{\top} \paraslp) \; - \; \Demand_t \big\}^2 \, ,
			\label{eq:paratwo}
		\end{align}
		where data from $\Termone+1$ to $\Termone + \Termtwo$ in Stage 2 is used.
		The refined pricing policy(before truncation) $ \pricetwo $ is then defined as $ \pricetwo(\feat) \defn \price(\feat; \paratwo) = - (\feat^{\top} \parainttwo) / (2 \, \feat^{\top} \paraslptwo)$, consistent with equation~\eqref{eq:priceone}.
		It is worth noting that, in practice, data from the first stage can also be included, allowing the use of all data up to $\Termone + \Termtwo$. However, for simplicity in theoretical analysis, we restrict the update to Stage 2 data, as this is sufficient to achieve the optimal rate of regret.
	

	This localized exploration strategy balances exploration and exploitation effectively. By using the initial estimate $\priceone$, we can secure a near optimal level of revenue, which corresponds to exploitation. At the same time, the exploration component (adding random noises to pricing) continues to play a role by allowing us to capture more nuanced patterns in market behavior (with relatively low regret), leading to a better pricing strategy.  For a linear demand model, it is well known from the design of experiment that setting the exploring prices driven by Rademacher random perturbations maximizes the information gain in learning $\btheta^\star$ for any distributions with the same support.  The revenue lost due to the use of non-optimal price is of order $\eta^2$.  Therefore, it will be shown in our proof that with a proper choice of $\eta$, the gain of statistical information in learning the parameter $\btheta^\star$ outbenefits the revenue loss due to the local exploration and this benefit diminishes when $T_2$ is sufficiently large that parameters are sufficiently accurately estimated.  Hence, we can now apply the learned optimal pricing strategy as if the true parameters were known.
	
	\vspace{-.5em}

	\paragraph{Stage 3: Committing.}
	At the final stage of the algorithm, we implement the refined policy $\pricetwo(\feat)$ for the remaining steps $t = \Termone + \Termtwo + 1, \Termone + \Termtwo + 2, \ldots, \Term$. Since the estimates have reached sufficient precision, further exploration is unnecessary, and pure exploitation of the learned knowledge ensures minimal regret.  \\

	\newcommand{\HRule}{\vspace{-2em} \\ \rule{\linewidth}{0.1mm}}
	\begin{algorithm}[ht!]
		\caption{Localized Explore-then-Commit (LetC) Algorithm in Dynamic Pricing}
		\label{alg:localized-explore-then-commit}
		
		\textbf{Input:} Numbers of steps in Stages 1 and 2, $\Termone$ and $\Termtwo$; \\ 
		\phantom{\textbf{Input:}} magnitude of local perturbation, $\perturb$; feasible price range $[\pricelb,\priceub]$.  \\
		\textbf{Output:}  Prices at each step, $\price_1, \price_2, \ldots, \price_{\Term}$. \\
		\HRule
		\begin{algorithmic}[0]
			\State \emph{Burn-in Exploration Stage:}
			\For{$t=1,\cdots,\Termone$}
			\State Generate price $\price_t$ from a binary distribution, taking $\pricelb$ or $\priceub$, each with probability $\frac{1}{2}$.
			\State Observes the context $\feat_t$ and demand $\Demand_t$.
			\EndFor
			\State Use data $\Data_{1:\Termone} = \{(\feat_t, \price_t, \Demand_t)\}_{t=1}^{\Termone}$ to derive a pricing policy $\priceone(\feat)$ based on equation~\eqref{eq:priceone}. \vspace{-.5em}
			
			\State
			\emph{Localized Exploration Stage:}
			\For{$t=\Termone+1,\cdots,\Termone + \Termtwo$}
			\State Observe context $\feat_t$.
			\State Set price $\price_t \defn \clip{\priceone(\feat_t) - \perturb}$ or $\clip{\priceone(\feat_t) + \perturb}$, each with probability $\frac{1}{2}$.
			\State Observe demand $\Demand_t$.
			\EndFor
			\State Update the pricing policy $\pricetwo({\feat})$ using data $\Data_{\Termone+1 : \Termone + \Termtwo} = \{({\feat}_t,\price_t,\Demand_t)\}_{t=\Termone+1}^{\Termone + \Termtwo}$. \vspace{.5em}
			
			\State \emph{Committing Stage:}
			\For{$t=\Termone + \Termtwo+1,\cdots,\Term$}
			\State Observe context $\feat_t$ and set price greedily by $\price_t \defn \clip{\pricetwo(\feat_t)}$. 
			\EndFor
		\end{algorithmic}  
	\end{algorithm}
	~
	\vspace{-2.1em}

	We comment that the first two stages of exploration occupy only a small portion of the overall learning process, allowing the algorithm to focus primarily on pure exploitation for most of the time so that the revenue loss in the first two stages is negligible. As will become clear in \Cref{thm:ub_simple} in \Cref{sec:ub}, by setting $\Termone = \bigOtil(\sqrt{\Term})$ and $\Termtwo = \bigO(\Term/\Dim)$ for $ T \gg d $, we achieve optimal rate of regret by effectively balancing exploration and exploitation.
	
	
	\noindent \paragraph{Doubling trick to make the algorithm fully online:}
	We propose a variant of our algorithm suited for a fully online procedure. In its current form, the choices of hyperparameters $\Termone$, $\Termtwo$, and $\perturb$ in \Cref{alg:localized-explore-then-commit} require knowledge about the termination time $\Term$. However, we often want to design an algorithm that achieves minimal regret (up to a constant) at any time $t$.
	
	To address this, we can apply the \emph{doubling trick} \citep{auer1995gambling,besson2018doubling}. Specifically, we double the time horizon iteratively,  which apply (reset) \Cref{alg:localized-explore-then-commit} sequentially for time horizons of lengths $\Term = T_0, 2T_0, 4T_0,8T_0, \ldots$ for a given $T_0$. For each new interval, we recompute the parameters and apply \Cref{alg:localized-explore-then-commit} with the corresponding~$\Term$. This ensures the algorithm performs efficiently without requiring prior knowledge of~$\Term$.
	With this modification, the theoretical guarantees of the original algorithm are preserved, ensuring statistical efficiency throughout the process.
	
	
	\noindent \paragraph{Further refining the algorithm by using time-varying perturbation $\perturb$:}
	We also conjecture that optimal regret could still be achieved by using a time-varying perturbation, $\perturb_t = \bigThetatil(\sqrt{\Dim/t})$, which would eliminate the need for Stage~3 entirely. This modification enables a more dynamic and seamless balance between exploration and exploitation throughout the learning process, potentially improving efficiency in an online setting. However, it requires refitting the model at each time step, which may impose a significant computational burden. Therefore, we maintain the procedure in \Cref{alg:localized-explore-then-commit} due to its simplicity in both practical implementation and theoretical analysis.


\section{Theoretical guarantees}
\label{sec:main}

We now present the main theretical results of our LetC algorithm.
In \Cref{sec:ub_simple}, we start by providing an upper bound on the cumulative regret of our algorithm, which scales as $\sqrt{\Term}$ and is independent of the dimension $\Dim$ of the context vectors $\feat_t$ when $T$ is large. This leads to our first key finding: when the planning horizon $\Term$ is sufficiently large (specifically, when $\Term = \Omegatil(\Dim^4)$), the regret becomes dimension-free.
In \Cref{sec:ub}, we extend this result to a more general, non-asymptotic upper bound on the regret that applies to any planning horizon $\Term \geq 0$. This bound explicitly accounts for the effect of dimensionality when $\Term$ is relatively small. The analysis hinges on a \emph{critical inequality} that balances exploration and exploitation, which becomes particularly important in high-dimensional settings with a limited planning horizon.
Finally, in \Cref{sec:lb}, we establish a minimax lower bound that matches the dimension-free upper bound from \Cref{sec:ub_simple}, confirming the statistical efficiency of our method.

\subsection{Dimension-free regret upper bound for large time horizon $\Term$ \yaqidone}
\label{sec:ub_simple}

Our first main result establishes a dimension-free upper bound on the cumulative regret. 
To build towards this result, we start by outlining the assumptions under which the bound is valid.

\subsubsection{Assumptions}

\paragraph{Regularity on the feature distribution:}
We first impose regularity conditions on the distribution $\distr$ of the feature vectors $\feat_t$:

\begin{assumption}[Regularity on the feature $\feat_t$]
	\label{asp:pure-online-second-moment-nondegenerate}
	\begin{enumerate}  
		\item The feature vectors $\feat_t$ are i.i.d. sub-Gaussian random vectors. Their $\psi_2$-norm is bounded above by a \mbox{constant~$\stdfeat > 0$}, i.e. for all $t$, $\psinorm{2}{\feat_t} \leq \stdfeat$.
		\item The second-moment matrix 
			$\Covfeat \defn \Exp_{\feat \sim \distr}[\feat \feat^{\top}] \in \Real^{\Dim \times \Dim}$
		is well-conditioned.
		Its eigenvalues are bounded as
		\mbox{$\constCovfeat \le \eigmin(\Covfeat)\le \eigmax(\Covfeat) \leq \ConstCovfeat$}
		for some constants \mbox{$\constCovfeat, \ConstCovfeat > 0$}.
	\end{enumerate}
\end{assumption}

\begin{remark}
The second condition in Assumption \ref{asp:pure-online-second-moment-nondegenerate} can be relaxed to the setting where $\Covfeat$ can be expressed as a spiked low-rank part plus a well-conditioned remaining part using factor model techniques~\citep{fan2013large,fan2020factor}.
\end{remark}

\paragraph{Regularity on model parameters:}

In addition to the assumptions on the feature distribution $\distr$, we also impose some regularity conditions on the demand model \eqref{eq:demand}:

\begin{assumption}[Regularity on model parameters]
	\label{asp:basic-regularity}
	There exist constants $\pricelb, \priceub, \featparalb, \featparaub \! > 0$ such that:
	\begin{enumerate}
		\item For every feature ${\feat} \in \FeatSp$, the corresponding optimal price $\pricestar(\feat)$ defined in equation~\eqref{eq:opt_price} lies in the feasibility interval and is bounded away from the boundaries. That is, for some small constant $\bar\delta>0$, $\pricestar({\feat}) \in [\pricelb+\bar\delta, \priceub-\bar\delta]$. Additionally, the range $\priceub - \pricelb$ is of constant order.
		\item For every feature ${\feat} \in \FeatSp$, the intercept $\feat^{\top} \paraintstar$ and the slope $\feat^{\top} \paraslpstar$ in equation \eqref{eq:demand} are also of constant order, satisfying $-{\feat}^{\top} \paraslpstar, {\feat}^{\top} \paraintstar \in [\featparalb, \featparaub]$.  
\end{enumerate}
\end{assumption}

\paragraph{Sufficient exploration conditions:}

Recall that in our model~\eqref{eq:demand}, the expected demand is a linear function of the augmented covariate $(\feat^{\top}, \feat^{\top} \price)^{\top}$. To achieve sublinear regret, $p$ typically converges to $\pricestar(\feat)$ over time. Consequently, the following second-moment matrix plays a crucial role in our analysis:
\begin{align}
	\label{eq:Covparatworealpop}
	\Covparatworealpop =
	\Exp_{\feat \sim \distr} \Bigg[ \begin{bmatrix}
		\feat \\ \feat \, \pricestar(\feat)
	\end{bmatrix}\begin{bmatrix}
		\feat \\ \feat \, \pricestar(\feat)
	\end{bmatrix}^{\top} \Bigg]
	\in \Real^{(2 \Dim) \times (2 \Dim)} \, .
\end{align}

The matrix $\Covparatworealpop$ represents the ``limiting'' second-moment matrix associated with the estimation process, especially when the pricing strategy is estimated with high accuracy. 
To ensure that the data we collect during the learning process effectively explores the feature space, it is crucial that $\Covparatworealpop$ has favorable spectral properties.
In the following, we provide a characterization of the spectral structure of the matrix $\Covparatworealpop$ under relatively mild and natural assumptions, as shown below:
\begin{assumption}[Sufficient exploration]
	\label{asp:exploration}
	\begin{enumerate}
		\item The distribution~$\distr$ of feature $\featt{t}$ satisfies an \emph{anti-concentration} property:
		There exists a constant $\constcon > 0$ such that for any symmetric matrix $\AMt \in \Sym^{\Dim}$ with \mbox{$\rank(\AMt) \leq 4$}, 
		\vspace{-.5em}
		\begin{align}
			\label{eq:ac}
			\Exp_{\feat \sim \distr} \big[ ( \feat^{\top} \AMt \, \feat )^2 \big] \; \geq \; \constcon \cdot \distrnorm{\AMt}{F}^2 \, .
		\end{align} ~ \vspace{-2.5em}
		\item The demand model~\eqref{eq:demand} is non-degenerate, meaning there exists a constant $\const{} > 0$ such that $\distrnorm{\paraintstar}{2} \geq \const{}$ and $\distrnorm{\paraslpstar}{2} \geq \const{}$. Moreover, the vectors $\paraintstar$ and $\paraslpstar$ are not collinear, satisfying $\abs[\big]{(\paraintstar)^{\top} \paraslpstar} \leq (1 - \const{}) \cdot \distrnorm{\paraintstar}{2} \distrnorm{\paraslpstar}{2} $ for a constant $\const{} \in (0,1)$.
	\end{enumerate}
\end{assumption}

\begin{remark}
The condition $\text{rank}(\AMt)\le 4$ is somewhat technical, and we will verify \Cref{asp:exploration} with more transparent conditions later in Section~\ref{sec:exploration}.
\end{remark}

Under the conditions specified in Assumption~\ref{asp:exploration}, we can characterize the spectrum of the matrix $\Covparatworealpop$ as stated in \Cref{lemma:exploration} below. The proof is provided in \Cref{sec:proof:lemma:exploration}.
\vspace{-.5em}
\begin{lemma}
	\label{lemma:exploration}
	The second-moment matrix $\Covparatworealpop \in \Sym^{2 \Dim}$ has the following property:
	\vspace{-.3em}
	\begin{enumerate}  \itemsep = -.2em
		\item The rank of $\Covparatworealpop$ satisfies $\rank(\Covparatworealpop) \leq 2 \Dim - 1$, with the vector $(\paraintstar, 2 \, \paraslpstar) \in \Real^{2 \Dim}$ always lying in its null space.
	\end{enumerate} \vspace{-.3em}
	Furthermore, under Assumptions~\ref{asp:basic-regularity} and~\ref{asp:exploration}, we have \vspace{-.3em}
	\begin{enumerate}   \itemsep = -.2em
		\setcounter{enumi}{1}
		\item \mbox{$\rank(\Covparatworealpop) = 2 \Dim - 1$}, and its null space is one-dimensional, spanned by the vector \mbox{$(\paraintstar, 2 \, \paraslpstar) \in \Real^{2 \Dim}$}.
		\item The second smallest eigenvalue of $\Covparatworealpop$ is bounded below by a positive constant.
	\end{enumerate}
\end{lemma}
As highlighted in \Cref{lemma:exploration}, when the second-moment matrix is singular, this dynamic pricing problem falls into the category of ``incomplete learning'' \citep{keskin2014dynamic, bastani2021mostly} under a greedy algorithm. Without explicit efforts to introduce price dispersion, there is no guarantee that parameter estimates will converge to their true values. The existing literature suggests addressing this issue by employing semi-myopic policies \citep{keskin2014dynamic} such as ETC or UCB. In contrast, we address this problem by learning through active local exploration.

We conclude the introduction of \Cref{asp:exploration} by making a few remarks:
(i) The anti-concentration property ensures that the feature $\featt{t}$ sufficiently explores all directions in the feature space $\FeatSp$, preventing any direction from being underrepresented.
(ii) By ensuring that the vectors $\paraintstar$ and $\paraslpstar$ are not strongly collinear, the optimal pricing function $\pricestar(\feat)$ remains sensitive to changes in the features. This prevents $\pricestar(\feat)$ from being constant and allows for better exploration across the entire feature space.
At the end of \Cref{sec:ub_simple}, we will outline some moment conditions for Assumption~\ref{asp:exploration}.1, and present a simple example to demonstrate that Assumptions~\ref{asp:pure-online-second-moment-nondegenerate}, \ref{asp:basic-regularity} and \ref{asp:exploration} are well-posed and not overly restrictive.

\subsubsection{Main result}

With all the necessary assumptions at hand, we are ready to present the main result of this section: a dimension-free regret bound that holds when the planning horizon~$\Term$ is sufficiently large.

A major contribution of our theory is providing practical recommendations for selecting the hyperparameters $\Termone$, $\Termtwo$, and $\perturb$ in \Cref{alg:localized-explore-then-commit}. We propose setting the lengths of the burn-in period and localized exploration stage as:
\begin{subequations}
\begin{align}
	\label{eq:thm_simple_Termone}
	& \Termone \; \defn \;  \ceil[\big]{{\sqrt{\Term}} \, \log\Term} \, ,  \\
	\label{eq:thm_simple_Termtwo}
	& \Termtwo \; \defn \; \ceil[\Big]{\frac{\Term}{2\Dim}} \, .
\end{align}
With this setup, the initial exploration phase ($\Termone$) becomes negligible in the long run, i.e., $\lim_{\Term \to \infty} \Termone / \Term = 0$ and the revenue loss is at most $O(T_1) =  \bigOtil \big (\sqrt{T} \big )$, while the second stage ($\Termtwo$) uses only a small fraction of the time horizon. This leaves the majority of the horizon focused on committing to and exploiting the learned policy.

For the perturbation level $\perturb$ during localized exploration, our theory requires
\begin{align}
	\label{eq:thm_simple_perturb}
	\Const{0} \sqrt{\big(\Dim/\sqrt{\Term}\big) \log \Term} & \; \leq \; \perturb \; \leq \; \Const{0}'
\end{align}
where $\Const{0}, \Const{0}' > 0$ are constants to be determined later.
The optimal choice of $\perturb$ roughly corresponds to setting $\perturb = \bigOtil \big(\sqrt{\Dim / \Termone}\big)$, which is approximately on the same scale as the estimation error in Stage~1.
\end{subequations}

We now present our main result, \Cref{thm:ub_simple}, which establishes statistical guarantees for the cumulative regret of \Cref{alg:localized-explore-then-commit} using the hyperparameters specified above. The detailed proof can be found in \Cref{sec:proof:thm:ub_simple_0}.

\begin{theorem}[Dimension-free regret upper bound]
	\label{thm:ub_simple}
	Suppose Assumptions~\ref{asp:pure-online-second-moment-nondegenerate}, \ref{asp:basic-regularity} and \ref{asp:exploration} hold. There exist constants $\const{}, \Const{}, \Const{0}, \Const{0}', \Const{1} > 0$, determined solely by the parameters specified in these assumptions, such that the following results hold. If the planning horizon $\Term$ is sufficiently large, satisfying
	\begin{align}
		\label{eq:ub_simple_cond}
		\Term \; \geq \; \Const{1}  \cdot \Dim^4 \, \log^2 \Term 
	\end{align}
	and the hyperparameters are set according to equations \eqref{eq:thm_simple_Termone}-\eqref{eq:thm_simple_perturb}, then \Cref{alg:localized-explore-then-commit} achieves the following guarantees:
	\begin{subequations}
	\begin{enumerate}
	\item[(a)] With probability at least $1 - \const{} \, \Term^{-2}$, the cumulative regret is bounded by
	\begin{align}
		\label{eq:ub_simple_whp}
		\Reg(\Term) \; \leq \; \Const{} \; \Big\{ \sqrt{\Term} \; \log\Term \; + \; \frac{\perturb^2 \, \Term}{\Dim} \Big\} \, .
	\end{align}
	\item[(b)] Additionally, the expected cumulative regret satisfies
	\begin{align}
		\label{eq:ub_simple_exp}
		\mathbb{E}[\Reg(\Term)]  \; \leq \; \Const{} \; \Big\{ \sqrt{\Term} \; \log\Term \; + \; \frac{\perturb^2 \, \Term}{\Dim} \Big\} \, .
	\end{align}
	\end{enumerate}
	\end{subequations}
\end{theorem} ~

We highlight that by choosing the perturbation level $\perturb$ as
\begin{align*}
	\perturb \; \asymp \; \sqrt{\big(\Dim/\sqrt{\Term}\big) \log \Term} \,, 
\end{align*}
the regret bound in \eqref{eq:ub_simple_whp} (and similarly in \eqref{eq:ub_simple_exp}) achieves its optimal order:
\begin{align}
	\label{eq:ub_simple}
	\Reg(\Term) \; \leq \; \Const{} \; \sqrt{\Term} \; \log\Term  \, .
\end{align}
This bound is \emph{dimension-free}, meaning the regret does not grow with the dimension $\Dim$. Intuitively, this is because when $T$ is large, we only need to actively explore in the singular direction, as indicated by Assumption \ref{asp:exploration}.
To reach this dimension-free regime, a larger planning horizon $\Term$ is needed in higher-dimensional settings, as indicated by condition \eqref{eq:ub_simple_cond}. Nevertheless, as $\Term$ increases, the cumulative regret eventually converges to the dimension-free rate of $\bigOtil(\sqrt{\Term})$, regardless of the feature dimension.
We will further illustrate this behavior with simulation studies in \Cref{sec:sim}.

\paragraph{Comparison to \cite{ban2021personalized}.}
	The contextual linear pricing framework was first introduced in \cite{ban2021personalized}. We highlight a few key differences in assumptions and results.
	The major difference is that we impose the benign eigen structure of $\Covparatworealpop$, the ``limiting'' second-moment matrix associated with the estimation
    process. We provide a clear and transparent condition to ensure this, as outlined in Assumption \ref{asp:exploration}. In the same linear modeling framework as ours, \cite{ban2021personalized} establishes a regret bound of $\calO(d\sqrt{T})$.
	To address high-dimensional settings, they further derive a regret bound of  $\calO(s\sqrt{\Term})$ under a sparsity assumption. In contrast, we are able to even eliminate $s$ using better algorithmic design and relatively mild assumptions.

\subsubsection{Comments on the anti-concentration property in Assumption \ref{asp:exploration}}
\label{sec:exploration}

To conclude this subsection, we highlight a few moment conditions that guarantee the anti-concentration property (Assumption \ref{asp:exploration}.1), and then present a simple problem instance that meets \mbox{Assumptions~\ref{asp:pure-online-second-moment-nondegenerate}-\ref{asp:exploration}}, as required in \Cref{thm:ub_simple}. The example illustrates that the assumptions are reasonable and compatible.

\begin{lemma}
Let $\bepsilon = (\epsilon_{1}, \epsilon_{2}, \ldots, \epsilon_{\Dim})^{\top}   \in \Real^{\Dim}$ be a random vector with independent coordinates, and $\feat=\bB\bepsilon  \in \Real^{\Dim}$.   Assume the following moment conditions:
\begin{enumerate}
\item \label{item:mo1}$\Exp[\epsilon_{i}] = \Exp[\epsilon_{i}^3] = 0$ for $i = 2, 3, \ldots, \Dim$; 
\item \label{item:mo2}$\Exp[\epsilon_{i}^2] = 1$ and $\Exp[\epsilon_{i}^4] > 1+c_{\textsf{mo}}$ for some constant $c_{\textsf{mo}}>0$ and $i = 1, 2, \ldots, \Dim$.
\item \label{item:mo3} the minimum singular value of $\bB$ is bounded as $\lambda_{\min}(\bB)\ge c_0>0$.
\end{enumerate}
Then, the anti-concentration property \eqref{eq:ac} holds for any symmetric matrix $\AMt$.
\end{lemma}



The second condition essentially states that the kurtosis of each coordinate is strictly greater than $1$, which holds for nondegenerate distributions. Many common distributions, such as the Gaussian and Uniform distributions, satisfy both conditions \ref{item:mo1} and \ref{item:mo2}. We now prove the lemma.

\begin{proof}  Let first consider the specific case $\bB = \bI_d$ so that $\feat = \bepsilon$.  For any symmetric matrix $\AMt \in \Sym^{\Dim}$,  by leveraging the conditions that $\Exp[\feati{i}] = \Exp[\feati{i}^3] = 0$ for $i = 2, 3, \ldots, \Dim$, we derive through some straightforward algebraic calculations
	\begin{align*}
		\Exp\big[ (\feat^{\top} \AMt \, \feat)^2 \big]
		\; = \; \sum_{i=1}^{\Dim} \AMth{i,i}^2 \cdot \Exp\big[ \feati{i}^4 \big] + \sum_{i \neq j} \big\{ \AMth{i,i} \AMth{j,j} + 2 \, \AMth{i,j}^2 \big\} \cdot \Exp\big[ \feati{i}^2 \, \feati{j}^2 \big] \, .
	\end{align*}
	Next, using the relations $\Exp[\feati{i}^2] = 1$ and $\Exp[\feati{i}^4] > 1+c_{\textsf{mo}}$ for $i = 1, 2, \ldots, \Dim$, we obtain
	\begin{align*}
		\Exp\big[ (\feat^{\top} \AMt \, \feat)^2 \big]
		& \; \geq \; (1+c_{\textsf{mo}}) \, \sum_{i=1}^{\Dim} \AMth{i,i}^2 + \sum_{i \neq j} \big\{ \AMth{i,i} \AMth{j,j} + 2 \, \AMth{i,j}^2 \big\}  \\
		& \; \ge \; \bigg\{ \sum_{i=1}^{\Dim} \AMth{i,i}^2 + \sum_{i \neq j} \AMth{i,i} \AMth{j,j}  \bigg\} + \min(2,c_{\textsf{mo}}) \, \bigg\{ \sum_{i=1}^{\Dim} \AMth{i,i}^2 + \sum_{i \neq j} \AMth{i,j}^2 \bigg\}  \\
		& \; = \; \trace(\AMt)^2 + \min(2,c_{\textsf{mo}}) \, \distrnorm{\AMt}{F}^2
		\; \geq \; \min(2,c_{\textsf{mo}}) \, \distrnorm{\AMt}{F}^2 \, .
	\end{align*}
	
 Now, for the general case, for any symmetric $\AMt$, it follows from the above result that
	\begin{align*}
		\Exp\big[ ( \feat^{\top} \AMt \, \feat )^2 \big]= \Exp\big[ ( \bepsilon^{\top} \bB^\top\AMt \bB \bepsilon )^2 \big]\geq \min(2,c_{\textsf{mo}})  \distrnorm{\bB^\top\AMt\bB}{F}^2\ge c_0^2\min(2,c_{\textsf{mo}})  \distrnorm{\AMt}{F}^2\, .
	\end{align*}
In other words, we have shown that the anti-concentration property \eqref{eq:ac} is indeed satisfied with $\constcon = \min(2,c_{\textsf{mo}}) c_0^2 >0$. 
\end{proof}

\begin{example}
	In this example, we consider features $\feat = (\feati{1}, \feati{2}, \ldots, \feati{\Dim})^{\top} \in \Real^{\Dim}$, where the entries $\{\feati{i}\}_{i=1}^{\Dim}$ are independently distributed according to
		\begin{align*}
			& \Prob\Big\{ \feati{1} = \frac{1}{2} \Big\} = \frac{4}{5},
			\quad  \Prob\{ \feati{1} = 2 \} = \frac{1}{5} \, ,  \\
			& \Prob\{ \feati{i} = 2 \} = \Prob\{ \feati{i} = - 2 \} = \frac{1}{8},
			\quad \Prob\{ \feati{i} = 0 \} = \frac{3}{4} \qquad
			\mbox{for $i = 2, 3, \ldots, \Dim$} \, .
		\end{align*}
	It is straightforward to show that \Cref{asp:pure-online-second-moment-nondegenerate} naturally holds for this feature distribution~$\distr$. Also, $\distr$ satisfies the above moment conditions \ref{item:mo1} and \ref{item:mo2}, and hence Assumption \ref{asp:exploration}.1.

	We now turn to selecting a demand model \eqref{eq:demand} with parameters $\paraintstar$ and $\paraslpstar \in \Real^{\Dim}$ that satisfy Assumptions~\ref{asp:basic-regularity} and \ref{asp:exploration}.2.
	
	Let $\vece_1 = (1, 0, 0, \ldots, 0)^{\top} \in \Real^{\Dim}$ be the unit vector with its first entry as one and all others as zero. To satisfy \Cref{asp:basic-regularity}, we pick vectors $\paraintstar \approx \vece_1$ and $\paraslpstar \approx -\vece_1$.
	Specifically, we set $\paraintstar$ such that $\distrnorm{\paraintstar - \vece_1}{1} \leq \frac{1}{8}$. From this choice, we can derive
	\begin{align*}
		\feat^{\top} \paraintstar \; \geq \; \feat^{\top} \vece_1 - \abs[\big]{\feat^{\top}(\paraintstar - \vece_1)} \; \geq \; \feati{1} - \supnorm{\feat} \distrnorm{\paraintstar - \vece_1}{1} \; \geq \; \frac{1}{2} - 2 \cdot \frac{1}{8} \; = \; \frac{1}{4} \, .
	\end{align*}
	Similarly, it can be shown that $\feat^{\top} \paraintstar \leq 3$.
	We also select $\paraslpstar$ such that $\distrnorm{\paraslpstar + \vece_1}{1} \leq \frac{1}{8}$, which ensures $\frac{1}{4} \leq - \feat^{\top} \paraslpstar \leq 3$.
	These bounds on $\feat^{\top} \paraintstar$ and $\feat^{\top} \paraslpstar$ imply that the optimal price $\pricestar$ is also bounded, thereby satisfying \Cref{asp:basic-regularity}.
	
	As for \Cref{asp:exploration}.2, we can easily satisfy the condition by choosing reasonable vectors for $\paraintstar$ and $\paraslpstar$ within their feasible sets $\distrnorm{\paraintstar - \vece_1}{1} \leq \frac{1}{8}$ and $\distrnorm{\paraslpstar + \vece_1}{1} \leq \frac{1}{8}$. For instance, setting
	$\paraintstar = (1, \frac{1}{8}, 0, 0, \ldots, 0)^{\top} \in \Real^{\Dim}$ and $\paraslpstar = (-1, \frac{1}{8}, 0, 0, \ldots, 0)^{\top} \in \Real^{\Dim}$
	ensures that the model is non-degenerate and these vectors are not collinear. Therefore, \Cref{asp:exploration}.2 holds.

\end{example}


\subsection{Non-asymptotic regret upper bound for any time horizon $\Term$ \yaqidone}
\label{sec:ub}

In \Cref{sec:ub_simple}, we established a regret upper bound that, surprisingly, turned out to be dimension-free. However, the result only applies when the planning horizon $\Term$ is sufficiently large, specifically requiring $\Term = \Omegatil(\Dim^4)$. It naturally raises the following questions: what happens when~$\Term$ is smaller? Can we still effectively control regret in such cases?

To address this limitation, we extend our analysis to shorter planning horizons by conducting a more detailed study of the spectral properties of the covariance matrix~$\Covparatworealpop$ defined in equation \eqref{eq:Covparatworealpop}. This extended analysis builds on Assumptions~\ref{asp:pure-online-second-moment-nondegenerate} and \ref{asp:basic-regularity} from \Cref{sec:ub_simple}, without relying on \Cref{asp:exploration}.

We start by establishing a critical inequality in \Cref{sec:CI}, which forms the backbone of our analysis. In \Cref{sec:thm_main}, we derive a new regret bound that holds for any $\Term \geq 0$, expanding the applicability of our results. Our theory also provides practical guidance on tuning the hyperparameters in \Cref{alg:localized-explore-then-commit} to achieve an effective balance between exploration and exploitation in a broader range of scenarios.

Finally, in \Cref{sec:ridge}, we draw connections to ridge regression and show that (i) localized exploration in online dynamic pricing resembles adding a ridge penalty in regression, and (ii) our critical inequality closely mirrors that in kernel ridge regression. While the critical inequality in kernel ridge regression characterizes the bias-variance tradeoff, ours focuses on the exploration-exploitation tradeoff. This comparison highlights how our method adapts regularization techniques to dynamic settings while introducing novel adjustments for online decision-making.

\subsubsection{Critical inequality \yaqidone}
\label{sec:CI}

Let us begin by introducing a critical inequality that serves as a cornerstone of our analysis. This inequality centers on the spectrum of the second-moment matrix $\Covparatworealpop$, which, for reference, is defined as follows:
\begin{align}
	\Covparatworealpop =
	\Exp_{\feat \sim \distr} \Bigg[ \begin{bmatrix}
		\feat \\ \feat \, \pricestar(\feat)
	\end{bmatrix}\begin{bmatrix}
		\feat \\ \feat \, \pricestar(\feat)
	\end{bmatrix}^{\top} \Bigg]
	\in \Real^{(2 \Dim) \times (2 \Dim)} \, .
	\tag{\ref{eq:Covparatworealpop}}
\end{align}
Let $\eigpop_1 \geq \eigpop_2 \geq \ldots \geq \eigpop_{2\Dim-1} \geq \eigpop_{2 \Dim} = 0$ denote the eigenvalues of matrix~$\Covparatworealpop$, ordered non-increasingly. 

\paragraph{Degenerate dimension:}
To simplify our analysis of the online learning process and define the critical inequality, we first introduce a concept called the \emph{degenerate dimension}, defined as
\begin{align}
	\label{eq:EffDim}
	\EffDim(\perturb) \defn \sum_{k=1}^{2\Dim} \min \Big\{ \frac{\perturb^2}{\eigpop_k}, 1 \Big\} \, .
\end{align}
From \Cref{lemma:exploration},  the smallest eigenvalue $\eigpop_{2\Dim}$ is zero. Therefore, the degenerate dimension always satisfies $1 \leq \EffDim(\perturb) \leq 2\Dim$, although it may not necessarily be an integer.

\paragraph{Critical inequality:}
We set the perturbation level $\perturb$ as a small, positive solution to the \emph{critical inequality}:
\begin{align}
	\label{eq:CI}
	\Singular(\perturb) \defn \sqrt{\frac{\EffDim(\perturb)}{2\Dim} }
	\ \geq \; \underbrace{\ConstCI \, \sqrt{\frac{2\Dim}{\Term}} \, \log \Term}_{\SNR^{-1}} \, \cdot \, \frac{1}{\perturb^2} \, .
\end{align}
In this expression, $\ConstCI > 0$ is a constant that will be specified later in \Cref{thm:ub_main}. 

The left-hand side of the critical inequality~\eqref{eq:CI}   features a singularity function at level $\perturb$, which we denote as $\Singular(\perturb)$. This function captures the ``proportion'' of eigenvalues that fall below the threshold~$\perturb^2$. 
It starts at a minimum value of at least $1/\sqrt{2\Dim}$ when $\perturb$ is close to zero and increases continuously to $1$ as $\perturb$ grows larger.
\footnote{
	To understand why $\Singular(\perturb)$ ranges between $1/\sqrt{2\Dim}$ and $1$, let us revisit the properties of the covariance matrix $\Covparatworealpop$. According to \Cref{lemma:exploration}, this matrix is rank deficient, meaning it has some zero eigenvalues. As a result, when $\perturb \rightarrow 0^+$, the singularity function behaves as
	$\Singular(0^+) = \sqrt{{\#\{ k \mid \eigpop_k = 0 \}}/{(2\Dim)}} \geq {1}/{\sqrt{2 \Dim}}$.
	On the other hand, as $\perturb$ increases, once it exceeds the largest eigenvalue, i.e., $\perturb \geq \sqrt{\eigpop_1}$, all the eigenvalues become smaller than $\perturb^2$. In this case, the singularity function reaches its maximum value
	$\Singular(\perturb) = 1$.
}

The right-hand side of the critical inequality \eqref{eq:CI} is a decreasing function with respect to the parameter $\perturb$. It diverges to $+\infty$ as $\perturb \rightarrow 0^+$ and decreases to $0$ as $\perturb \rightarrow +\infty$. The inequality involves the inverse of a \emph{signal-to-noise ratio} (SNR), defined as
\begin{align}
	\label{eq:SNR}
\SNR \; \defn \; \frac{\sqrt{\Term} \, / \log\Term}{\ConstCI \, \sqrt{2 \Dim}} \, .
\end{align}
The value of $\SNR$ increases with a longer planning horizon $\Term$, indicating stronger signals as more information is collected over time. In contrast, $\SNR$ decreases as the dimension $\Dim$ increases since the signal strength is diluted across multiple directions, weakening its impact.

In \Cref{fig:CI} below, we illustrate the basic geometry of the critical inequality \eqref{eq:CI} to provide a more intuitive understanding of its structure. This figure shows how varying levels of $\SNR$ influence the behavior of the critical inequality and, consequently, affect the solutions. 

\begin{figure}[!ht]
	\centering
	\begin{tabular}{ccc}
		\hspace{-2em}
		\includegraphics[width = .5 \linewidth]{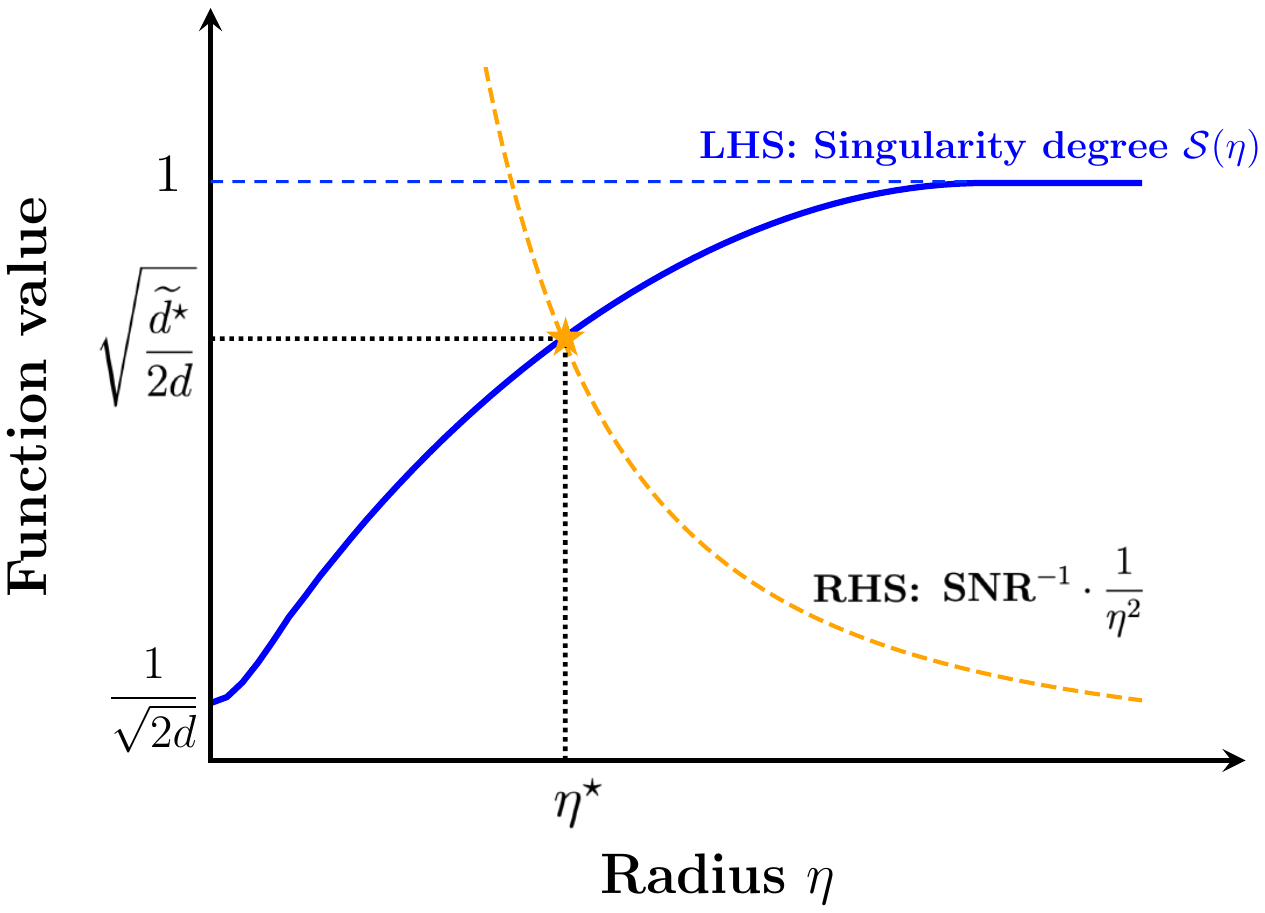} &&
		\hspace{-2em}
		\includegraphics[width = .5 \linewidth]{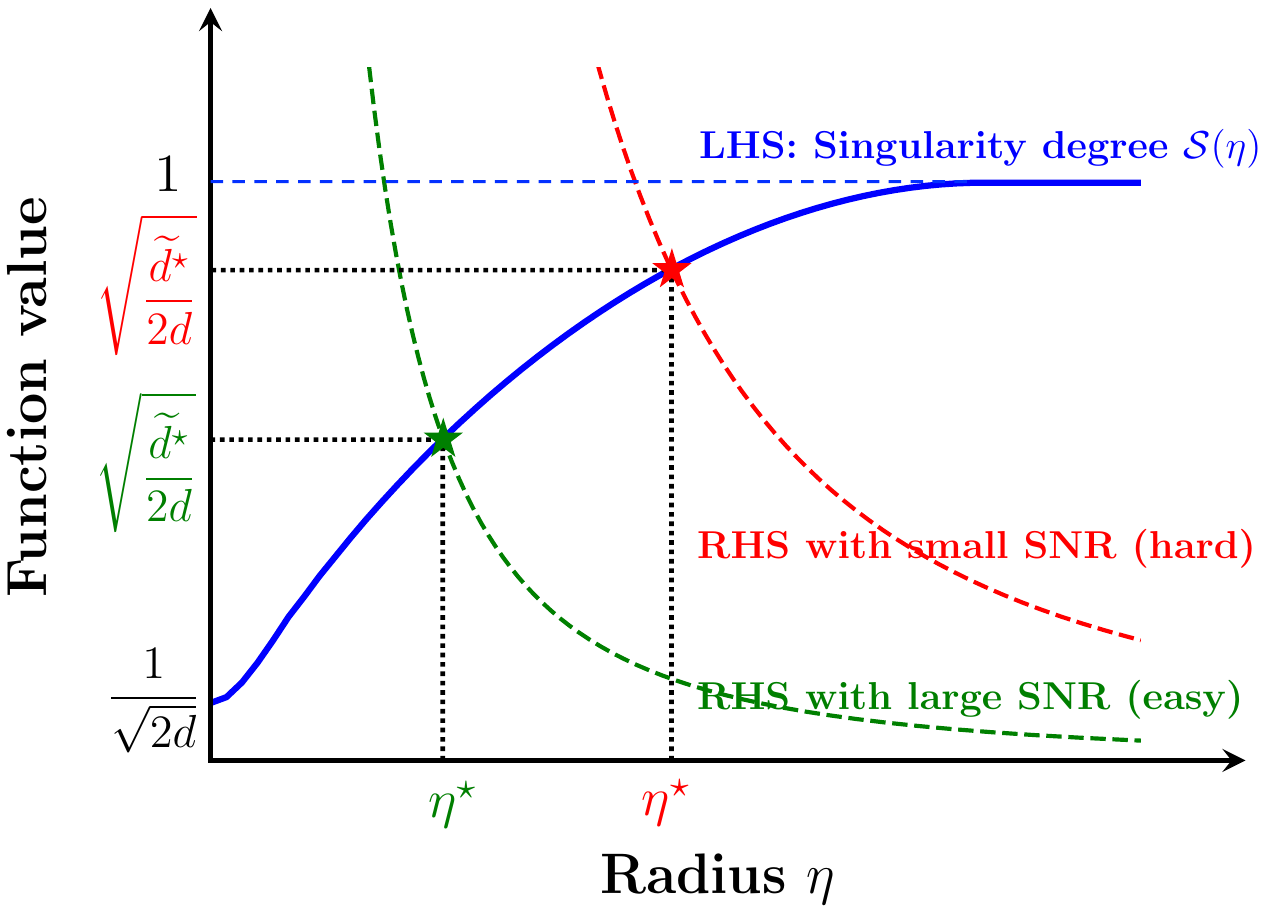} \\
		\hspace{0.7em} (a) && \hspace{0.7em} (b)
	\end{tabular}
	\caption{Illustrations of the structure of the critical inequality~\eqref{eq:CI}.
	(a) The left plot shows the singularity function \mbox{$\perturb \mapsto \Singular(\perturb)$} on the left-hand side and the function \mbox{$\perturb \mapsto \SNR^{-1} \perturb^{-2}$} on the right-hand side. The critical point $\perturbstar$ is identified at the intersection of these two curves, marked with a star.
	(b) The right plot illustrates the impact of varying the signal-to-noise ratio $\SNR$ on the right-hand side curve. As $\SNR$ decreases, indicating a more challenging problem, the curve shifts upwards, causing the critical value $\perturbstar$ to move to the right, corresponding to larger values of solutions. On the $y$-axis of both plots, $\tilde d^*$ stands for the degenerate dimension~\eqref{eq:EffDim} at $\eta=\perturbstar$. \vspace{-1em}}
	\label{fig:CI}
\end{figure}

\paragraph{Critical radius $\perturbstar$ and regular solution:}
Due to the monotonic behavior of the two sides of the critical inequality~\eqref{eq:CI}, they intersect at exactly one point, leading to a unique solution for $\perturb$ that satisfies the equality.
We refer to this intersection point as the critical radius, denoted by $\perturbstar$. Formally, we define it as
$\perturbstar \defn \min \{ \perturb > 0 \mid \text{$\perturb$ satisfies \eqref{eq:CI}} \}$.
In other words, $\perturbstar$ represents the smallest positive solution to the inequality.

In practice, however, computing the exact value of $\perturbstar$ is generally infeasible, so we aim to approximate it instead. For this purpose, we introduce the concept of a \emph{regular} solution:
A solution $\perturb$ is termed \emph{$\zeta$-regular} for some parameter $\zeta > 1$ if the scaled value $\perturb/\zeta$ no longer satisfies the critical inequality \eqref{eq:CI}. In cases where $\zeta$ is unspecified, we refer to $\perturb$ simply as \emph{regular}, with $\zeta$ treated as an adjustable constant for controlling the approximation quality.

\vspace{1em}

The concept of degenerate dimension extends the idea of effective or statistical dimension, which is widely used in kernel ridge regression (see \cite{yang2017randomized}). Traditionally, effective dimension is determined by counting the eigenvalues that exceed a given threshold. In contrast, our approach emphasizes eigenvalues below the cutoff $\perturb^2$. This reframing shifts the focus to identifying directions where the model is less confident. These directions of singularity indicate areas where further exploration is needed, thereby guiding the learning process more effectively.

\subsubsection{Main theorem \yaqidone}
\label{sec:thm_main}

With the stage now set, we are ready to present our main theorem. Our theory recommends setting the perturbation level $\perturb$ in the localized exploration stage to satisfy the critical inequality~\eqref{eq:CI}:
\begin{subequations}
\begin{align}
	\label{eq:thm_main_perturb}
	\mbox{Choose $\perturb \leq \Const{0}'$ such that it solves the critical inequality~\eqref{eq:CI}.}
\end{align}
The degenerate dimension $\EffDim(\perturb)$ at the chosen perturbation level $\perturb$ is hard to determine in practice. We make a guess $\EffDim$ such that $\EffDim(\perturb)\le\EffDim \le 2\Dim$, and then set the lengths of exploration stages as follows:
\begin{align}
	\label{eq:thm_main_Termone}
	& \Termone \; \defn \;  \ceil[\Big]{{\sqrt{\EffDim \, \Term}} \, \log\Term} \\
	\label{eq:thm_main_Termtwo}
	& \Termtwo \; \defn \; \ceil[\bigg]{\frac{\EffDim}{2 \, \Dim} \cdot \Term} \, .
\end{align}
\end{subequations}
With the above configuration, we establish non-asymptotic guarantees on the cumulative regret, as stated in \Cref{thm:ub_main}, without \Cref{asp:exploration} for a wider range of $T$. The full proof can be found in \Cref{sec:proof:thm:ub_main_0}.

\begin{theorem}[A general regret upper bound]
	\label{thm:ub_main}
	Suppose Assumptions~\ref{asp:pure-online-second-moment-nondegenerate} and \ref{asp:basic-regularity} hold. Let $\ConstCI, \const{}, \Const{}, \Const{0}', \Const{1} > 0$ be constants determined solely by the parameters specified in these assumptions. If the planning horizon $\Term$ is sufficiently large, such that
	\begin{align}
		\label{eq:ub_main_cond}
		\Term \; \geq \; \Const{1}  \cdot \Dim \, \log^2 \Term 
	\end{align}
	and the hyperparameters are set according to equations \eqref{eq:thm_main_perturb}-\eqref{eq:thm_main_Termtwo}, then \Cref{alg:localized-explore-then-commit} achieves the following guarantees:  
	\begin{subequations}
		\begin{enumerate}
			\item[(a)] With probability at least $1 - \const{} \, \Term^{-2}$, the cumulative regret is bounded by
			\begin{align}
				\label{eq:ub_main_whp}
				\Reg(\Term) \; \leq \; \Const{} \; \Big\{ \sqrt{\EffDim \, \Term} \, \log \Term 
				+ \frac{\EffDim}{2 \, \Dim} \cdot \perturb^2 \, \Term \Big\} \, .
			\end{align}
			\item[(b)] Additionally, the expected cumulative regret satisfies
			\begin{align}
				\label{eq:ub_main_exp}
				\mathbb{E}[\Reg(\Term)]  \; \leq \; \Const{} \; \Big\{ \sqrt{\EffDim \, \Term} \, \log \Term + \frac{\EffDim}{2 \, \Dim} \cdot \perturb^2 \, \Term \Big\} \, .
			\end{align}
		\end{enumerate}
	\end{subequations}
\end{theorem}

We now provide some clarifications on how to refine the regret bound by properly scaling the parameters $\perturb$ and $\EffDim$.
\begin{subequations}
\begin{itemize}
\item If $\perturb$ satisfies the critical inequality~\eqref{eq:CI} as a regular solution\footnote{Recall that a regular solution defined at the end of \Cref{sec:CI} means the solution is near the critical point of \eqref{eq:CI}.}, the regret bound stated in \eqref{eq:ub_main_whp} (and similarly in \eqref{eq:ub_main_exp}) can be tightened to
\begin{align}
	\label{eq:ub_main_regular}
	\Reg(\Term) \; \leq \; \Const{} \; \Bigg\{ \sqrt{\EffDim} + \frac{\EffDim}{\sqrt{\EffDim(\perturb)}} \Bigg\} \, \sqrt{\Term} \, \log\Term \, .
\end{align}
The proof of this claim can be found in \Cref{sec:proof:thm:ub_main_0}.
\item  If, in addition, we have $\EffDim \, \asymp \, \EffDim(\perturb)$, the regret bound simplifies even further to
\begin{align}
	\label{eq:ub_main}
	\Reg(\Term) \; \leq \; \Const{} \; \sqrt{\EffDim \, \Term} \; \log\Term \, .
\end{align}
In this case, the extra complexity term vanishes, resulting in the optimal scaling of the regret.
\end{itemize}
\end{subequations}
~

Below, we provide key insights into the implications of our main theorem.

\paragraph{Characterization of regret across the full range of $\Term$:}

One of the key contributions of \Cref{thm:ub_main} is its ability to characterize the regret behavior across the entire range of planning horizons $\Term$.  Importantly, the lower bound~\eqref{eq:ub_main_cond} on the planning horizon $\Term$ required for our analysis is quite mild. When $\Term < \Const{1} \cdot \Dim \log^2 \Term$, the regret can always be bounded as
\begin{align}
	\label{eq:ub_T}
	\Reg(\Term) \; \leq \; \Const{} \; \Term \, .
\end{align}
This, combined with the bound \eqref{eq:ub_main}, provides a comprehensive characterization of the regret for any $\Term \geq 0$.

We identify two critical boundaries where the behavior of the regret changes:
\begin{itemize}
	\item For $\Term \asymp \Dim \, \log^2 \Term$, we have $\EffDim \asymp \Dim$, meaning that the regret bounds~\eqref{eq:ub_main} and \eqref{eq:ub_T} align up to constant factors. This indicates that our bound seamlessly transitions from the linear regime to the square-root regime as $\Term$ increases.
	\item For $\Term \asymp \Dim^4 \, \log^2 \Term$ and under Assumption \ref{asp:exploration}, we find that $\EffDim \asymp 1$, which reduces the bound~\eqref{eq:ub_main} in \Cref{thm:ub_main} to match the dimension-free bound in \Cref{thm:ub_simple}, as given by \eqref{eq:ub_simple}.
\end{itemize}
We will rigorously establish these transitions in \Cref{sec:turning_points}, focusing on how the degenerate dimension $\EffDim$ scales at these key boundaries.

For $\Term$ values below the first transition point, the linear bound \eqref{eq:ub_T} applies. For $\Term$ beyond the second transition point, the dimension-free bound \eqref{eq:ub_simple} dominates. Between these two points, the bound \eqref{eq:ub_main} adapts smoothly by adjusting the degenerate dimension~$\EffDim$ according to the value of $\Term$. This dynamic adjustment allows the bound to seamlessly cover the entire range of planning horizons.

\begin{figure}[!ht]
	\centering
	\includegraphics[width = .53 \linewidth]{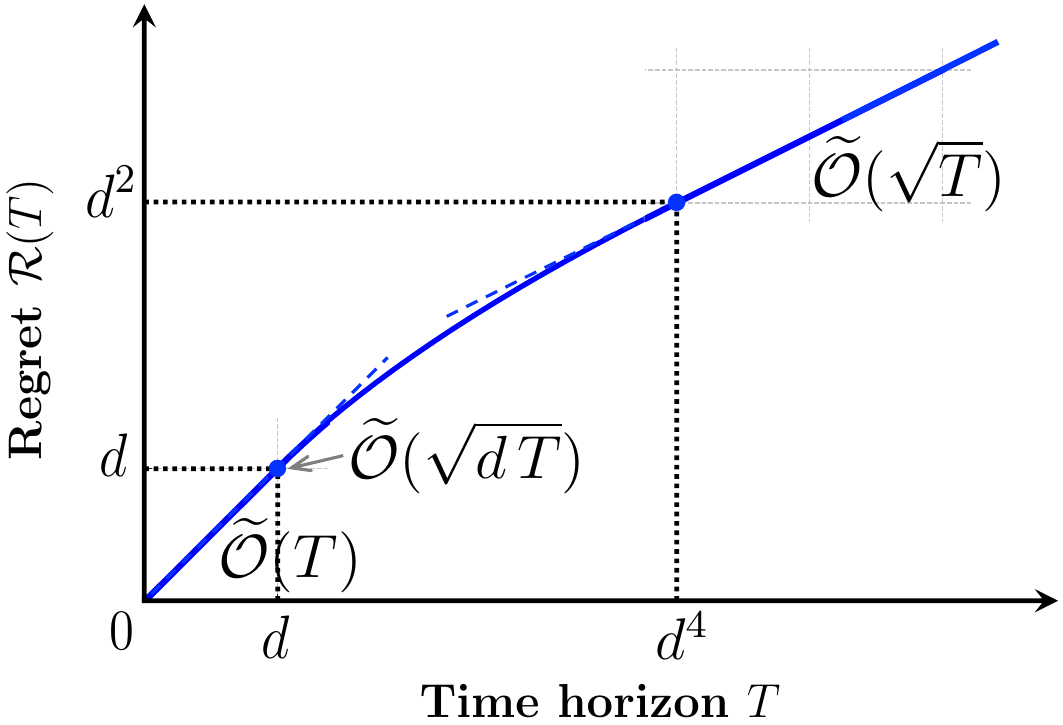} \vspace{.5em}
	\caption{(Log-log) plot of the regret upper bound versus the time horizon $\Term$, omitting constant and logarithmic factors. (i) Initially, when $\Term = \bigOtil(\Dim)$, the regret grows linearly. (ii) At the first transition point $\Term = \bigThetatil(\Dim)$, it shifts smoothly into the square-root regime~$\bigOtil(\sqrt{\Dim \, \Term})$. \mbox{(iii) In the} intermediate range $\Term = \Omegatil(\Dim)$ and $\Term = \bigOtil(\Dim^4)$, the regret scales as $\bigO\big(\sqrt{\EffDim \, \Term}\big)$, with $\EffDim$ decreasing from $\Dim$ to $1$. (iv) Beyond the second transition point at $\Term = \bigThetatil(\Dim^4)$, the regret becomes dimension-free, scaling as $\bigOtil(\sqrt{\Term})$. \vspace{-1em}}
	\label{fig:rate}
\end{figure}

The full regret bound, covering all values of $\Term \geq 0$, is summarized as
\begin{align}
	\label{eq:ub_full}
	\Reg(\Term) \; \lesssim \;
	\begin{cases}
		\Term  &  \mbox{if } \Term \lesssim \Dim \, \log^2 \Term \, , \\
		\sqrt{\EffDim \, \Term} \; \log\Term  &  \mbox{if } \Dim \, \log^2 \Term \lesssim \Term \lesssim \Dim^4 \, \log^2 \Term \, ,  \\
		\sqrt{\Term} \; \log\Term  &  \mbox{if } \Term \gtrsim \Dim^4 \, \log^2 \Term \, .
	\end{cases}
\end{align}
A visual representation of this bound is provided in \Cref{fig:rate}.

\paragraph{Exploitation-exploration tradeoff:}

The critical inequality~\eqref{eq:CI} encapsulates the tradeoff between exploitation and exploration. To better understand this, we can equivalently reformulate it as
\begin{align}
	\label{eq:CI_reform}
	\perturb^2 \; \geq \; \frac{1}{\SNR \cdot \Singular(\perturb)} \, .
\end{align}

In the reformulated inequality, the left-hand side represents the regret loss caused by exploration during Stage 2. Increasing the strength of the exploration noise $\perturb$ results in higher regret from this exploration phase. On the other hand, the right-hand side captures the revenue loss due to inaccurate estimation of demand function during exploitation in Stage 3, which depends on the data collected using the specified level of $\perturb$. A  larger value of $\perturb$ increases $\Singular(\perturb)$ and reduces the right-hand side of the inequality. This means that using a higher level of exploration noise improves the coverage of the data collected, leading to lower error during the exploitation phase.

The inequality~\eqref{eq:CI_reform} highlights the delicate balance between these two opposing effects. When it holds as equality, it indicates an optimal balance, achieving the best possible tradeoff between exploration and exploitation.

\subsubsection{Connection with ridge regression \yaqidone}
\label{sec:ridge}

We now discuss the connection between our localized exploration-then-commit method for dynamic pricing and ridge regression.

\paragraph{Localized exploration as a ridge penalty:}
In terms of estimating the parameter $\paratwo$, adding exploration noise with magnitude $\perturb$ has an effect similar to adding a ridge regularization term $\perturb^2 \norm{\para}_2^2$ in the linear regression problem~\eqref{eq:paratwo}. To illustrate this connection, we present an informal version of \Cref{lemma:paratwo_err_simple}. The full, detailed version is provided in \Cref{lemma:paratwo_err} in \Cref{sec:key_lemma}.

\begin{lemma}[Estimation error in $\paratwo$ from Stage~2 (informal)]
	\label{lemma:paratwo_err_simple}
	Under suitable assumptions and with properly chosen hyperparameters, there exists a constant $\Const{} > 0$ such that, with high probability
	\begin{align}
		\label{eq:paratwo_err_0}
		\norm[\big]{\paratwo-\parastar}_2^2 \; \leq \; \Const{} \cdot \bigg\{ \sum_{k=1}^{2\Dim} \frac{1}{\eigpop_k + \perturb^2} \bigg\} \frac{\log \Term}{\Termtwo} \, .
	\end{align}
\end{lemma}

The bound \eqref{eq:paratwo_err_0} reveals that increasing the noise level $\perturb$ effectively increases the eigenvalues of the covariance matrix $\Covparatworealpop$ by $\perturb^2$, analogous to the effect of ridge regularization. This improved conditioning reduces the estimation error, similar to how ridge regression stabilizes estimates in ill-conditioned problems.

\paragraph{Comparison with the critical inequality:}
In this work, one of our key contributions is to extend the framework of critical inequalities, originally developed for kernel ridge regression (e.g., \cite{bartlett2005local,yang2017randomized,wainwright2019high,duan2021optimal,duan2022policy}), to an online learning setting where the focus shifts from the bias-variance tradeoff to the exploration-exploitation tradeoff. These critical inequalities share a common principle: they examine the spectrum of the covariance matrix to characterize effective or degenerate dimensions. However, they target different balances depending on the specific regime of interest.

To delineate the connection between these two frameworks, let us revisit the critical inequality used in kernel ridge regression:
\begin{align}
	\label{eq:CI_KRR}
	\mathcal{C}(\delta) \defn \sqrt{\sum_{j=1}^{\infty} \min\Big\{ \frac{\mu_j}{\delta^2}, 1 \Big\}} \; \leq \; \SNR \cdot \delta \, .
\end{align}
Here $\mu_1 \geq \mu_2 \geq \ldots \geq 0$ are the eigenvalues of a kernel-based covariance operator. The left-hand side $\mathcal{C}(\delta)$, known as the kernel-complexity function at radius $\delta$, reflects the \emph{variance} incurred by using basis directions where the signal strength exceeds the threshold $\delta^2$. The quantity $\mathcal{C}^2(\delta)$ is known as the \emph{statistical dimension}.
The right-hand side, a linear function of $\delta$, represents the \emph{bias} introduced by cutting off bases below this threshold $\delta^2$, with the slope determined by a signal-to-noise ratio $\SNR$. Tuning KRR according to the solution to this inequality captures the optimal balance between variance and bias.

Our critical inequality~\eqref{eq:CI} parallels \eqref{eq:CI_KRR} but is adapted to the online learning context, where the focus shifts to exploring the degenerate dimensions and the singularity $\Singular(\perturb)$ of the covariance matrix. Unlike the kernel ridge regression setting, where the complexity $\mathcal{C}(\delta)$ emphasizes function class richness, we instead focus on how well the exploration phase covers directions of singularity to optimize the tradeoff between exploration and exploitation.

In essence, while both inequalities utilize similar language and concepts, they address different challenges: kernel ridge regression focuses on balancing bias and variance, whereas our approach targets efficient exploration and exploitation in dynamic settings. We present their comparisons in Table \ref{tab:CI}.

\begin{table}[h]
    \centering
    \begin{tabular}{c|cc}
    \hline
       & Linear Dynamic Pricing & Kernel Ridge Regression \\\hline
    Tradeoff & Exploration-exploitation tradeoff & Bias-variance tradeoff \\ 
       Dimension& Degenerate Dimension 
 & Statistical Dimension
       \\ \hline
    \end{tabular}
    \caption{The comparison of critical inequalities for linear dynamic pricing and kernel ridge regression}
    \label{tab:CI}
    \vspace{-1em}
\end{table}

	
\subsection{Minimax lower bound}
\label{sec:lb}
Thus far, we have established some regret upper bounds on the performance of our proposed LetC algorithm. To what extent are these bounds improvable? In order to answer this question, it is natural to investigate the fundamental (statistical) limitations of the dynamic pricing problem itself. Although previous studies, such as \cite{ban2021personalized}, have explored this question, the provided lower bounds do not account for the precise dependence on the context dimension. In this section, we fill this gap by deriving a precise minimax regret lower bound on this problem.

We start by 
defining the model class $\mathfrak{M}$. To facilitate fair comparison with the upper bound,  $\mathfrak{M}$ should satisfy Assumptions \ref{asp:basic-regularity}, \ref{asp:exploration}. More specifically, for some constants $b_1,b_2,
\rho_1,\rho_2$, such that $0<b_1<b_2$, $\rho_1>0$ and $0<\rho_2<1$, let 
\begin{align*}\mathfrak{M}_{b_1,b_2,
		\rho_1,\rho_2}=\Big\{\btheta=(\balpha{}^\top,\paraslp{}^\top)^\top:&-{\feat}^{\top}\paraslp,{\feat}^{\top}\balpha\in [b_1,b_2] \ \forall {\feat}\in \FeatSp;
	\\&\|\balpha\|_2^2\ge 4\rho_1,\|\paraslp\|_2^2\ge \rho_1; |\balpha^{T}\paraslp|\le \rho_2\|\balpha\|_2\|\paraslp\|_2\Big\}\end{align*} 
be the pricing model class under consideration.
The lower bound should be uniform for any measurable policy $\policy$ that maps the data collected prior to time $t$ to the price at time $t$, for any time $t\in [T]$.

We can prove the following lower bound, whose proof is delegated to Appendix  \ref{sec:proof-lower-bound}. 
The proof utilizes the multivariate van Trees inequality, as similarly applied in \cite{keskin2014dynamic,ban2021personalized}.
For notational simplicity, we introduce two shorthands, $\mathfrak{M}=\mathfrak{M}_{b_1,b_2,
\rho_1,\rho_2}$ and $\feat_{1:T}=\{\feat_t\}_{t=1}^{\Term}$  in the following stated theorem.

\begin{theorem} 
		\label{thm:pure-online-lower-bound}
Suppose there exists some $\overline{\para}\in \mathfrak{M}_{2b_1,b_2/2,
		2\rho_1,\widetilde \rho_2}$
        with $0<\widetilde \rho_2<\rho_2$. Let $\noise_i$ be i.i.d. $\calN(0,\sigma^2)$. Let $\sigma\asymp 1$ and $T>d^6$.
\begin{enumerate}
    \item For deterministic ${\feat}_1,\cdots,{\feat}_T\in \FeatSp$ such that $\|{\feat}_t\|_2\le C_f \sqrt{d},\ \forall t\in [\Term]$, we have that  \begin{align}\label{equ:lower-bound-1}\inf_{\policy}\sup_{\btheta\in \mathfrak{M}}\Exp_{\price_{1:T} \sim \policy}^{\para} \Bigg[\sum_{t=1}^{\Term} \big( \returnstar({\feat}_t;\btheta)- \return({\feat}_t,\price_t;\btheta)\big)\Biggm|{\feat_{1:\Term}}\Bigg]\gtrsim \sqrt{\Term} \, .\end{align}
        where we recall the definitions of $\returnstar({\feat}_t;\btheta)$ and $\return({\feat}_t,\price_t;\btheta)$ as in equations~\eqref{eq:revenue}~and~\eqref{eq:max-revenue}. 
\item For ${\feat}_t\sim \distr$ satisfying Assumptions \ref{asp:pure-online-second-moment-nondegenerate} and \ref{asp:exploration}, we have \begin{align}\label{equ:lower-bound-2}\inf_{\policy}\sup_{\btheta\in \mathfrak{M}}\Exp_{\price_{1:T} \sim \policy}^{\para} \big[\Reg(\Term ; \para)\big]\gtrsim \sqrt{\Term}.\end{align}
        where we recall the definition of $\Reg(\Term ; \para)$ as in equation \eqref{eq:regret_def}.
\end{enumerate}
\end{theorem}


The key takeaway from this result is that when the time horizon is sufficiently large, even an oracle algorithm knowing all the contexts beforehand cannot beat the $\sqrt{T}$ lower bound on the regret. By comparing the upper and lower bounds, we conclude that both are tight and unimprovable. Notably, while parameter estimation becomes more challenging as the dimension increases, the complexity of decision-making can remain unchanged.

	

	
\section{Numerical experiments}
	\label{sec:exp}
In this section, we conduct simulations and real data experiments to corroborate our theory, as well as test the efficacy of our proposed algorithm. The code and all numerical results are available at \href{https://github.com/jasonchaijh/LetC}{https://github.com/jasonchaijh/LetC}.

\subsection{Simulation studies}

\label{sec:sim}

A key property of our algorithm is that its regret remains independent of the dimensionality and scales proportionally to the square root of the time horizon. The primary goal of our simulation study is to demonstrate this characteristic.

We begin by generating sales data using the linear demand model in \eqref{eq:demand}, with the data-generating process defined as follows:
\begin{enumerate}
    \item We sample the context $\feat_t$ independently and identically distributed (i.i.d.) from $\distr$, where the first coordinate of $\distr$ is a constant $1$, and the other coordinates are i.i.d. $\calU[-1,1]$.
    
    \item For $d \ge 4$, we fix the ground truth coefficients $\balpha^{\star} = [1, 1/5, 1/5, 0, \ldots, 0]$ and $\bbeta^{\star} = [1, -1/5, -1/5, 0, \ldots, 0]$.
    
    \item The noises are generated from $\varepsilon_t \sim \mathcal{N}(0, 0.01^2)$.

    \item At each time $t$, given the feature $\feat_t$ and the price $p_t$ chosen by the chosen algorithm, we generate the demand as $D_t=f^{\star}(\feat_t,p_t)+\varepsilon_t$ where $f^{\star}(\feat_t,p_t)=\feat_t^{\top}\balpha^{\star}+(\feat_t^{\top}\bbeta^{\star})p_t$ is a shorthand for the mean demand.
\end{enumerate}

\begin{figure}[htbp]
    \centering
    \begin{subfigure}{0.48\textwidth}
        \centering
        \includegraphics[width=1.05\textwidth]{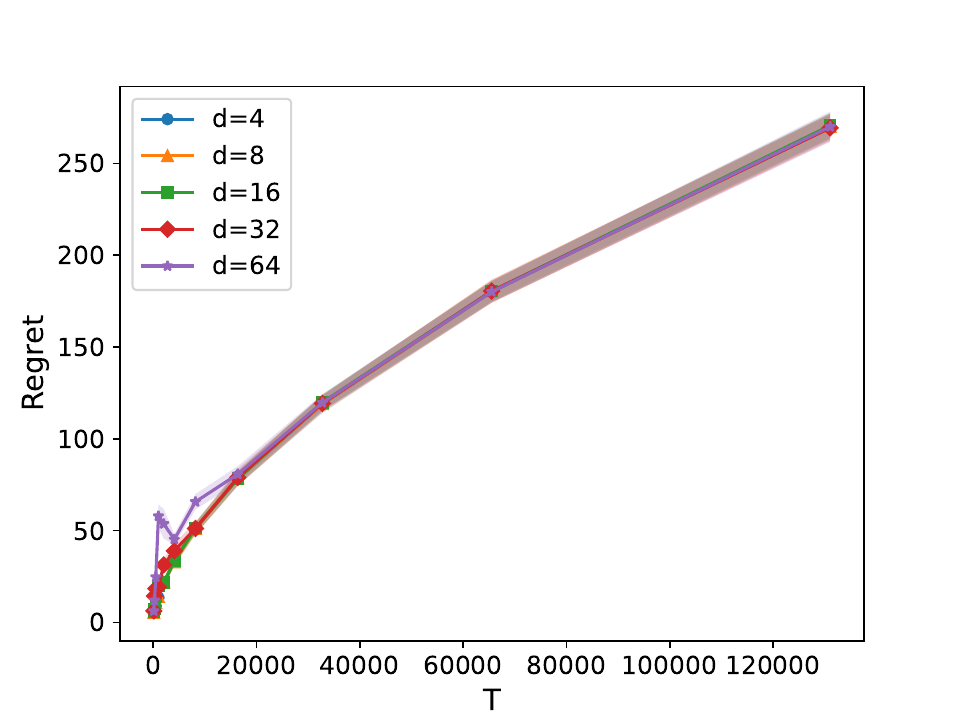}
        \caption{}
        \label{fig:sim1}
    \end{subfigure}
    \hspace{-0\textwidth}
    \begin{subfigure}{0.48\textwidth}
        \centering
        \includegraphics[width=1.05\textwidth]{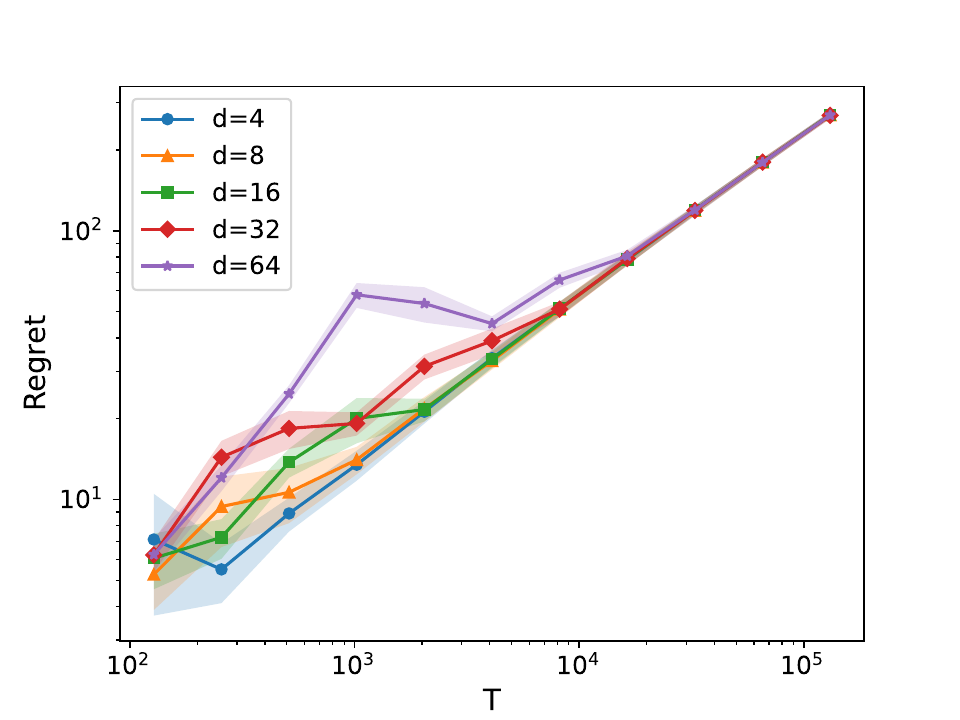}
        \caption{}\label{fig:sim2}
    \end{subfigure}
    \caption{\textbf{Regret of the LetC algorithm}. Panel (a) depicts the LetC regret versus the time $T$ in the original scale, while Panel (b) displays the LetC regret versus the time $T$ in the log-log scale. For both panels, the $x$-axis is the time $T$, and the $y$-axis is the regret up to time $T$. Different curves stand for different dimensions. The correspondence between the dimension and the color of the curve is as follows, $d=4$ is `Blue', $d=8$ is `Orange', $d=16$ is `Green', $d=32$ is `Red', $d=64$ is `Purple'. It is evident that for $T$ large enough, the regret depends on the square-root manner of the time $T$ and is independent of the dimension.
    }
    \label{fig:twopics}
\end{figure}

According to our theoretical framework, we set
\begin{align*} T_1=\bigg\lceil\frac{\sqrt{T}\log T}{C_1}\bigg\rceil, \quad T_2=\frac{T}{dC_3}, \quad and  \quad \eta=\sqrt{\frac{C_2 d\log T}{\sqrt{T}}} \end{align*}
based on the tuning hyperparameters $C_1, C_2, C_3$. Note that these hyperparameters should be scale-dependent, and we take $C_1=10, C_2=0.005, C_3=0.5$  which remain fixed throughout the experiment. For each $d \in \{4, 8, 16, 32, 64\}$ and 
$T \in \{2^7,2^8,\cdots,2^{17}\}$, we run LetC with $m = 100$ simulation trials. The resulting average regret and log-log regret plots are presented in Figure \ref{fig:sim1} and \ref{fig:sim2}, respectively. The shaded areas indicate the standard deviation across the parallel trials. 

The figures highlight two key phenomena. First, the regret of our algorithm remains largely independent of the dimension as the time horizon increases and scales in a square-root manner, which can be seen in the log-log plot that the slope tends to $1/2$. Second, consistent with our theoretical findings, there are certain thresholds above which the regret exhibits standard $\sqrt{T}$ scaling, with these thresholds being higher for larger dimensions.

\begin{figure}[!ht]
    \centering
    \includegraphics[width=0.55\textwidth]{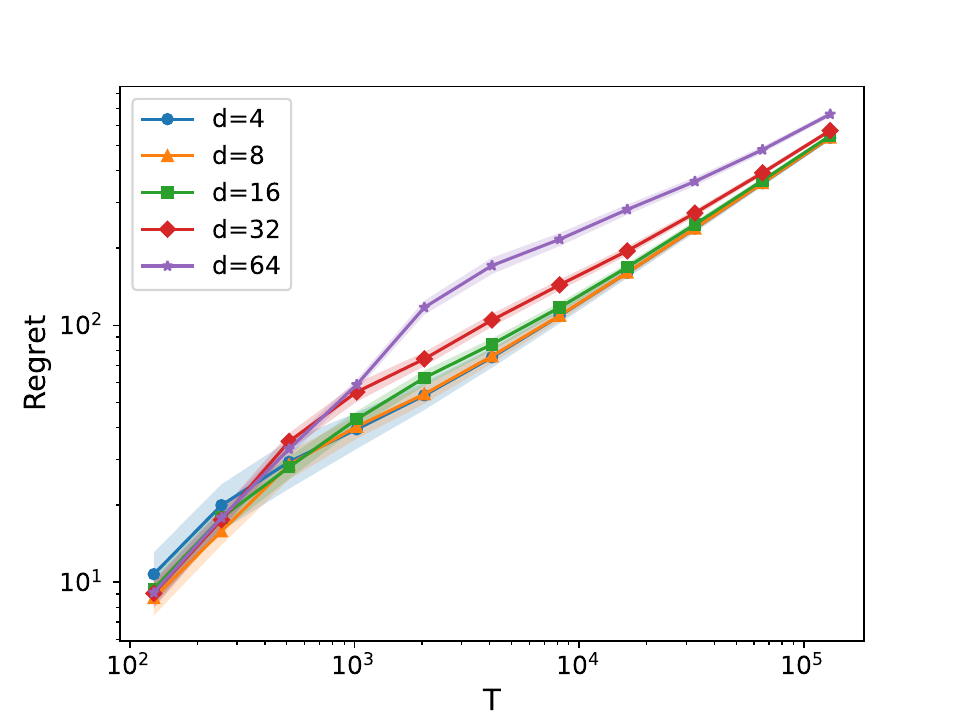}
    \caption{\textbf{Regret of LetC algorithm with doubling trick.} The figure shows the regret of the LetC algorithm with the doubling trick, plotted against time $T$ on a log-log scale. In this approach, each time segment expands by a factor of 2, and the estimation is restarted at the beginning of each new segment. The correspondence between the dimension and the color of the curve is as follows, $d=4$ is `Blue', $d=8$ is `Orange', $d=16$ is `Green', $d=32$ is `Red', $d=64$ is `Purple'.}
    \label{fig:sim3}
\end{figure}

In practice, 
the seller does not have prior knowledge of the time horizon $T$. 
To address this, we employ the doubling trick (see Section \ref{sec:main} for details) to generate a single trajectory of increasing regret over time. Figure \ref{fig:sim3} demonstrates similar behavior, showing that the regret scales in a square-root manner and is roughly independent of $d$.

	\subsection{Real-data examples}
	\label{sec:app}

We evaluate our proposed algorithm using a real-world dataset available at the \href{https://sites.google.com/view/dmdaworkshop2023/data-challenge}{INFORMS 2023 BSS Data Challenge Competition}. 
Since the ground truth model, required for generating online responses to any input algorithm, is inaccessible, we adopt a parametric bootstrap (or calibration) approach \citep{ban2021personalized,fan2022policy}, treating the fitted model as the underlying ground truth for conducting online experiments. Specifically, using the historical sales dataset, we fit a linear demand model and simultaneously estimate the distribution of features. These components are then used to generate semi-synthetic data for the testing period.

To evaluate the efficacy of our LetC algorithm, we aim to compare it with some benchmark offline pricing policy. It is reasonable to assume that the prices in the dataset result from some ``good" pricing strategy (given the monetary stakes involved) deployed by the seller, which we aim to learn from the data. Specifically, we will fit the price as a function of features using a kernel method and the resulting pricing function will be regarded as the seller's policy.  We apply these two policies (LetC and fitted pricing function) in the testing period. The details of the raw data and the testing data-generating process are elaborated below.

\paragraph{Data Description} The dataset consists of daily price and sales data for over 200 office products from Blue Summit Supplies (BSS), an e-commerce company,  covering a span of approximately seven seasons from January 1, 2022, to September 23, 2023.  Due to product heterogeneity, we employ distinct models for different products. We extracted nine features from the dataset: the first seven are indicator variables representing the days of the week to capture weekday effects. The eighth and ninth features correspond to the minimum and maximum competitor prices for the same product observed in the recent period, respectively.
It is worth noting that the allowable price bounds $[\pricelb,\priceub]$\footnote{For each product, we denote the \emph{minimum allowable price} as $\pricelb$, and the \emph{maximum allowable price} as $\priceub$, for notation consistency.} act as constraints on the realized prices but are not included in the model fitting process.

\paragraph{Semi-synthetic Data Generation} As illustrated, the data contains 9 features includes 7 binary variables and 2 continuous variables. For the binary variables that capture weekday effects, it is straightforward to generate in time order. For the continuous features, we fit a mixture of multivariate Gaussian to approximate its distribution. Specifically, for each day of the week, we fit the pair ($\pricelb_{\text{comp}},\priceub_{\text{comp}}$)\footnote{We denote the \emph{minimum price for competitors} as $\pricelb_{\text{comp}}$ and the \emph{maximum price for competitors} as $\priceub_{\text{comp}}.$} separately. Using rejection sampling, we continue generating features, accepting a feature 
$\feat$ if and only if both of the following conditions are satisfied.
\begin{enumerate}
\item $\pricelb_{\text{comp}} < \priceub_{\text{comp}}$;
\item
$f^{\star}(\feat, p) > 0$ for all $p \in [l,u]$, where we recall $f^{\star}(\feat_t,p_t)=\feat_t^{\top}\balpha^{\star}+(\feat_t^{\top}\bbeta^{\star})p_t$ is the mean demand.  
\end{enumerate}

To ensure that the realized demand remains positive, we generate the realized demand $D_t$ using the Poisson distribution: $D_t \sim \operatorname{Poisson}(f^{\star}(\feat_t, p_t))$, where we recall that $f^{\star}(\feat_t,p_t)=\feat_t^{\top}\balpha^{\star}+(\feat_t^{\top}\bbeta^{\star})p_t$ is the expected demand.

\paragraph{Offline Policy Estimation} We will demonstrate the efficacy of our policy by comparing it with a benchmark offline pricing policy, which we assume was deployed by the seller. To fit this offline pricing strategy, we use kernel ridge regression with the radial basis function (RBF) kernel on the original dataset, with the 9 constructed features.   Hyperparameters are found as regularization strength `alpha'= 0.2, kernel bandwidth `gamma'= 0.05 were selected through 5-fold cross-validation. The fitted policy $p_{\operatorname{off}}(\cdot)$ is a mapping from the features to a selling price. 

\begin{figure}[!ht]
    \centering
    \includegraphics[width=0.8\textwidth]{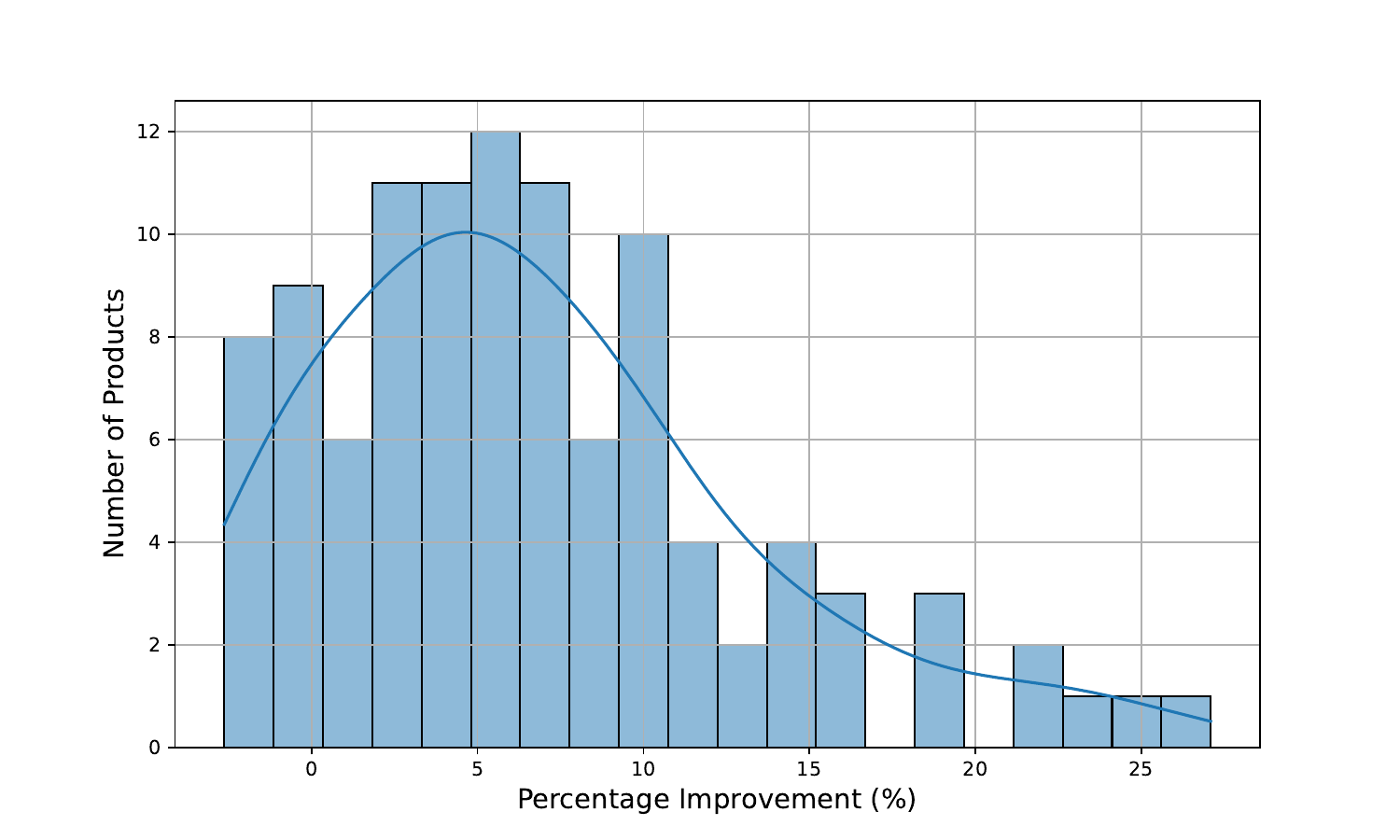}
    \caption{\textbf{Histogram of Revenue Improvement.} We plot the histogram of revenue improvement of our LetC algorithm over offline policy for a total of 105 products. 
    The revenue is calculated as \emph{expected demand $\times$ price}, aggregated over a one-year period. The revenue improvement is calculated as \mbox{$\big(\mbox{\emph{revenue}(LetC)}-\mbox{\emph{revenue}(offline)}\big) / \mbox{\emph{revenue}(offline)} \times 100\%$}. We also plot the fitted (density) curve by kernel density estimation) Compared to offline policy, our algorithm achieves more than 5\% revenue improvement in most products.
    }
    \label{fig:improvement}
\end{figure}

\paragraph{Experiment 1}
In the first experiment, we compare the offline policy with the policy chosen by our LetC algorithm in the testing period of one year (365 days), across different products. For the hyperparameters of LetC algorithm, we pick $C_1=3, C_2=0.1, C_3=0.5$. From the complete set of products, we select 105 whose demand models are suitable for linear representation. Specifically, we run the semi-synthetic feature generation process for a particular product, and if a feature is not accepted after 10,000 generations (typically due to violation of condition 2), we discard the product, as it is likely unsuitable for linear demand modeling. The histogram result in Figure \ref{fig:improvement} demonstrates significant revenue improvements when using our policy compared to the offline policy.

\begin{figure}[!ht]
    \centering
    \begin{subfigure}{0.48\textwidth}
        \centering
        \includegraphics[width=1.05\textwidth]{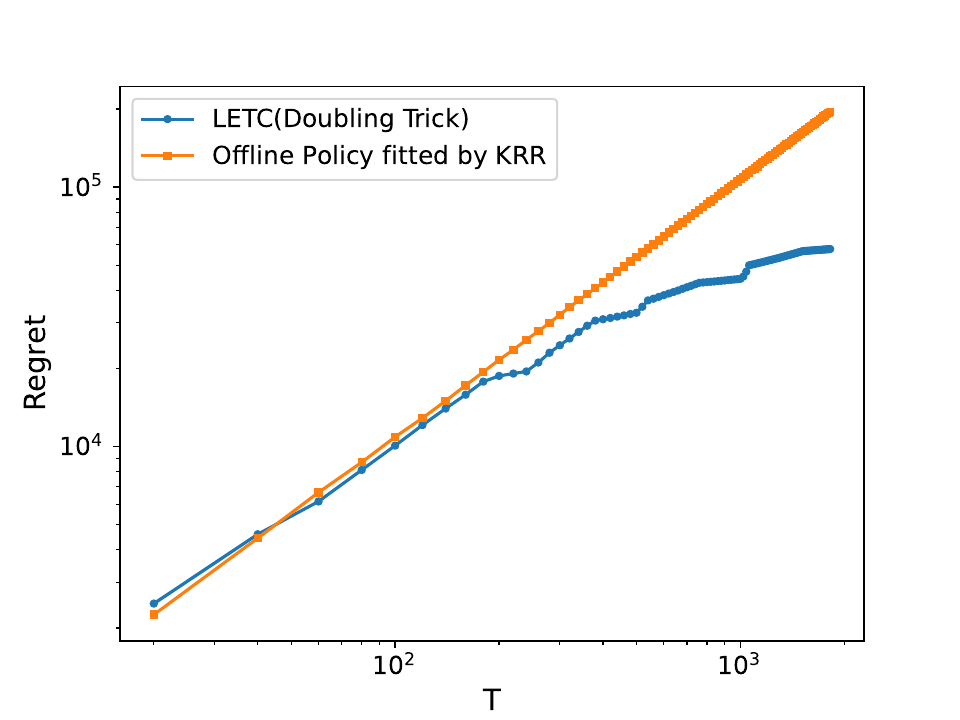}
        \caption{`Misc School Supplies SKU 17'}
        \label{fig:prod1}
    \end{subfigure}
    \hspace{0.01\textwidth}
    \begin{subfigure}{0.48\textwidth}
        \centering
        \includegraphics[width=1.05\textwidth]{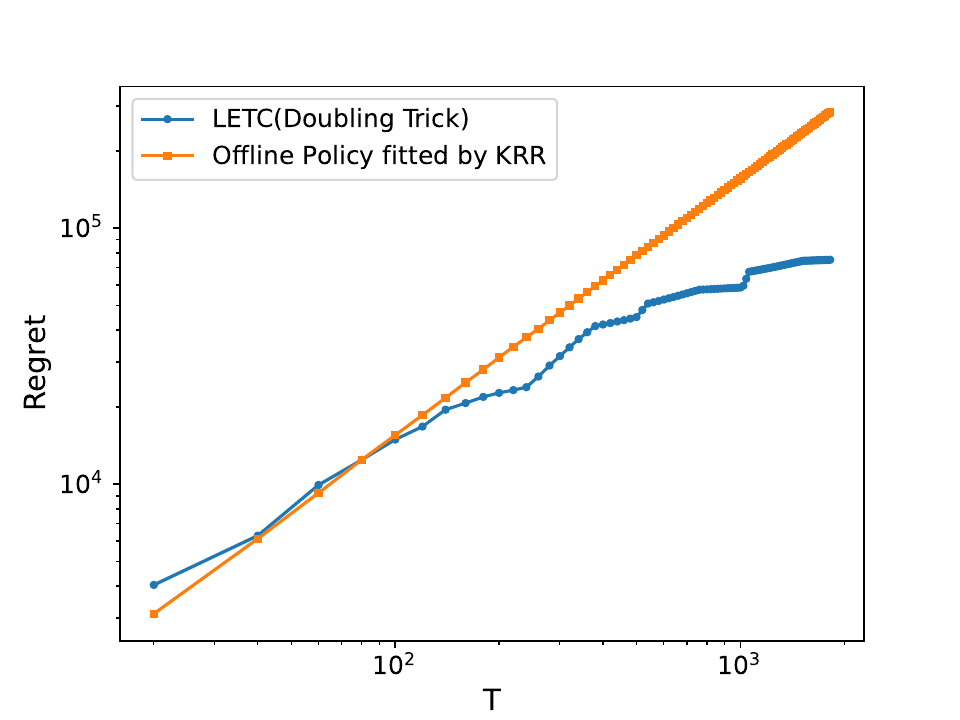}
        \caption{`Classification Folders SKU 11'}
        \label{fig:prod2}
    \end{subfigure}
    \caption{\textbf{Regret of LetC for two products in five-year period}.
    Panel (a) visualizes the regret of the product named
    `Misc School Supplies SKU 17', and Panel (b) visualizes the regret of the product named `Classification Folders SKU 11'. In both panels, the $x$-axis is the time $T$, the $y$-axis stands for the regret up to time $T$. the orange curve is for the static offline policy, the blue curve is for our LetC algorithm. We run $m=100$ trials each and plot shaded regions to represent standard error. The improvements of LetC algorithm to the static offline policy is nontrivial.}
    \label{fig:twopics}
\end{figure}

\paragraph{Experiment 2}
In the second experiment, we compare the offline policy with the policy chosen by our LetC algorithm with a doubling trick in a testing period of five years ($365\times 5$ days). We randomly pick two products, ``Misc School Supplies SKU 17'' and ``Classification Folders SKU 11'',   and draw their regret curves over time. 
The results are shown in Figure \ref{fig:prod1} and \ref{fig:prod2}. We conduct $m=100$ trials for each policy and depict the standard error as the shaded area in the plot. As we can see, our proposed policy performs almost uniformly better than the static offline policy, which incurs linear regret across testing time. Our proposed policy scales sub-linearly with time horizon, implying a decrease in average regret as more information about the pricing model is accumulated.

	\section{Conclusion}\label{sec:con}
    
    In this paper, we study the problem of contextual dynamic pricing under a linear demand model. We propose a novel three-stage online algorithm leveraging the idea of local exploration. Theoretic analysis shows the regret of the algorithm achieves a minimax lower bound when the time horizon exceeds a mild threshold polynomial in the dimension. Furthermore, our method demonstrates a smooth transition in regret behavior across the entire range of time horizons, from small to large. Additionally, we establish the critical inequality for dynamic pricing, which captures the trade-off between exploration and exploitation. Our analysis provides valuable insights into the eigenstructure of the         ``augmented'' covariance matrix. The effectiveness of the proposed algorithm is validated through both simulations and real-world data experiments, confirming its theoretical guarantees and practical utility.
    
    There are several directions worth exploring in the future. First, while we focus on the linear model for simplicity and clarity, the proposed algorithmic framework could potentially be extended to more complex settings like a generalized linear model, parametric model, or nonparametric model. Second, we believe our new critical inequality can be further generalized to other problems of statistical decision-making. And we leave these intriguing questions in future endeavor.
	
	


    \bibliographystyle{apalike2}
    {\small \bibliography{ref-main}}

	
	\newpage
	
	\appendix

\section{Proof of non-asymptotic upper bounds \yaqidone}
\label{sec:proof:thm:ub_main}

This section provides detailed proofs for the non-asymptotic upper bounds stated in \Cref{thm:ub_simple,thm:ub_main}. The analysis hinges on bounding the estimation errors for the parameters~$\para$ and $\price$ at each stage.
To begin, in \Cref{sec:proof:thm:ub_main_overview}, we present an overview of the proof. This section introduces key lemmas that quantify the estimation errors and leverages these results to derive regret bounds under both the large and limited time horizon $\Term$ regimes.
The detailed proofs of the key lemmas introduced in \Cref{sec:proof:thm:ub_main_overview} are provided in \Cref{sec:paraone_err,sec:price_stage_2,sec:paratwo_err,sec:price_stage_3}. Finally, \Cref{sec:ub_cond} includes auxiliary results and technical details that are essential for completing the proofs of \Cref{thm:ub_simple,thm:ub_main}.


\subsection{Overview \yaqidone}
\label{sec:proof:thm:ub_main_overview}

The proof of \Cref{thm:ub_simple,thm:ub_main} is divided into two main steps. First, we establish bounds on the estimation errors for the parameters $\paraone$ in Stage 1, $\paratwo$ in Stage 2, as well as for the pricing strategies $\priceone$ and $\pricetwo$ used in Stages 2 and 3. These key results are presented in \Cref{sec:key_lemma}.
Building on these bounds, we then proceed to finalize the proofs of \Cref{thm:ub_simple} and \Cref{thm:ub_main} in \Cref{sec:proof:thm:ub_simple_0} and \Cref{sec:proof:thm:ub_main_0}, respectively. In each case, we carefully choose the hyperparameters $\Termone$, $\Termtwo$, and $\perturb$ to balance the regret contributions from different stages.

\subsubsection{Analysis of each stage \yaqidone}
\label{sec:key_lemma}

In this part, we present non-asymptotic upper bounds on the estimation errors for the parameter $\paraone$ in Stage 1, the pricing strategy $\priceone$ in Stage 2, the parameter $\paratwo$ in Stage 2, and the pricing strategy $\pricetwo$ in Stage 3.

It is important to note that the results in this section hold for any hyperparameters $\Termone, \Termtwo, \perturb$ that satisfy the conditions specified in the lemmas. Although the same notations for constants (e.g., $\Const{}, \Const{1}, \const{}$) may be used across different lemmas, their values may vary depending on the context. \\

The first result \Cref{lemma:paraone_err} provides an upper bound on the estimation error for $\paraone$, obtained from the burn-in Stage~1. The proof can be found in \Cref{sec:paraone_err}.
\begin{subequations}
\begin{lemma}[Estimation error in $\paraone$ from Stage~1]
	\label{lemma:paraone_err}
	Under Assumptions~\ref{asp:pure-online-second-moment-nondegenerate}~and~\ref{asp:basic-regularity}, there are constants $\Const{}, \Const{1}$ and $\const{} > 0$ such that,
	if
	\begin{align}
		\label{eq:cond_paraone_err}
		\Termone\ge \Const{1} \cdot \max\{\Dim, \, \log \Term\},
	\end{align}
	then with probability at least $1 - \const{} \, \Term^{-2}$, it holds that
	\begin{align}
		\label{eq:paraone_err}
		\norm[\big]{\paraone-\parastar}_2^2 \; \leq \; \Const{} \cdot \frac{\Dim \, \log \Term}{\Termone} \, .
	\end{align}
\end{lemma}
\end{subequations} ~

Building on \Cref{lemma:paraone_err}, our next result \Cref{lemma:thm_proof_stage2} characterizes the deviation of the prices $\pricet{t}$ from the optimal price $\pricestar(\feat_t)$ in Stage~2. The proof is provided in \Cref{sec:price_stage_2}.
\begin{subequations}
\begin{lemma}[Pricing accuracy in Stage~2]
	\label{lemma:thm_proof_stage2}
	Under Assumptions~\ref{asp:pure-online-second-moment-nondegenerate}~and~\ref{asp:basic-regularity}, there are constants $\Const{1}, \Const{2}$ and $\const{} > 0$ such that,
	if 
	\begin{align}
		\label{eq:cond_thm_proof_stage2}
		\Termone\ge \Const{1} \cdot \Dim \, \log^2 \Term
		\qquad \mbox{and} \qquad
		\perturb \geq \Const{2} \cdot \sqrt{\Dim/ \Termone}  \, \log\Term,
	\end{align}
	then with probability at least $1 - \const{} \, \Term^{-2}$,
	\begin{align}
		\abs[\big]{\price_t - \pricestar(\feat_t)} \, \leq \; & 2 \, \perturb & & \mbox{for $\Termone < t \leq \Termone + \Termtwo$ in Stage~2}.
		\label{eq:thm_proof_stage2}
	\end{align}
\end{lemma} ~
\end{subequations}

Now it comes to our most technical \Cref{lemma:paratwo_err}, which concerns the estimation error in parameter $\paratwo$ from Stage~2. This lemma extends its simplified version presented earlier in \Cref{lemma:paratwo_err_simple}
and is key to our analysis as it shows the effect of localized exploration.
The proof of \Cref{lemma:paratwo_err} is presented in \Cref{sec:paratwo_err}.
\begin{subequations}
\begin{lemma}[Estimation error in $\paratwo$ from Stage~2, full version of \Cref{lemma:paratwo_err_simple}]
	\label{lemma:paratwo_err}
	Under the conditions of \Cref{lemma:thm_proof_stage2}, there are constants $\Const{}, \Const{1}, \Const{2}$ and $\const{} > 0$ such that, if 
	\begin{align}
		\label{eq:cond_paratwo_err}
		\Termtwo \geq \Const{1} \cdot \perturb^{-4} \max\{\Dim, \log \Term\}
		\qquad \mbox{and} \qquad
		\perturb \leq \Const{2}^{-1}, 
	\end{align}
	then with probability at least $1 - \const{} \, \Term^{-2}$,
	\begin{align}
		\label{eq:paratwo_err}
		\norm[\big]{\paratwo-\parastar}_2^2 \; \leq \; \Const{} \cdot \bigg\{ \sum_{k=1}^{2\Dim} \frac{1}{\eigpop_k + \perturb^2} \bigg\} \frac{\log \Term}{\Termtwo} \, .
	\end{align}
	Recall that $\eigpop_k$ denotes the $k$-th eigenvalue of the matrix $\Covparatworealpop$ defined in equation~\eqref{eq:Covparatworealpop}.
\end{lemma}
\end{subequations}

As shown in the bound~\eqref{eq:paratwo_err}, introducing a localized perturbation effectively increases the eigenvalues of the matrix $\Covparatworealpop$ by $\perturb^2$, which helps address any singularities in the covariance matrix: Even if some eigenvalues are nearly zero, 
adding $\perturb^2$ ensures that the bound remains stable and does not diverge. This stabilizing effect is essential for keeping the bound under control.

The role of adding $\perturb^2$ is similar to \emph{regularization in ridge regression}, where boosting the eigenvalues also stabilizes the solution.  However, unlike in regression where regularization emphasizes the most informative directions, in on-line learning, the term $\perturb^2$ helps mitigate the impact of under-explored or degenerate directions in the feature space. \\

Building on the estimation error of $\paratwo$ established in \Cref{lemma:paratwo_err}, \Cref{lemma:thm_proof_stage3} characterizes the accuracy of the prices used in Stage~3. The proof for \Cref{lemma:thm_proof_stage3} can be found in \Cref{sec:price_stage_3}.

\begin{subequations}
\begin{lemma}[Pricing accuracy in Stage~3]
	\label{lemma:thm_proof_stage3}
	Under the conditions of \Cref{lemma:paratwo_err}, there are constants $\Const{}, \Const{1}$ and $\const{} > 0$ such that, if
	\begin{align}
		\label{eq:cond_thm_proof_stage3}
		\textstyle
		\Termtwo \ge \Const{1} \cdot \big\{ \sum_k (\eigpop_k + \perturb^2)^{-1} \big\} \log^2 \Term,
	\end{align}
	then with probability at least $1 - \const{} \, \Term^{-2}$,
	\begin{align}
		\abs[\big]{\price_t - \pricestar(\feat_t)} \, \leq \; & \Const{} \cdot \sqrt{\bigg\{ \sum_{k=1}^{2\Dim} \frac{1}{\eigpop_k + \perturb^2} \bigg\} \frac{1}{\Termtwo}} \; \log\Term & & \mbox{for $\Termone + \Termtwo < t \leq \Term$ in Stage~3}.
		\label{eq:thm_proof_stage3}
	\end{align}
\end{lemma}
\end{subequations} ~


In the sequel, we will use \Cref{lemma:paraone_err,lemma:thm_proof_stage2,lemma:paratwo_err,lemma:thm_proof_stage3} to establish our \Cref{thm:ub_simple,thm:ub_main}.


\subsubsection{Proof of Theorem~\ref{thm:ub_simple} \yaqidone}
\label{sec:proof:thm:ub_simple_0}
	
	The proof of \Cref{thm:ub_simple} proceeds in three main steps.
	We first derive a general upper bound on the regret using results from \Cref{lemma:paraone_err,lemma:thm_proof_stage2,lemma:paratwo_err,lemma:thm_proof_stage3}. This bound holds for any set of hyperparameters that satisfy the conditions specified in these key lemmas.
	Next, we consider the specific choice of hyperparameters $\Termone, \Termtwo, \perturb$ as suggested by equations \eqref{eq:thm_simple_Termone}-\eqref{eq:thm_simple_perturb}. We show that, under the condition \eqref{eq:ub_simple_cond} on the planning horizon $\Term$, this choice effectively balances the regret contributions from each stage, achieving an optimal exploration-exploitation tradeoff. 
	Finally, we translate the high-probability regret bound into a bound on the expected regret, completing the proof. \\
	
	Let us dive into each step of the proof in detail.

	\paragraph{Step 1: Regret bound by combining \Cref{lemma:thm_proof_stage2,lemma:thm_proof_stage3}.}
	
	Given that \Cref{lemma:paraone_err,lemma:thm_proof_stage2,lemma:paratwo_err,lemma:thm_proof_stage3} hold, we are now ready to derive an upper bound on the regret~$\Reg(\Term)$. To begin, let us revisit the definition of cumulative regret from equation~\eqref{eq:regret_def}. By breaking it down across the three stages, we decompose the regret as follows
	\begin{align*}
		\Reg(\Term) \; = \; \bigg( \sum_{t = 1}^{\Termone} + \!\! \sum_{t = \Termone + 1}^{\Termone + \Termtwo} + \!\!\!\! \sum_{t = \Termone + \Termtwo + 1}^{\Term} \bigg) \big\{ \return_t - \returnstar(\feat_t) \big\} \, ,
	\end{align*}
	We will now address the regret contributions from each part separately.
	
	\begin{subequations}
		During Stage~1, the burn-in phase, under \Cref{asp:basic-regularity}, the single-step regret $\return_t - \returnstar(\feat_t)$ is controlled by a (uniform) upper bound on $\return_t$. Therefore, we derive
		\begin{align}
			\label{eq:regretbd_Stage1}
			\sum_{t = 1}^{\Termone} \big\{ \return_t - \returnstar(\feat_t) \big\} \; \leq \; \Const{} \cdot \Termone \, . 
		\end{align}
		
		For the remaining two stages, similar to the bound derived in \eqref{eq:regret_ub}, the regret can be expressed in terms of the difference between the implemented price and the optimal price:
		\begin{align*}
			\return_t - \returnstar(\feat_t)
			\; \leq \; (-\feat_t^{\top} \paraslpstar) \big\{ \price_t - \pricestar(\feat_t) \big\}^2
			\; \leq \; \Const{} \cdot \big\{ \price_t - \pricestar(\feat_t) \big\}^2 \, ,
		\end{align*}
		where the second inequality follows from \Cref{asp:basic-regularity}, ensuring that the slope term satisfies $(-\feat_t^{\top} \paraslpstar) \leq \featparaub$.
		Using the bounds from \Cref{lemma:thm_proof_stage2,lemma:thm_proof_stage3} for price deviations in \mbox{Stages 2 and 3}, we obtain
		\begin{align}
			\label{eq:regretbd_Stage2}
			\sum_{t = \Termone + 1}^{\Termone + \Termtwo}  \big\{ \return_t - \returnstar(\feat_t) \big\} & \; \leq \; \Const{} \cdot \perturb^2 \cdot \Termtwo \qquad \mbox{and}
			\\
			\label{eq:regretbd_Stage3}
			\sum_{t = \Termone + \Termtwo + 1}^{\Term} \!\!\!\! \big\{ \return_t - \returnstar(\feat_t) \big\}
			& \; \leq \; \Const{} \cdot \bigg\{ \sum_{k=1}^{2\Dim} \frac{1}{\eigpop_k + \perturb^2} \bigg\} \frac{1}{\Termtwo} \; \log^2\Term \cdot (\Term - \Termone - \Termtwo) \, .
		\end{align}
	\end{subequations}
	
	By combining the bounds from \eqref{eq:regretbd_Stage1}, \eqref{eq:regretbd_Stage2}, and \eqref{eq:regretbd_Stage3}, we arrive at the conclusion
	\begin{align}
		\label{eq:regretbd}
		\Reg(\Term) \; \leq \;
		\Const{} \cdot \Bigg\{ \Termone +  \perturb^2 \cdot \Termtwo + \bigg\{ \sum_{k=1}^{2\Dim} \frac{1}{\eigpop_k + \perturb^2} \bigg\} \frac{1}{\Termtwo} \; \log^2\Term \cdot \Term \Bigg\} \, .
	\end{align}

	\paragraph{Step 2: Balancing the terms by choosing proper hyperparameters.}
	
	Now, let us derive a specific form of the regret bound \eqref{eq:regretbd} by choosing the hyperparameters $\Termone$, $\Termtwo$ and $\perturb$ as suggested by equations \eqref{eq:thm_simple_Termone}-\eqref{eq:thm_simple_perturb}.
	
	First, using our choices for $\Termone$ and $\Termtwo$ from equations \eqref{eq:thm_simple_Termone} and \eqref{eq:thm_simple_Termtwo}, we can control the regret contributions for Stages~1~and~2 as follows:
	\begin{subequations}
	\begin{align}
		\label{eq:thm_simple_stage1}
		\Termone & \; \leq \; \Const{} \, \sqrt{\Term} \, \log \Term \, ,  \\
		\label{eq:thm_simple_stage2}
		\perturb^2 \cdot \Termtwo & \; \leq \; \Const{} \cdot \frac{\perturb^2 \, \Term}{\Dim}  \, .
	\end{align}
	\end{subequations}
	
	\begin{subequations}
		The analysis for Stage 3 (Commitment Phase) is a bit more involved.
		From our choice of $\perturb$ using equation \eqref{eq:thm_simple_perturb}, we get:
		\begin{align}
			\label{eq:thm_simple_perturbinv}
			\frac{1}{\perturb^2} & \; \leq \; \Const{} \cdot \frac{\sqrt{\Term}}{\Dim \, \log \Term} \, .
		\end{align}
		Additionally, the planning horizon condition \eqref{eq:ub_simple_cond} ensures
		\begin{align}
			\label{eq:thm_simple_dbd}
			\Dim & \; \leq \; \Const{} \cdot \frac{\sqrt{\Term}}{\Dim \, \log \Term} \, .
		\end{align}
		\end{subequations}
		From \Cref{lemma:exploration}, we know that the eigenvalues of the matrix $\Covparatworealpop$ satisfy $\eigpop_1 \geq \ldots \geq \eigpop_{2\Dim - 1} \geq \const{} > \eigpop_{2 \Dim} = 0$ for some constant $\const{} > 0$.
		Therefore, we can bound the sum of inverses as
		\begin{align}
			\label{eq:thm_simple_traceinvbd}
			\sum_{k=1}^{2\Dim} \frac{1}{\eigpop_k + \perturb^2}
			\; \leq \; \frac{2 \Dim - 1}{\eigpop_{2\Dim-1}} + \frac{1}{\perturb^2}
			\; \leq \; \Const{} \cdot \frac{\sqrt{\Term}}{\Dim \, \log \Term} \, ,
		\end{align}
		where the last inequality follows from the bounds~\eqref{eq:thm_simple_perturbinv}~and~\eqref{eq:thm_simple_dbd}.
		With our choice of $\Termtwo$ from equation \eqref{eq:thm_simple_Termtwo}, the regret contribution from Stage 3 simplifies to
		\begin{align}
			\label{eq:thm_simple_stage3}
			\bigg\{ \sum_{k=1}^{2\Dim} \frac{1}{\eigpop_k + \perturb^2} \bigg\} \frac{1}{\Termtwo} \; \log^2\Term \cdot \Term
			& \; \leq \; \Const{} \cdot \sqrt{\Term} \, \log \Term \, .
		\end{align}
		
		Finally, combining the bounds from Stages 1, 2, and 3 (\eqref{eq:thm_simple_stage1}, \eqref{eq:thm_simple_stage2}, and \eqref{eq:thm_simple_stage3}), we reduce the overall bound \eqref{eq:regretbd} to
		\begin{align}
			\label{eq:thm_simple}
			\Reg(\Term) \; \leq \;
			\Const{} \cdot \Big\{ \sqrt{\Term} \, \log \Term + \frac{\perturb^2 \Term}{\Dim} \Big\} \, .
		\end{align}
		This matches inequality~\eqref{eq:ub_simple_whp} from \Cref{thm:ub_simple}. By using a union bound, we conclude that this regret bound holds with probability at least $1 - \const{} \Term^{-2}$, as long as the conditions in \Cref{lemma:paraone_err,lemma:thm_proof_stage2,lemma:paratwo_err,lemma:thm_proof_stage3} are satisfied.  \\
		
		It remains to verify that our chosen hyperparameters satisfy the conditions required in \Cref{lemma:paraone_err,lemma:thm_proof_stage2,lemma:paratwo_err,lemma:thm_proof_stage3}.
		
		We begin by verifying the conditions in \Cref{lemma:paraone_err} and \Cref{lemma:thm_proof_stage2}. For the burn-in stage, we have set $\Termone \asymp \sqrt{\Term} \log \Term$ as specified in \eqref{eq:thm_simple_Termone}. The lower bound on the planning horizon~$\Term$ given in \eqref{eq:ub_simple_cond} ensures that $\Termone \gtrsim \max\{\Dim, \log \Term\}$ and $\Termone \gtrsim \Dim \log^2 \Term$, which satisfies condition~\eqref{eq:cond_paraone_err} and the first part of condition \eqref{eq:cond_thm_proof_stage2}.
		Moreover, our chosen perturbation level $\perturb$ in \eqref{eq:thm_simple_perturb} guarantees that $\perturb \gtrsim \sqrt{\Dim / \Termone} \log \Term$, fulfilling the second part of \eqref{eq:cond_thm_proof_stage2}. Therefore, the conditions required by both \Cref{lemma:paraone_err} and \Cref{lemma:thm_proof_stage2} are confirmed to be satisfied.
		
		We now proceed to verify the conditions for \Cref{lemma:paratwo_err,lemma:thm_proof_stage3}. First, let us confirm that the condition $\perturb \leq \Const{2}^{-1}$ in \eqref{eq:cond_paratwo_err} is met. This holds naturally by setting $\Const{0}’ = \Const{2}^{-1}$ in our choice of $\perturb$ as specified in \eqref{eq:thm_simple_perturb}.
		Next, we need to verify that $\Termtwo \geq \Const{1} \cdot \perturb^{-4} \max\{\Dim, \log \Term\}$, as required by \eqref{eq:cond_paratwo_err}. According to the bound derived in \eqref{eq:thm_simple_perturbinv}, we have
		\begin{align*}
			\perturb^{-4} \max\{\Dim, \log \Term\} \; \lesssim \; \Term/ (\Dim \, \log \Term) \; \lesssim \; \Term / (2\Dim) \, .
		\end{align*}
		By selecting a sufficiently large constant $\Const{0}$ in \eqref{eq:thm_simple_perturb}, we ensure that $\perturb^{-4} \max\{\Dim, \log \Term\} \leq \Term/(2\Dim) \leq \Termtwo$. Therefore, the condition \eqref{eq:cond_paratwo_err} is satisfied.
		Lastly, for condition \eqref{eq:cond_thm_proof_stage3}, we observe that
		\begin{align*}
			{ \textstyle \big\{ \sum_k (\eigpop_k + \perturb^2)^{-1} \big\} \log^2 \Term }
			\lesssim \sqrt{\Term} \, \log \Term / \Dim \; \lesssim \; \Term / (2\Dim) 
		\end{align*}
		based on \eqref{eq:thm_simple_traceinvbd}. This confirms that \eqref{eq:cond_thm_proof_stage3} is also satisfied.
		
		In summary, we have shown that our chosen hyperparameters, as specified in \eqref{eq:thm_simple_Termone}, \eqref{eq:thm_simple_Termtwo} and \eqref{eq:thm_simple_perturb}, satisfy all the conditions required by \Cref{lemma:paraone_err,lemma:thm_proof_stage2,lemma:paratwo_err,lemma:thm_proof_stage3}. Consequently, this ensures that the high-probability regret bound \eqref{eq:thm_simple} holds, thereby completing the proof of inequality \eqref{eq:ub_simple_whp} stated in \Cref{thm:ub_simple}.

\paragraph{Step 3: Converting high-probability bounds to expected value bounds.}

Above, we have shown that the regret admits an upper bound~\eqref{eq:ub_main_whp} with probability at least $1 - \Term^{-2}$. To translate this into a bound on the expected regret $\Exp[\Reg(\Term)]$, we begin by defining an event $\GoodEvent = \{ \mbox{The regret bound~\eqref{eq:ub_main_whp} holds.} \}$ and decomposing the expectation as follows
\begin{align*}
	\mathbb{E}[\Reg(\Term)] 
	& \; = \; \mathbb{E}\big[\Reg(\Term) \cdot \indicator{\GoodEvent}\big]
	+  \mathbb{E}\big[\Reg(\Term) \cdot \indicator{\GoodEventcomp}\big] \, .
\end{align*}
Here $\mathbb{E}\big[\Reg(\Term) \cdot \indicator{\GoodEvent}\big]$ can be controlled by the high-probability upper bound on the right-hand side of inequality~\eqref{eq:ub_main_whp}.
As for the other term $\mathbb{E}\big[\Reg(\Term) \cdot \indicator{\GoodEventcomp}\big]$, we notice that $\return_t - \returnstar(\feat_t)$ is always bounded above by a constant and $\Reg(\Term) \leq \Const{} \cdot \Term$ even on the complement event~$\GoodEventcomp$. 
Therefore, we have
\begin{align*}
	\mathbb{E}\big[\Reg(\Term) \cdot \indicator{\GoodEventcomp}\big] 
	\; \leq \; (\Const{} \cdot \Term) \cdot \Prob(\GoodEventcomp)
	\; \leq \; \Const{} \cdot \Term^{-1} \, ,
\end{align*}
where we have used the property that $\Prob(\GoodEventcomp) \leq \const{} \, \Term^{-2}$.
Combining these bounds, we obtain
\begin{align*}
	\mathbb{E}[\Reg(\Term)] \; \leq \; \Const{} \cdot \Big\{ \sqrt{\Term} \, \log \Term + \frac{\perturb^2 \Term}{\Dim} \Big\} + \Const{} \cdot \Term^{-1} \; \leq \; \Const{} \cdot \Big\{ \sqrt{\Term} \, \log \Term + \frac{\perturb^2 \Term}{\Dim} \Big\} \, , 
\end{align*}
which completes the derivation of the expected regret bound as stated in \Cref{thm:ub_simple}. With this, we conclude the proof of \Cref{thm:ub_simple}.


\subsubsection{Proof of Theorem~\ref{thm:ub_main} \yaqidone}
\label{sec:proof:thm:ub_main_0}

The proof of \Cref{thm:ub_main} follows the same structure as the proof of \Cref{thm:ub_simple} in \Cref{sec:ub_simple}, with the key difference being in Step 2: balancing the terms by selecting appropriate hyperparameters. 
In this section, we explicitly calculate each term in the regret bound~\eqref{eq:regretbd} using our chosen values for the hyperparameters $\perturb$, $\Termone$, and $\Termtwo$, as defined in equations~\eqref{eq:thm_main_perturb}, \eqref{eq:thm_main_Termone} and \eqref{eq:thm_main_Termtwo}. This will allow us to derive a final bound in the form of inequality~\eqref{eq:ub_main_whp}.
The detailed verification that these hyperparameter choices satisfy the conditions required by \Cref{lemma:paraone_err,lemma:thm_proof_stage2,lemma:paratwo_err,lemma:thm_proof_stage3} is provided in \Cref{sec:verification}. \\

We start by analyzing the contributions to the regret from the first two exploration stages in the expression~\eqref{eq:regretbd}. By applying our chosen parameters $\Termone$ and $\Termtwo$ as defined in equations \eqref{eq:thm_main_Termone} and \eqref{eq:thm_main_Termtwo}, we obtain
\begin{subequations}
	\begin{align}
		\label{eq:thm_main_stage1}
		\Termone & \; \leq \; \Const{} \, \sqrt{\EffDim \, \Term} \, \log \Term \, ,  \\
		\label{eq:thm_main_stage2}
		\perturb^2 \cdot \Termtwo & \; \leq \; \Const{} \cdot (\EffDim/\Dim) \; \perturb^2 \, \Term \, .
	\end{align}
\end{subequations}

We now focus on the regret incurred during the exploitation phase (Stage~3).
First, using the definition of the degenerate dimension $\EffDim(\perturb)$ from equation \eqref{eq:EffDim}, we have
\begin{align}
	\label{eq:thm_main_aux0}
	\sum_{k=1}^{2\Dim} \frac{1}{\eigpop_k + \perturb^2} 
	\leq \frac{1}{\perturb^2}
	\sum_{i=1}^{2\Dim} \min \Big\{ \frac{\perturb^2}{\eigpop_i}, 1 \Big\} 
	= \frac{\EffDim(\perturb)}{\perturb^2} \, .
\end{align}
Recall that our choice of the noise level $\perturb$ satisfies the critical inequality~\ref{eq:CI}, which implies
\begin{align*}
	\frac{1}{\perturb^2}
	\; \leq \; \SNR \cdot \Singular(\perturb)
	\; = \; \frac{1}{2 \ConstCI} \, \frac{\sqrt{\Term}}{\log\Term} \; \frac{\sqrt{\EffDim(\perturb)}}{\Dim} \, .
\end{align*}
Given that our parameter choice ensures $\EffDim \geq \EffDim(\perturb)$, we can simplify this to
\begin{align}
	\label{eq:thm_main_aux1}
	\frac{1}{\perturb^2}
	\; \leq \; \Const{} \cdot \frac{\sqrt{\Term}}{\log\Term} \; \frac{\sqrt{\EffDim}}{\Dim} \, .
\end{align}

We then use inequalities \eqref{eq:thm_main_aux0} and \eqref{eq:thm_main_aux1} to bound the regret term from Stage~3 in the overall regret expression \eqref{eq:regretbd}.
Using \eqref{eq:thm_main_aux0} along with the definition of $\Termtwo$ in \eqref{eq:thm_main_Termtwo}, we obtain
\begin{align*}
	\bigg\{ \sum_{k=1}^{2\Dim} \frac{1}{\eigpop_k + \perturb^2} \bigg\} \frac{1}{\Termtwo} \; \log^2\Term \cdot \Term 
	& \; \leq \; \Const{} \cdot \frac{\EffDim(\perturb)}{\perturb^2} \cdot \frac{\Dim}{\EffDim} \; \log^2 \Term
	\; \leq \; 
	 \Const{} \cdot \frac{\Dim}{\perturb^2} \, \log^2 \Term \, .
\end{align*}
Substituting inequality \eqref{eq:thm_main_aux1}, we further simplify this to
\begin{align}
	\label{eq:thm_main_stage3}
	\bigg\{ \sum_{k=1}^{2\Dim} \frac{1}{\eigpop_k + \perturb^2} \bigg\} \frac{1}{\Termtwo} \; \log^2\Term \cdot \Term 
	& \; \leq \; \Const{} \, \sqrt{\EffDim \, \Term} \, \log \Term \, .
\end{align}

By using the upper bounds derived in equations \eqref{eq:thm_main_stage1}, \eqref{eq:thm_main_stage2} and \eqref{eq:thm_main_stage3}, we can express the overall regret as
\begin{align*}
	\Reg(\Term) \; \leq \; \Const{} \, \Big\{ \sqrt{\EffDim \, \Term} \, \log \Term + (\EffDim/\Dim) \; \perturb^2 \, \Term \Big\} \, ,
\end{align*}
which corresponds to the bound stated in inequality~\eqref{eq:ub_main_whp} in \Cref{thm:ub_main}. This holds with probability at least $1 - \const{} \, \Term^{-2}$ by union bound.
The bound on the expected regret can be derived in a similar manner by following the argument used in Step~3 of \Cref{sec:proof:thm:ub_simple_0}. This completes the proof of \Cref{thm:ub_main}.  \\

In what follows, we will prove the claims \eqref{eq:ub_main_regular} and \eqref{eq:ub_main} mentioned in the comments immediately after \Cref{thm:ub_main}.

\paragraph{Proof of claim~\eqref{eq:ub_main_regular}:}
When $\perturb$ is a regular solution to the critical inequality \ref{eq:CI}, by definition, there exists a constant $\zeta > 1$ such that $\perturb / \zeta$ no longer satisfies \ref{eq:CI}. This implies
\begin{align}
	\label{eq:thm_main_aux2}
	\Singular(\perturb / \zeta) = \sqrt{\frac{1}{2 \Dim} \sum_{k=1}^{2 \Dim}  \min \Big\{ \frac{(\perturb/\zeta)^2}{\eigpop_k}, 1 \Big\}}
	\; < \; \ConstCI \, \sqrt{\frac{2\Dim}{\Term}} \, \log \Term \, \cdot \, \frac{1}{(\perturb/\zeta)^2} \, .
\end{align}
We also observe that
\begin{align}
	\label{eq:thm_main_aux3}
	\Singular(\perturb / \zeta) 
	\; = \sqrt{\frac{1}{2 \Dim} \sum_{k=1}^{2 \Dim}  \min \Big\{ \frac{(\perturb/\zeta)^2}{\eigpop_k}, 1 \Big\}}
	\; \geq \sqrt{\frac{1}{2 \Dim} \sum_{k=1}^{2 \Dim}  \min \Big\{ \frac{(\perturb/\zeta)^2}{\eigpop_k}, (1/\zeta)^2 \Big\}}
	\; = \; \Singular(\perturb) / \zeta \, .
\end{align}
By combining inequalities \eqref{eq:thm_main_aux2} and \eqref{eq:thm_main_aux3}, we obtain
\begin{align}
	\label{eq:thm_main_aux4}
	\perturb^2 \; \leq \; \Const{} \, \sqrt{\frac{2\Dim}{\Term}} \, \log \Term \cdot \Singular(\perturb / \zeta)^{-1}
	\; \leq \; \Const{} \, \sqrt{\frac{2\Dim}{\Term}} \, \log \Term \cdot \Singular(\perturb)^{-1}
	\; \leq \; \Const{} \cdot \frac{\Dim  \, \log \Term}{\sqrt{\EffDim(\perturb) \, \Term}} \, .
\end{align}
By plugging this upper bound for $\perturb^2$ from \eqref{eq:thm_main_aux4} into the regret expression in \eqref{eq:ub_main_whp}, we derive the refined regret bound as claimed in \eqref{eq:ub_main_regular}.

\paragraph{Proof of claim~\eqref{eq:ub_main}:}
The derivation of the bound \eqref{eq:ub_main} from the bound \eqref{eq:ub_main_regular} is straightforward. If we assume $\EffDim \asymp \EffDim(\perturb)$, then the second term in \eqref{eq:ub_main_regular} simplifies, leading to
\begin{align*}
\Reg(\Term) \; \leq \; \Const{} \, \sqrt{\EffDim \, \Term} \, \log \Term \, ,
\end{align*}
as stated in \eqref{eq:ub_main}.


\subsection{Proof of Lemma~\ref{lemma:paraone_err}, estimation error in $\paraone$ from Stage~1 \yaqidone}
\label{sec:paraone_err}


In Stage 1, we estimate the parameters \mbox{$\parastar = (\paraintstar, \paraslpstar)$} using least-squares regression. Specifically, we compute an estimate $\paraone = (\paraintone, \paraslpone)$ based on the dataset $\Data_{1 : \Termone} = \{ (\feat_t, \price_t, \Demand_t) \}_{t=1}^{\Termone}$.

According to standard results from fixed-design regression (see \Cref{lemma:ls} in \Cref{sec:ls}), with high probability $1 - \Term^{-2}$, we can bound the error between our estimate $\paraone$ and the true parameter~$\parastar$ as follows:
\begin{align}
	\label{eq:paraone_err1}
	\distrnorm[\big]{\paraone - \parastar}{2}^2
	\; \leq \; \Const{} \cdot \frac{\trace(\Covparaone^{-1})}{\Termone} \, \log \Term \, ,
\end{align}
where the matrix $\Covparaone$ is the empirical covariance matrix of the features and prices, given by
\begin{align*}
	\Covparaone \; \defn \; 
	\frac{1}{\Termone} \sum_{t = 1}^{\Termone}
	\begin{bmatrix}
		\feat_t \\ \feat_t  \, \price_t
	\end{bmatrix}
	\begin{bmatrix}
		\feat_t \\ \feat_t  \, \price_t
	\end{bmatrix}^{\top} \, .
\end{align*}

It is then important to ensure that matrix $\Covparaone$ is well-conditioned. In particular, we can show that, if $\Termone \geq \Const{1} \cdot \max\{ \Dim, \log \Term \}$, then the inverse of $\Covparaone$ satisfies
\begin{align}
	\label{eq:trace_Covparaone}
	\trace(\Covparaone^{-1}) \; \leq \; \Const{} \cdot \Dim
	\qquad
	\mbox{with probability at least $1 - \Term^{-2}$}.
\end{align}
Applying a union bound, we conclude from inequalities~\eqref{eq:paraone_err1}~and~\eqref{eq:trace_Covparaone} that
\begin{align}
	\label{eq:paraone_err2}
	\distrnorm[\big]{\paraone - \parastar}{2}^2
	\; \leq \; \Const{} \cdot \frac{\Dim}{\Termone} \, \log \Term \, ,
\end{align}
with probability at least $1 - 2 \, \Term^{-2}$.

This result~\eqref{eq:paraone_err2} completes the proof of \Cref{lemma:paraone_err}.
The remaining task is to prove the condition on the covariance matrix in inequality~\eqref{eq:trace_Covparaone}, which we will address next.

\subsubsection{Proof of inequality~\eqref{eq:trace_Covparaone}, trace of inverse covariance matrix \yaqidone} 

We define the population counterpart of the matrix $\Covparaone$ as
\begin{align*}
	\Covparaonerealpop \defn \frac{1}{2} \, \bigg\{ 
	\begin{bmatrix}
		\Covfeat & \pricelb \, \Covfeat \\
		\pricelb \, \Covfeat & \pricelb^2 \, \Covfeat
	\end{bmatrix}
	+
	\begin{bmatrix}
		\Covfeat & \priceub \, \Covfeat \\
		\priceub \, \Covfeat & \priceub^2 \, \Covfeat
	\end{bmatrix}
	\bigg\}
	\; \in \Real^{2\Dim \times 2\Dim} \, .
\end{align*}
This matrix represents the expected value of the random matrix $\Covparaone$,
i.e. $\Exp [ \Covparaone ] = \Covparaonerealpop$.
\begin{subequations}
	
	Our analysis relies on two key results:
	\begin{align}
		\label{eq:eigmin_Covparaonerealpop}
		& \eigmin(\Covparaonerealpop) \; \geq \; \Const{}^{-1}
		\qquad \qquad \mbox{and} \\
		\label{eq:diff_Covparaonerealpop}
		& \Prob \big( \norm{\Covparaone - \Covparaonerealpop}_2
		\leq \eigmin(\Covparaonerealpop) / 2 \, \big) \; \geq \; 1 - \Term^{-2}
		\qquad \mbox{if $\Termone \geq \Const{1} \cdot \max\{ \Dim, \log \Term \}$}.
	\end{align}
\end{subequations}
The bound in \eqref{eq:eigmin_Covparaonerealpop} comes from our exploration scheme, which considers two extreme price points. Inequality \eqref{eq:diff_Covparaonerealpop} is established using a concentration bound for sub-Gaussian random vectors.

By combining \eqref{eq:eigmin_Covparaonerealpop} and \eqref{eq:diff_Covparaonerealpop}, we see that with high probability,
\begin{align}
	\label{eq:diff_Covparaonerealpop_cor}
	\eigmin(\Covparaone) \geq \eigmin(\Covparaonerealpop) - \norm{\Covparaone - \Covparaonerealpop}_2 \geq \eigmin(\Covparaonerealpop) / 2 \geq \Const{}^{-1}.
\end{align}
Therefore, we can conclude
\begin{align*}
	\trace(\Covparaone^{-1}) \leq \eigmin^{-1}(\Covparaone) \cdot \Dim \leq \Const{} \cdot \Dim,
\end{align*}
which validates inequality~\eqref{eq:trace_Covparaone}. 

We now turn to proving inequalities \eqref{eq:eigmin_Covparaonerealpop} and \eqref{eq:diff_Covparaonerealpop}, which we will handle separately.
\\

\paragraph{Proof of inequality~\eqref{eq:eigmin_Covparaonerealpop}:}
We first show that the population covariance matrix $\Covparaonerealpop$ is well-conditioned, i.e., $\eigmin(\Covparaonerealpop) \geq \Const{}^{-1}$.

We begin by reformulating the matrix $\Covparaonerealpop$ as
\begin{align*}
	\Covparaonerealpop =
	\frac{1}{2}
	\begin{bmatrix}
		\IdMt  &  \IdMt  \\
		\pricelb \, \IdMt & \priceub \, \IdMt
	\end{bmatrix}
	\begin{bmatrix}
		\Covfeat &  \\
		& \Covfeat
	\end{bmatrix}
	\begin{bmatrix}
		\IdMt  &  \pricelb \, \IdMt  \\
		\IdMt & \priceub \, \IdMt
	\end{bmatrix} \, .
\end{align*}
Using the properties of the operator norm $\norm{ \, \cdot \, }_2$, we bound $\norm{\Covparaonerealpop^{-1}}_2$ as
\begin{align*}
	\norm[\big]{\Covparaonerealpop^{-1}}_2
	\; \leq \; 	\frac{1}{2} \,
	\norm[\bigg]{\begin{bmatrix}
			\IdMt  &  \IdMt  \\
			\pricelb \, \IdMt & \priceub \, \IdMt
		\end{bmatrix}^{-1}}_2^2 \;
	\norm[\bigg]{	\begin{bmatrix}
			\Covfeat^{-1} &  \\
			& \Covfeat^{-1}
	\end{bmatrix}}_2
	= \frac{1}{2} \,
	\norm[\bigg]{\begin{bmatrix}
			1  &  1  \\
			\pricelb  & \priceub
		\end{bmatrix}^{-1}}_2^2 \;
	\norm[\big]{\Covfeat^{-1}}_2 \, .
\end{align*}
Next, we compute
\begin{align*}
	\norm[\bigg]{\begin{bmatrix}
			1  &  1  \\
			\pricelb  & \priceub
		\end{bmatrix}^{-1}}_2
	= \norm[\bigg]{
		\frac{1}{\priceub - \pricelb}
		\begin{bmatrix}
			\priceub  &  - 1  \\
			- \pricelb  & 1
	\end{bmatrix}}_2
	\leq \frac{1}{\priceub - \pricelb} \,
	\norm[\bigg]{\begin{bmatrix}
			\priceub  &  - 1  \\
			- \pricelb  & 1
	\end{bmatrix}}_{\ell_1}
	= \frac{2 + \pricelb + \priceub}{\priceub - \pricelb} \, .
\end{align*}
By combining these bounds, we obtain
\begin{align*}
	\norm[\big]{\Covparaonerealpop^{-1}}_2
	\; \leq \;
	2 \, \Big\{ 1 + \frac{\pricelb + \priceub}{2} \Big\}^2
	\frac{\norm{\Covfeat^{-1}}_2}{(\priceub - \pricelb)^2} \, .
\end{align*}
This ensures that
\begin{align}
	\label{eq:eigmin_Covparaonerealpop2}
	\eigmin(\Covparaonerealpop)
	\; = \; \norm[\big]{\Covparaonerealpop^{-1}}_2^{-1}
	\; \geq \; \frac{1}{2} \, \Big\{ \frac{\priceub - \pricelb}{1 + (\pricelb + \priceub)/2} \Big\}^2 \cdot \eigmin(\Covfeat) \, .
\end{align}
\Cref{asp:basic-regularity} guarantees that the right-hand side of inequality \eqref{eq:eigmin_Covparaonerealpop2} is bounded below by a positive constant. Therefore, the property in inequality~\eqref{eq:eigmin_Covparaonerealpop} holds.

\paragraph{Proof of inequality~\eqref{eq:diff_Covparaonerealpop}:}
Next, we show that the empirical covariance matrix $\Covparaone$ is concentrated around its population counterpart $\Covparaonerealpop$. A key observation is as follows: since contexts~$\{ \feat_t \}_{t=1}^{\Termone}$ are i.i.d. sub-Gaussian with variance proxy $\stdfeat^2$ (as per \Cref{asp:pure-online-second-moment-nondegenerate}) and the prices $\{ \price_t \}_{t=1}^{\Termone}$ are uniformly bounded by $\priceub > 0$ (by \Cref{asp:basic-regularity}), the combined context-price vector $\{ ( \feat_t, \feat_t \, \price_t ) \}_{t = 1}^{\Termone} \subseteq \Real^{2 \Dim}$ in Stage~1 are also i.i.d. sub-Gaussian. The variance proxy is $(1 + \priceub^2) \stdfeat^2$, which is of constant order.

Now, applying Theorem~6.5 from \cite{wainwright2019high} (see also inequality~(5.25) in \cite{vershynin2010introduction}), we find that with probability at least $1 - \Term^{-2}$:
\begin{align}
	\label{eq:diff_Covparaonerealpop_orig}
	\norm[\big]{\Covparaone - \Covparaonerealpop}_2 
	\; \leq \; \Const{} \cdot \Big\{ \sqrt{\frac{\Dim + \log \Term}{\Termone}} + \frac{\Dim + \log \Term}{\Termone} \Big\} \, .
\end{align}
To ensure that $\norm[\big]{\Covparaone - \Covparaonerealpop}_2 \leq \eigmin(\Covparaonerealpop) / 2$, it is sufficient to have a large enough sample size~$\Termone$, such that $\Termone \geq \Const{1} \cdot \max\{ \Dim, \log \Term \}$.


\subsection{Proof of Lemma~\ref{lemma:thm_proof_stage2}, pricing accuracy in Stage~2 \yaqidone}
\label{sec:price_stage_2}

To begin with, recall that the price during localized exploration is set as $\price_t = \clip{\priceone(\feat_t) + \perturb \cdot \Rad_t}$ with $\Rad_t \in \{ \pm 1 \}$. We claim that, to prove inequality \eqref{eq:thm_proof_stage2}, it suffices to show that $2\eta\le\bar\delta$ and
\begin{align}
    \abs[\big]{\priceone(\feat_t) - \pricestar(\feat_t)} \, \leq \; & \perturb & & \mbox{for $\Termone < t \leq \Termone + \Termtwo$ in Stage~2}.
	\label{eq:thm_proof_stage2_alter}
\end{align}

Specifically, if $2\eta\le\bar\delta$ and \eqref{eq:thm_proof_stage2_alter} holds, then $\abs{\priceone(\feat_t) + \perturb \cdot \Rad_t-\pricestar(\feat_t)}\le \abs[\big]{\priceone(\feat_t) - \pricestar(\feat_t)} + \perturb\le \bar\delta$, 
this together with $\pricestar(\feat_t)\in [\ell-\bar\delta,r+\bar\delta]$ in Assumption \ref{asp:basic-regularity}.1 leads to
$\priceone(\feat_t) + \perturb \cdot \Rad_t\in [\ell,u]$, hence $\price_t=\priceone(\feat_t) + \perturb \cdot \Rad_t$ and \eqref{eq:thm_proof_stage2}. As $\eta\le C_0'$, the condition $2\eta\le \bar\delta$ is ensured by setting $C_0'\le \frac{\bar\delta}{2}$. In what follows, we aim to prove \eqref{eq:thm_proof_stage2_alter}.

Using the expressions for $\priceone$ and $\pricestar$ from equations \eqref{eq:priceone} and \eqref{eq:opt_price}, we have
\begin{align}
	& \phantom{\; \leq \;} \abs[\big]{\priceone(\feat_t) - \pricestar(\feat_t)} \; \leq \; \abs[\bigg]{ \frac{\feat_t^{\top} \paraintone}{2 \, \feat_t^{\top} \paraslpone} - \frac{\feat_t^{\top} \paraintstar}{2 \, \feat_t^{\top} \paraslpstar} }  \notag \\
	& 
	\; \leq \; \frac{1}{2 \, \big\{ \abs{\feat_t^{\top} \paraslpstar} - \abs{\feat_t^{\top} (\paraslpone - \paraslpstar)} \big\}} \,
	\bigg\{ \abs[\big]{ \feat_t^{\top} (\paraintone - \paraintstar) }
	+ \frac{\abs{\feat_t^{\top} \paraintstar}}{\abs{\feat_t^{\top} \paraslpstar}} \cdot \abs[\big]{ \feat_t^{\top} (\paraslpone - \paraslpstar) } \bigg\} \, .
	\label{eq:price_diff_stage2}
\end{align}
By applying \Cref{asp:basic-regularity}, which states that $\abs{\feat_t^{\top} \paraintstar}, \abs{\feat_t^{\top} \paraslpstar} \in [\featparalb, \featparaub]$ with $0 < \featparalb \leq \featparaub$, we find that, given the condition
\begin{align}
	\label{eq:cond_featparaslpone}
	\abs[\big]{\feat_t^{\top} (\paraslpone - \paraslpstar)} \; \leq \; \frac{\featparalb}{2} \, ,
\end{align}
the bound~\eqref{eq:price_diff_stage2} reduces to
\begin{align}
	\label{eq:diff_price_stage2}
	\abs[\big]{\priceone(\feat_t) - \pricestar(\feat_t)} 
	\; \leq \; \Const{} \cdot \Big\{ \abs[\big]{ \feat_t^{\top} (\paraintone - \paraintstar) }
	+ \abs[\big]{ \feat_t^{\top} (\paraslpone - \paraslpstar) } \Big\} \, .
\end{align}

Now, we claim that the following bounds hold for $t = \Termone + 1, \ldots, \Termone + \Termtwo$ with probability at least $1 - \const{} \, \Term^{-2}$:
\begin{subequations}
	\begin{align}
		\label{eq:featparaintone}
		\abs[\big]{ \feat_t^{\top} (\paraintone - \paraintstar) }
		& \; \leq \; \Const{} \cdot \sqrt{\frac{\Dim}{\Termone}} \, \log \Term \, ,  \\
		\label{eq:featparaslpone}
		\abs[\big]{ \feat_t^{\top} (\paraslpone - \paraslpstar) }
		& \; \leq \; \Const{} \cdot \sqrt{\frac{\Dim}{\Termone}} \, \log \Term \, .
	\end{align}
\end{subequations}
Combining~the bounds~\eqref{eq:diff_price_stage2}, \eqref{eq:featparaintone}~and~\eqref{eq:featparaslpone}, we derive that there exists a constant $\Const{2} > 0$ such that
\begin{align*}
	\abs[\big]{\priceone(\feat_t) - \pricestar(\feat_t)} 
	\; \leq \; \Const{2} \cdot \sqrt{\frac{\Dim}{\Termone}} \, \log \Term \, .
\end{align*}
Therefore, by choosing $\perturb \geq \Const{2} \cdot \sqrt{\Dim/ \Termone}  \, \log\Term $, we ensure that $\abs[\big]{\priceone(\feat_t) - \pricestar(\feat_t)} \leq \perturb$ for all $\Termone < t \leq \Termone + \Termtwo$ in Stage~2, which verifies inequality~\eqref{eq:thm_proof_stage2_alter}.

Finally, to satisfy the condition \eqref{eq:cond_featparaslpone}, we need $\Termone \geq \Const{1} \cdot \Dim \, \log^2 \Term$.
This completes the proof of \Cref{lemma:thm_proof_stage2}.

In the next step, we will prove the inequalities \eqref{eq:featparaintone} and \eqref{eq:featparaslpone} based on the error bound for $\paraone$ given in \Cref{lemma:paraone_err}.  \\


\paragraph{Proof of inequalities~\eqref{eq:featparaintone}~and~\eqref{eq:featparaslpone}:}
We will focus on proving inequality \eqref{eq:featparaintone}, as the proof for inequality \eqref{eq:featparaslpone} is similar.

For $\Termone + 1 \leq t \leq \Termone + \Termtwo$, note that under the assumption that $\feat_t$ is sub-Gaussian, and due to the independence between $\feat_t$ and $\paraintone - \paraintstar$, the term $\feat_t^{\top}(\paraintone - \paraintstar)$ is also sub-Gaussian, conditional on $\paraintone$. Its variance proxy is $\stdfeat^2 \, \norm{\paraintone - \paraintstar}_2^2$. This allows us to apply concentration results for sub-Gaussian random variables.

Specifically, for any single $t$, with some sufficiently large constant $\Const{} > 0$ and with probability at least $1 - \Term^{-3}$, the following holds:
\begin{align}
	\label{equ:theorem1-sub-gaussian0}
	\abs[\big]{{\feat}_t^{\top}(\paraintone - \paraintstar)} \; \leq \; \Const{} \, \cdot \sqrt{\log \Term} \, \norm{\paraintone - \paraintstar}_2 \, .
\end{align}
We use a union bound to extend this result over all $t$ in the range $\Termone + 1 \leq t \leq \Termone + \Termtwo$. By doing so, we conclude that with probability at least $1 - \Term^{-2}$, the bound in \eqref{equ:theorem1-sub-gaussian0} holds for all~$t$ in this range.

Now, we apply \Cref{lemma:paraone_err}, which gives us a bound on $\norm{\paraintone - \paraintstar}_2$:
\begin{align}
	\label{equ:theorem1-sub-gaussian1}
	\norm{\paraintone - \paraintstar}_2
	\leq \norm[\big]{\paraone - \parastar}_2 \leq \Const{} \cdot \sqrt{\frac{\Dim \log \Term}{\Termone}} \, ,
\end{align}
with probability at least $1 - \Term^{-2}$.

Finally, combining inequality \eqref{equ:theorem1-sub-gaussian0} with the bound~\eqref{equ:theorem1-sub-gaussian1}, we obtain the desired result, completing the proof of inequality \eqref{eq:featparaintone}.


\subsection{Proof of Lemma~\ref{lemma:paratwo_err}, estimation error in $\paratwo$ from Stage~2 \yaqidone} 
\label{sec:paratwo_err}


In this section, we aim to analyze the localized exploration stage and control the estimation error of the parameter $\paratwo$. 

Our parameter $\paratwo$ is estimated via linear regression using the dataset $\Data_{\Termone + 1 : \Termone + \Termtwo}$.
By applying \Cref{lemma:ls}, we obtain that with probability at least $1 - \Term^{-2}$,
\begin{align}
	\label{eq:paratwo_err1}
	\distrnorm[\big]{\paratwo - \parastar}{2}^2
	\; \leq \; \Const{} \cdot \frac{\trace(\Covparatwo^{-1})}{\Termtwo} \, \log \Term \, ,
\end{align}
where the empirical covariance matrix $\Covparatwo \in \Real^{2\Dim}$ is defined
\begin{align}
	\label{eq:Covparatwo}
	\Covparatwo \; \defn \; 
	\frac{1}{\Termtwo} \sum_{t = \Termone + 1}^{\Termone + \Termtwo}
	\begin{bmatrix}
		\feat_t \\ \feat_t  \, \price_t
	\end{bmatrix}
	\begin{bmatrix}
		\feat_t \\ \feat_t  \, \price_t
	\end{bmatrix}^{\top} \, .
\end{align}

The core of our analysis shows that under certain conditions, the trace of the inverse covariance matrix $\trace(\Covparatwo^{-1})$ satisfies the following bound
\begin{align}
	\label{eq:S2}
	\trace(\Covparatwo^{-1}) \; \leq \; \Const{} \cdot \sum_{k=1}^{2\Dim} \frac{1}{\eigpop_k + \perturb^2}
	\qquad \mbox{with probability at least $1 - \const{} \, \Term^{-2}$.}
\end{align}
Here $\eigpop_1 \geq \eigpop_2 \geq \ldots \geq \eigpop_{2\Dim}$ represents the eigenvalues of matrix $\Covparatworealpop$, which is introduced in equation~\eqref{eq:Covparatworealpop}.

By combining inequalities~\eqref{eq:paratwo_err1}~and~\eqref{eq:S2}, we arrive at the conclusion of our analysis, as stated in \Cref{lemma:paratwo_err}. \\
%
%
%
%
%
%
%
\paragraph{Proof of inequality~\eqref{eq:S2}, trace of inverse covariance matrix:}

The next critical step is to establish the trace bound in \eqref{eq:S2}. To achieve this, we first define a ``statistical dimension''~$\StatDim$ as
\begin{align}
	\label{eq:StatDim}
\StatDim \equiv \StatDim(\perturb) \; \defn \; \min\{ k \mid \eigpop_k \geq \Const{\StatDim} \cdot \perturb^2 \}
\end{align}
with the constant $\Const{\StatDim} \geq 1$ to be determined later.
Let the eigenvalue of matrix $\Covparatwo$ be $\eigtwo_1 \geq \eigtwo_2 \geq \cdots \geq \eigtwo_{2\Dim}$.
At this moment, we claim the following relationships hold:
\begin{subequations}
\begin{align}
	& \!\!\!\!\!\!\!\! \mbox{\emph{Small eigenvalues bound:}} \notag  \\
	& \eigtwo_{\StatDim + 1} \geq \eigtwo_{2\Dim - \StatDim + 2} \geq \cdots \geq \eigtwo_{2\Dim} \; \geq \; \Const{0}^{-1} \cdot \perturb^2
	\qquad \mbox{with probability $1 - \const{} \, \Term^{-2}$}
	\label{eq:eigtwo_small}   \\
	& \mbox{when }  \Termtwo \geq \Const{1} \cdot \perturb^{-2} \max\{\Dim, \log \Term\}
	\mbox{ and } \perturb \leq \sqrt{\constCovfeat / 8} \, ;
	\notag \\
	& \!\!\!\!\!\!\!\! \mbox{\emph{Large eigenvalues bound:}} \notag  \\
	\label{eq:eigtwo_large}
	&  \eigtwo_k \geq \Const{0}^{-1} \cdot \eigpop_k
	\quad \mbox{for $k = 1, 2, \ldots, \StatDim$}
	\qquad \qquad \quad \mbox{with probability $1 - \const{} \, \Term^{-2}$}   \\
	& \mbox{under the conditions of \Cref{lemma:thm_proof_stage2} and when } 
	\Termtwo \geq \Const{2} \cdot \max\{ \perturb^{-2}, \perturb^{-4} \big\} \max\{\Dim, \log \Term\}.
	\notag
\end{align}
\end{subequations}

Armed with these claims, we can now proceed to prove the desired trace bound.
We begin by breaking down the trace of $\Covparatwo^{-1}$ as
\begin{align*}
\trace(\Covparatwo^{-1}) 
\; = \; \sum_{k=1}^{2d}\frac{1}{\eigtwo_k} 
\; \leq \; \sum_{k=1}^{\StatDim} \frac{1}{\eigtwo_k} + \sum_{i=\StatDim+1}^{2\Dim} \frac{1}{\eigtwo_k} \, .
\end{align*}
For the first summation, we apply inequality~\eqref{eq:eigtwo_large} and find that
\mbox{$\eigtwo_k^{-1} \leq \Const{0} \cdot \eigpop_k^{-1} \leq \Const{} \cdot (\eigpop_k + \perturb^2)^{-1}$}
since $\eigpop_k \geq \Const{\StatDim} \, \perturb^2$ for $k = 1,2\ldots,\StatDim$.
For the second summation, inequality~\eqref{eq:eigtwo_small} implies
$\eigtwo_k^{-1} \leq \Const{0} \cdot \perturb^{-2} \leq \Const{} \cdot (\eigpop_k + \perturb^2)^{-1}$ since $\eigpop_k < \Const{\StatDim} \, \perturb^2$ for $i = \StatDim+1, \ldots, 2\Dim$.
Combining these two parts, we obtain 
$ \trace(\Covparatwo^{-1}) 
\; \leq \; \Const{} \cdot (\eigpop_k + \perturb^2)^{-1}$,
which gives the desired trace bound~\eqref{eq:S2}.

We now rigorously prove the claims \eqref{eq:eigtwo_small} and \eqref{eq:eigtwo_large} in turn.


\subsubsection{Proof of inequality~\eqref{eq:eigtwo_small}, small eigenvalues bound \yaqidone}

We introduce an auxiliary matrix that marginalizes out the randomness in $\Covparatwo$ associated with localized exploration. Specifically, we define this matrix as
\begin{align}
\label{eq:Covparatwopop}
\Covparatwopop 
& \; \defn \; \Exp_{\Rad}\big[ \Covparatwo \bigm| \priceone, \{ \feat_t \}_{t=\Termone+1}^{\Termone + \Termtwo} \big]  
\\ 
\notag
& \; = \; \frac{1}{2\Termtwo}\sum\limits_{t=\Termone+1}^{\Termone + \Termtwo}
\! \Bigg\{ \!\!
\begin{bmatrix}
{\feat}_t\\{\feat}_t(\priceone({\feat}_t)+\perturb)
\end{bmatrix}\begin{bmatrix}
{\feat}_t\\{\feat}_t(\priceone({\feat}_t)+\perturb)
\end{bmatrix}^{\top}
\!\! +
\begin{bmatrix}
{\feat}_t\\{\feat}_t(\priceone({\feat}_t)-\perturb)
\end{bmatrix}\begin{bmatrix}
{\feat}_t\\{\feat}_t(\priceone({\feat}_t)-\perturb)
\end{bmatrix}^{\top} \! \Bigg\} \, .
\end{align}
The bound~\eqref{eq:eigtwo_small} on the smallest eigenvalue of $\Covparatwo$ relies on two key properties of matrix $\Covparatwopop$:
\begin{subequations}
\begin{align}
\label{eq:Covparatwopop1}
& \!\!\!\! \mbox{\emph{Lower bound on $\eigmin(\Covparatwopop)$:}} \!\!\!\!\!\!\!\! \notag  \\
& \Prob\big\{ \eigmin(\Covparatwopop) \; \geq \;  \Const{0}^{-1} \cdot \perturb^2 \big\}
\geq 1 - \Term^{-2}
\quad && \mbox{if $\Termtwo \geq \Const{3} \cdot \max\{ \Dim, \log \Term \}$
	and $\perturb \leq \sqrt{\constCovfeat / 8}$} \, ,   \\
& \!\!\!\! \mbox{\emph{Concentration bound on the difference $\Covparatwo - \Covparatwopop$:}}\!\!\!\!\!\!\!\!\!\!\!\!\!\!\!\!\!\!\!\!\!\!\!\! \notag  \\
\label{eq:Covparatwopop2}
& \Prob\big\{ \norm{\Covparatwo - \Covparatwopop}_2 \leq (2\Const{0})^{-1} \cdot \perturb^2 \big\} \geq 1 - \Term^{-2}
&& \mbox{if } \Termtwo \geq \Const{4} \cdot \frac{\max\{\Dim, \log \Term\}}{\min\{ \perturb, \perturb^2 \}}.
\end{align}
\end{subequations}
With these two properties in hand, we can now bound the smallest eigenvalue of $\Covparatwo$.
By combining inequalities~\eqref{eq:Covparatwopop1}~and~\eqref{eq:Covparatwopop2}, we establish that if $\Termtwo \geq \Const{1} \cdot \perturb^{-2} \max\{\Dim, \log \Term\}$, then with probability at least $1 - \Term^{-2}$,
\begin{align*}
\eigmin(\Covparatwo) \; \geq \; \eigmin(\Covparatwopop) - \norm[\big]{\Covparatwo - \Covparatwopop}_2 \; \geq \; \perturb^2 / \Const{0} - \perturb^2 / (2 \Const{0}) = \perturb^2 / (2 \Const{0}) \, .
\end{align*}
This verifies the bound in inequality~\eqref{eq:eigtwo_small}. 

We will now proceed to prove the two key claims \eqref{eq:Covparatwopop1}~and~\eqref{eq:Covparatwopop2}, which form the foundation for the result.


\paragraph{Proof of inequality~\eqref{eq:Covparatwopop1}, lower bound on $\eigmin(\Covparatwopop)$:}
In this part, our goal is to show that 
for any unit vector $(\vecu, \vecv) \in \Sphere^{2d-1}$,
\begin{align}
\label{eq:Covparatwopop_eigmin}
\begin{bmatrix}
\vecu\\\vecv
\end{bmatrix}^{\top}\Covparatwopop\begin{bmatrix}
\vecu\\\vecv
\end{bmatrix}
\; = \; \frac{1}{\Termtwo}\sum\limits_{t=\Termone+1}^{\Termone + \Termtwo} \Big\{ \underbrace{\left(\vecu^{\top}{\feat}_t+\priceone({\feat}_t) \cdot \vecv^{\top}{\feat}_t\right)^2}_{Z_1} + \underbrace{\left(\perturb \cdot \vecv^{\top}{\feat}_t\right)^2}_{Z_2} \Big\}
\; \geq \; \Const{0}^{-1} \cdot \perturb^2 \, .
\end{align}
To approach this, we analyze two distinct cases based on the value of $\norm{\vecv}_2^2$. Each case highlights whether the first term ($Z_1$) or the second term ($Z_2$) in \eqref{eq:Covparatwopop_eigmin} is more dominant.
We divide the analysis as follows:
\begin{align*}
& \mbox{Case (i):} ~~ \norm{\vecv}_2^2 \; \leq \, \const{0},
& \mbox{Case (ii):} ~~ \norm{\vecv}_2^2 \; > \, \const{0},
\end{align*}
where constant $\const{0}$ serves as the threshold between the two cases and will be determined later.

In the next steps, we will select a proper constant $\const{0} > 0$ and then analyze each case separately, showing that the bound~\eqref{eq:Covparatwopop_eigmin} holds in both scenarios.  \\

\noindent \emph{Choice of threshold $\const{0}$:}

In our analysis, we will frequently refer to the bounds on the eigenvalues of the empirical covariance matrix
\begin{align*}
\Covfeattwo \; \defn \; \frac{1}{\Termtwo} \sum_{t = 1}^{\Termtwo} \feat_t \feat_t^{\top} \; \in \Real^{\Dim \times \Dim} \, .
\end{align*}
Similar to previous results like \eqref{eq:diff_Covparaonerealpop} and \eqref{eq:diff_Covparaonerealpop_cor}, we can establish that, under Assumptions~\ref{asp:pure-online-second-moment-nondegenerate}~and~\ref{asp:basic-regularity}, for some constant $\Const{2} > 0$, when $\Termtwo \geq \Const{2} \cdot \max\{\Dim, \log \Term\}$, the following eigenvalue bounds hold:
\begin{align}
\label{eq:Covfeattwo_IdMt}
\constnew \, \, \IdMt \; \preceq \; \Covfeattwo \; \preceq \; \Constnew \, \IdMt
\qquad \mbox{with probability at least $1 - \Term^{-2}$}.
\end{align}

Using the constants $\constCovfeat$ and $\ConstCovfeat$, we can now define the splitting point $\const{0}$ for the two cases in our earlier analysis. We set it as
\begin{align*}
\const{0} \; \defn \; \frac{\constCovfeat}{2 \, \constCovfeat + 16 \ConstCovfeat \priceub^2} \; .
\end{align*}

With the splitting point $\const{0}$ now defined, we can move forward to analyze each case and verify the bound~\eqref{eq:Covparatwopop_eigmin}.  \\

\noindent \emph{Case (i):}

In this case, where the norm of $\vecv$ is small, we primarily focus on the term $Z_1$ from equation \eqref{eq:Covparatwopop_eigmin}.
We begin by lower bounding the quadratic form as follows:
\begin{align}
\begin{bmatrix}
\vecu\\\vecv
\end{bmatrix}^{\top}\Covparatwopop\begin{bmatrix}
\vecu\\\vecv
\end{bmatrix}
& \; \geq  \; \frac{1}{\Termtwo}\sum\limits_{t=\Termone+1}^{\Termone + \Termtwo}
\left(\vecu^{\top}{\feat}_t+\priceone({\feat}_t) \cdot \vecv^{\top}{\feat}_t\right)^2
\notag \\
\label{eq:Covparatwopop_eigmin1}
& \; \ge \; \frac{1}{\Termtwo} \sum\limits_{t=\Termone+1}^{\Termone + \Termtwo} \Big\{ \frac{1}{2}(\vecu^{\top}{\feat}_t)^2-u^2(\vecv^{\top}{\feat}_t)^2 \Big\} \, .
\end{align}
In the second inequality in equation~\eqref{eq:Covparatwopop_eigmin1}, we have used the relation $(a+b)^2 = a^2 + 2ab + b^2 \geq a^2 - (\frac{1}{2}a^2 + 2b^2) + b^2 = \frac{1}{2} a^2 - b^2$ for any $a,b \in \Real$.

We then express the right-hand side of inequality~\eqref{eq:Covparatwopop_eigmin1} in terms of the covariance matrix~$\Covfeattwo$. It follows that
\begin{align*}
\begin{bmatrix}
\vecu\\\vecv
\end{bmatrix}^{\top} \! \Covparatwopop\begin{bmatrix}
\vecu\\\vecv
\end{bmatrix}
\; \geq \; \frac{1}{2} \, \vecu^{\top}\Covfeattwo\vecu-\priceub^2 \cdot \vecv^{\top}\Covfeattwo\vecv 
& \; \stackrel{(a)}{\geq} \; \frac{\constCovfeat}{4} \, \norm{\vecu}_2^2 - \Constnew \, \priceub^2 \cdot \norm{\vecv}_2^2  \\
& \; \stackrel{(b)}{=} \; \frac{\constCovfeat}{4} - \Big\{ \frac{\constCovfeat}{4} + \Constnew \priceub^2 \Big\} \, \norm{\vecv}_2^2
\; \stackrel{(c)}{\geq} \; \frac{\ConstCovfeat}{8} \, .
\end{align*}
Here inequality~$(a)$ is based on the previously established eigenvalue bound~\eqref{eq:Covfeattwo_IdMt} for matrix~$\Covfeattwo$.
Inequality~$(b)$ is due to $\norm{\vecu}_2^2 + \norm{\vecv}_2^2 = 1$.
Inequality~$(c)$ is guaranteed by the condition $\norm{\vecv}_2^2 \leq \const{0}$ in Case~(ii).
Moreover, under the condition $\perturb \leq \sqrt{\constCovfeat / 8}$, we have $\constCovfeat / 8 \geq \Const{0}^{-1} \cdot \perturb^2$. Therefore, the bound~\eqref{eq:Covparatwopop_eigmin} holds, completing the proof for Case~(i). \\

\noindent \emph{Case (ii):}

In this case, where the norm of $\vecv$ is larger, we focus on the term $Z_2$ in expression \eqref{eq:Covparatwopop_eigmin}.
It follows that
\begin{align*}
\begin{bmatrix}
\vecu\\\vecv
\end{bmatrix}^{\top}\Covparatwopop\begin{bmatrix}
\vecu\\\vecv
\end{bmatrix}
\; \geq \; \frac{1}{\Termtwo}\sum\limits_{t=\Termone+1}^{\Termone + \Termtwo} \left(\perturb \cdot \vecv^{\top}{\feat}_{t}\right)^2 
\; \stackrel{(a)}{=} \; \perturb^2 \cdot \vecv^{\top}\Covfeattwo\vecv
\; \stackrel{(b)}{\geq} \; \perturb^2 \cdot \constnew \, \, \norm{\vecv}_2^2
\; \stackrel{(c)}{\geq} \; \frac{\constCovfeat \, \const{0}}{2} \cdot \perturb^2 \, .
\end{align*}
Here equality~$(a)$ comes from the definition of covariance~$\Covfeattwo$. Step~$(b)$ uses the results from bound~\eqref{eq:Covfeattwo_IdMt}. The last inequality~$(c)$ is ensured by the assumption $\norm{\vecv}_2^2 > \const{0}$ in Case~(ii).
Therefore, the bound~\eqref{eq:Covparatwopop_eigmin} holds in Case~(ii).  \\

The two cases discussed above together cover all possible scenarios for the unit vector $(\vecu, \vecv) \in \Sphere^{2d-1}$.
In both cases, we have shown that the bound~\eqref{eq:Covparatwopop_eigmin} holds. Therefore, we conclude that
$\eigmin(\Covparatwopop) \geq \Const{0}^{-1} \cdot \perturb^2$, as claimed in equation~\eqref{eq:Covparatwopop1}.

\paragraph{Proof of inequality~\eqref{eq:Covparatwopop2}, concentration bound on the difference $\Covparatwo - \Covparatwopop$:}

In this part, we establish a concentration bound on the difference between two covariance matrices $\Covparatwo$ and $\Covparatwopop$, as defined in equations~\eqref{eq:Covparatwo}~and~\eqref{eq:Covparatwopop}. 
For any unit vector $(\vecu, \vecv) \in \Real^{2\Dim}$, we express the difference as
\begin{align*}
\begin{bmatrix}
\vecu\\\vecv
\end{bmatrix}^{\top} 
\! \big( \Covparatwo - \Covparatwopop \big) 
\begin{bmatrix}
\vecu\\\vecv
\end{bmatrix}
\; = \; \frac{2 \perturb}{\Termtwo}
\sum_{t=\Termone+1}^{\Termone + \Termtwo}
\Rad_t \cdot \left(\vecu^{\top}{\feat}_t+\priceone({\feat}_t) \cdot \vecv^{\top}{\feat}_t\right)
\left(\vecv^{\top}{\feat}_t\right) \, .
\end{align*}
To simplify notation, we introduce a shorthand
$\zeta_t(\vecu, \vecv)
\defn \Rad_t \cdot \left(\vecu^{\top}{\feat}_t+\priceone({\feat}_t) \cdot \vecv^{\top}{\feat}_t\right)
(\vecv^{\top}{\feat}_t)$,
which allows us to rewrite the quadratic form as
\begin{align}
\label{eq:zeta}
\begin{bmatrix}
\vecu\\\vecv
\end{bmatrix}^{\top} 
\! \big( \Covparatwo - \Covparatwopop \big) 
\begin{bmatrix}
\vecu\\\vecv
\end{bmatrix}
\; = \; \frac{2 \perturb}{\Termtwo}
\sum_{t=\Termone+1}^{\Termone + \Termtwo}
\zeta_t(\vecu, \vecv) \, .
\end{align}

A key observation here is that $\{ \zeta_t(\vecu, \vecv) \}_{t=\Termone + 1}^{\Termone + \Termtwo}$ are i.i.d. sub-exponential random variables.
To bound the magnitude of $\zeta_t(\vecu, \vecv)$, we calculate its $\psi_1$-norm:
\begin{align*}
\psinorm{1}{\zeta_t(\vecu, \vecv)}  
& \; \leq \; \supnorm{\Rad_t} \cdot 2 \, 
\psinorm[\big]{2}{\vecu^{\top}{\feat}_t+\priceone({\feat}_t) \cdot \vecv^{\top}{\feat}_t} \, 
\psinorm{2}{\vecv^{\top}{\feat}_t}   \\
& \; \leq \; \supnorm{\Rad_t} \cdot 2
\big\{ \psinorm{2}{\vecu^{\top}{\feat}_t} \! + \supnorm{\priceone} \! \cdot \psinorm{2}{\vecv^{\top}{\feat}_t} \big\} 
\psinorm{2}{\vecv^{\top}{\feat}_t}   \\
& \; \leq \; 1 \cdot 2 \, \big\{ (1 + \priceub) \stdfeat \} \, \stdfeat  
= 2 (1 + \priceub) \stdfeat^2 = \bigO(1).
\end{align*}

We now apply Bernstein's inequality (Theorem~2.8.1 in book~\cite{vershynin2018high}) for sub-exponential random variable, combined with a standard covering argument over the unit sphere $\Sphere^{2\Dim-1}$. This gives us
\begin{align*}
\sup_{(\vecu, \vecv) \in \Sphere^{2\Dim - 1}} \,
\frac{1}{\Termtwo} \sum_{t=\Termone+1}^{\Termone + \Termtwo}
\zeta_t(\vecu, \vecv)
\; \leq \; \Const{} \cdot \bigg\{ \sqrt{\frac{\Dim + \log \Term}{\Termtwo}} + \frac{\Dim + \log \Term}{\Termtwo} \bigg\}
\end{align*}
with probability at least $1 - \Term^{-2}$.

From equation \eqref{eq:zeta}, we deduce that
\begin{align}
\label{eq:diff_Covparatwopop}
\norm[\big]{\Covparatwo - \Covparatwopop}_2
\; \leq \; \Const{} \cdot \perturb \, \Big\{ \sqrt{\frac{\Dim + \log \Term}{\Termtwo}} + \frac{\Dim + \log \Term}{\Termtwo} \Big\} \, .
\end{align}
Therefore, by ensuring that
\begin{align*}
\Termtwo \; \geq \; \Const{4} \cdot \frac{\max\{\Dim, \log \Term\}}{\min\{ \perturb, \perturb^2 \}} \, ,
\end{align*}
we reduce the bound~\eqref{eq:diff_Covparatwopop} to
$\norm[\big]{\Covparatwo - \Covparatwopop}_2 \leq (2\Const{0})^{-1} \cdot \perturb^2$, as claimed in equation~\eqref{eq:Covparatwopop2}.


\subsubsection{Proof of inequality~\eqref{eq:eigtwo_large}, large eigenvalues bound \yaqidone}
We now turn to bounding the first $\StatDim$ eigenvalues $\eigtwo_1, \eigtwo_2, \ldots, \eigtwo_{\StatDim}$ of matrix~$\Covparatwo$.
To begin with, we introduce a helpful auxiliary matrix
\begin{align}
\Covparatwopopstar
& \; \defn \; \frac{1}{\Termtwo} \sum\limits_{t=\Termone+1}^{\Termone + \Termtwo}
\begin{bmatrix}
{\feat}_t\\{\feat}_t \, \pricestar({\feat}_t)
\end{bmatrix}\begin{bmatrix}
{\feat}_t\\{\feat}_t \, \pricestar({\feat}_t)
\end{bmatrix}^{\top}
\; \in \Real^{\Dim \times \Dim}.
\end{align}
Let $\eigpopstar_1 \geq \eigpopstar_2 \geq \ldots \geq \eigpopstar_{2\Dim}$ be its associated eigenvalues, arranged in a decending order.

Our first task is to establish a lower bound on the eigenvalues of matrix $\Covparatwopopstar$ using the eigenvalues $\eigpop_1, \eigpop_2, \ldots$ of matrix $\Covparatworealpop$ (as defined in equation~\eqref{eq:Covparatworealpop}). By taking the expectation of $\Covparatwopopstar$, we can show that it converges to the matrix $\Covparatworealpop$. Therefore, by leveraging a concentration inequality, we can then prove the following claim:
\begin{subequations}
\vspace{.5em}

\indent \emph{Lower bound on $\eigpopstar_k$:}
\begin{align}
\label{eq:Covparatwopopstar1}
& \eigpopstar_k \; \geq \; \frac{\eigpop_k}{2}
\qquad \mbox{for $k = 1, 2, \ldots, \StatDim$}  \\
& \mbox{with probability at least $1 - \Term^{-2}$, if $\Termtwo \geq \Const{5} \cdot \max\{ \perturb^{-2}, \perturb^{-4} \} \, \max\{ \Dim, \log \Term \}$} \, .
\notag
\end{align}

Next, we consider the relationship between $\Covparatwo$ and $\Covparatwopopstar$ by examing the eigen spaces of matrix $\Covparatwopopstar$. 
For $k = 1,2\ldots,\StatDim$, let $\eigSp_k \subseteq \Real^{2\Dim}$ denote the linear space spanned by the first~$k$ eigenvectors of $\Covparatwopopstar$ associated with eigenvalues $\eigpopstar_1, \eigpopstar_2, \ldots, \eigpopstar_k$.
We then state the key claim:
\vspace{.5em}

\indent \emph{Difference between $\Covparatwo$ and $\Covparatwopopstar$:}
\begin{align}
\label{eq:Covparatwopopstar2}
& \abs[\bigg]{ \begin{bmatrix}
		\vecu\\\vecv
	\end{bmatrix}^{\top}
	\! \! (\Covparatwo - \Covparatwopopstar)
	\begin{bmatrix}
		\vecu\\\vecv
\end{bmatrix}}
\; \leq \; \frac{1}{2} \cdot \begin{bmatrix} \vecu \\ \vecv \end{bmatrix}^{\top} \Covparatwopopstar \begin{bmatrix} \vecu \\ \vecv \end{bmatrix} 
~~ \mbox{for any vector } (\vecu, \vecv) \in \eigSp_{\StatDim}, \\
& \mbox{if \Cref{lemma:thm_proof_stage2}~and~inequality~\eqref{eq:Covfeattwo_IdMt} hold}.
\notag
\end{align}
\end{subequations}

With the two key claims~\eqref{eq:Covparatwopopstar1}~and~\eqref{eq:Covparatwopopstar2} established, we are now equipped to prove inequality~\eqref{eq:eigtwo_large}.
We start by applying the max-min theorem for eigenvalues. It tells us that for any $k = 1, 2, \ldots, \StatDim$, the $k$-th eigenvalue of matrix $\Covparatwo$ can be characterized as
\begin{align*}
\eigtwo_k
\; = \max_{\begin{subarray}{c} \mathds{M} \subseteq \Real^{2\Dim} \\
	\dim(\mathds{M}) = k
	\end{subarray}}
	\min_{\begin{subarray}{c} (\vecu, \vecv) \in \mathds{M} \\ \norm{(\vecu, \vecv)}_2 = 1 \end{subarray}}
	\begin{bmatrix} \vecu \\ \vecv \end{bmatrix}^{\top} \Covparatwo \begin{bmatrix} \vecu \\ \vecv \end{bmatrix} \, .
\end{align*}
Notice that the eigenspace $\eigSp_k$ of matrix $\Covparatwopopstar$ is a valid $k$-dimensional subspace of $\Real^{2\Dim}$. Therefore, we choose $\mathds{M} \defn \eigSp_k$ in the max-min formulation and find that
\begin{align}
	\label{eq:Covparatwopopstar3}
	\eigtwo_k \; \geq \min_{\begin{subarray}{c} (\vecu, \vecv) \in \eigSp_k \\ \norm{(\vecu, \vecv)}_2 = 1 \end{subarray}}
	\begin{bmatrix} \vecu \\ \vecv \end{bmatrix}^{\top} \Covparatwo \begin{bmatrix} \vecu \\ \vecv \end{bmatrix} \, .
\end{align}

Using inequality~\eqref{eq:Covparatwopopstar2}, we know that for any unit vector $(\vecu, \vecv) \in \eigSp_k \subseteq \eigSp_{\StatDim}$, we can bound the difference between the quadratic forms of $\Covparatwo$ and $\Covparatwopopstar$ as follows
\begin{align*}
	\begin{bmatrix} \vecu \\ \vecv \end{bmatrix}^{\top} \Covparatwo \begin{bmatrix} \vecu \\ \vecv \end{bmatrix} 
	\; \geq \; \begin{bmatrix} \vecu \\ \vecv \end{bmatrix}^{\top} \Covparatwopopstar \begin{bmatrix} \vecu \\ \vecv \end{bmatrix}
	- \abs[\bigg]{ \begin{bmatrix}
			\vecu\\\vecv
		\end{bmatrix}^{\top}
		\! \! (\Covparatwo - \Covparatwopopstar)
		\begin{bmatrix}
			\vecu\\\vecv
	\end{bmatrix}} 
	\; \geq \; \frac{1}{2} \, \begin{bmatrix} \vecu \\ \vecv \end{bmatrix}^{\top} \Covparatwopopstar \begin{bmatrix} \vecu \\ \vecv \end{bmatrix} \, .
\end{align*}
%
Moreover, the property of the eigen space $\eigSp_k$ of matrix $\Covparatwopopstar$ ensures that
\begin{align*}
	\begin{bmatrix} \vecu \\ \vecv \end{bmatrix}^{\top} \Covparatwopopstar \begin{bmatrix} \vecu \\ \vecv \end{bmatrix} \; \geq \; \eigpopstar_k
	\qquad \mbox{for any unit vector } \begin{bmatrix} \vecu \\ \vecv \end{bmatrix}  \in \eigSp_k \, .
\end{align*}
By combining all the steps, we reduce inequality~\eqref{eq:Covparatwopopstar3} to $\eigtwo_k \geq \eigpopstar_k / 2$. Applying the bound~\eqref{eq:Covparatwopopstar1}, we arrive at the conclusion $\eigtwo_k \geq \eigpop_k / 4$, which completes the proof of inequality~\eqref{eq:eigtwo_large}. \\

Now let us turn to the proofs of claims~\eqref{eq:Covparatwopopstar1}~and~\eqref{eq:Covparatwopopstar2}.


\paragraph{Proof of inequality~\eqref{eq:Covparatwopopstar1}, lower bound on $\eigpopstar_k$:}


Just as we analyzed the difference between $\Covparaone$ and $\Covparaonerealpop$ in Stage 1 (as shown in bound~\eqref{eq:diff_Covparaonerealpop}), we now perform a similar analysis in Stage 2. This time, we are concerned with bounding the difference between the auxiliary matrix $\Covparatwopopstar$ and the true covariance matrix $\Covparatworealpop$.

To do this, we require Stage~2 to be sufficiently long so that
\begin{align*}
\Termtwo \; \geq \; C \cdot \frac{\max\{ \Dim, \log \Term \}}{\min\{ \perturb^2, \, \perturb^4 \}} \, .
\end{align*}
Under this condition,
with probability at least $1 - \Term^{-2}$, we can ensure that
$\norm[\big]{\Covparatwopopstar - \Covparatworealpop}_2 \leq \perturb^2/2$.

We then use Weyl’s inequality to relate the eigenvalues of $\Covparatwopopstar$ to those of $\Covparatworealpop$. 
It tells us that for any $k = 1, 2, \ldots, \StatDim$,
\begin{align*}
\eigpopstar_k
\; \geq \; \eigpop_k - \norm[\big]{\Covparatwopopstar - \Covparatworealpop}_2
\; \geq \; \eigpop_k - \frac{\perturb^2}{2} \; \geq \; \frac{\eigpop_k}{2} \, ,
\end{align*}
which establishes inequality~\eqref{eq:Covparatwopopstar1}.
Here, we have used the fact that $\perturb^2 \leq \eigpop_{\StatDim} \leq \eigpop_k$ due to our choice of statistical dimension $\StatDim$ in definition~\eqref{eq:StatDim}.

\paragraph{Proof of inequality~\eqref{eq:Covparatwopopstar2}, difference between $\Covparatwo$ and $\Covparatwopopstar$:}
It suffices to consider any unit vector $(\vecu, \vecv) \in \eigSp_{\StatDim}$.
We find that by definitions of matrices $\Covparatwo$~and~$\Covparatwopopstar$, it holds that
\begin{multline}
\label{eq:diff_Covparatwopopstar}
\abs[\bigg]{ \begin{bmatrix}
	\vecu\\\vecv
\end{bmatrix}^{\top}
\! \! (\Covparatwo - \Covparatwopopstar)
\begin{bmatrix}
	\vecu\\\vecv
	\end{bmatrix}}
	= \abs[\bigg]{ \frac{1}{\Termtwo} \sum_{t=\Termone+1}^{\Termone + \Termtwo} \Big\{ \big(\vecu^{\top}{\feat}_t+\price_t \; \vecv^{\top}{\feat}_t\big)^2-\big(\vecu^{\top}{\feat}_t+\pricestar(\feat_t) \; \vecv^{\top}{\feat}_t\big)^2 \Big\} }  \\
	\leq \frac{1}{\Termtwo} \sum_{t=\Termone+1}^{\Termone + \Termtwo} \Big\{
	\frac{1}{4} \big(\vecu^{\top}{\feat}_t + \pricestar({\feat}_t) \; \vecv^{\top}{\feat}_t\big)^2
	+ 5 \cdot \big(\price_t - \pricestar(\feat_t)\big)^2 (\vecv^{\top}{\feat}_t)^2 \Big\} \, .
\end{multline}
Here we have used the relation $(x + \Delta x)^2 - x^2 = 2 x \Delta x + (\Delta x)^2 \leq \{ \frac{1}{4} x^2 + 4 (\Delta x)^2 \} + (\Delta x)^2 = \frac{1}{4} x^2 + 5 (\Delta x)^2$ for any $x, \Delta x \in \Real$.

We now use the fact that $\abs{\price_t - \pricestar(\feat_t)} \leq 2 \, \perturb$, as given in \Cref{lemma:thm_proof_stage2}. Substituting this into the bound above and rewriting the summation using matrix forms involving $\Covparatwopopstar$ and $\Covfeattwo$, we reduce inequality~\eqref{eq:diff_Covparatwopopstar} to
\begin{align}
	\label{eq:diff_Covparapopstar1}
\abs[\bigg]{ \begin{bmatrix}
	\vecu\\\vecv
\end{bmatrix}^{\top}
\! \! (\Covparatwo - \Covparatwopopstar)
\begin{bmatrix}
	\vecu\\\vecv
	\end{bmatrix}}
	& \leq \; \frac{1}{4} \cdot \begin{bmatrix} \vecu \\ \vecv \end{bmatrix}^{\top} \Covparatwopopstar \begin{bmatrix} \vecu \\ \vecv \end{bmatrix} + \underbrace{20 \, \perturb^2 \cdot \vecv^{\top} \Covfeattwo \vecv}_{\Gamma}.
\end{align}
We next focus on the second term~$\Gamma$ on the right-hand side and derive that
\begin{align}
	\label{eq:diff_Covparapopstar2}
	\Gamma \; = \; 20 \, \perturb^2 \cdot \vecv^{\top} \Covfeattwo \vecv
	& \; \stackrel{(a)}{\leq} \; 20 \,  \perturb^2 \cdot \Constnew  \; \stackrel{(b)}{\leq} \;  \frac{1}{4} \, \eigpop_{\StatDim}
	\; \stackrel{(c)}{\leq} \; \frac{1}{4} \cdot \begin{bmatrix} \vecu \\ \vecv \end{bmatrix}^{\top} \Covparatwopopstar \begin{bmatrix} \vecu \\ \vecv \end{bmatrix}.
\end{align}
Here inequality~$(a)$ follows from the eigenvalue upper bound on the covariance matrix $\Covfeattwo$, as given in inequality~\eqref{eq:Covfeattwo_IdMt}.
Inequality~$(b)$ uses the relation $\eigpop_{\StatDim} \geq \Const{\StatDim} \, \perturb^2$, which arises from the definition of the statistical dimension $\StatDim$ in equation~\eqref{eq:StatDim}. This inequality~$(b)$ holds when the constant $\Const{\StatDim}$ is chosen to be sufficiently large.
The final inequality~$(c)$ is based on property of the eigen space $\eigSp_{\StatDim}$.
By substituting the upper bound for the term $\Gamma$ from inequality~\eqref{eq:diff_Covparapopstar2} into inequality~\eqref{eq:diff_Covparapopstar1}, we derive inequality~\eqref{eq:Covparatwopopstar2}, as previously claimed.


\subsection{Proof of Lemma~\ref{lemma:thm_proof_stage3}, pricing accuracy in Stage~3 \yaqidone}
\label{sec:price_stage_3}

The analysis for pricing accuracy in Stage~3 mirrors that of Stage~2. 
By applying the same techniques used in proving \Cref{lemma:thm_proof_stage2} (as outlined in \Cref{sec:price_stage_2}), we can extend the result from \Cref{lemma:paratwo_err}, which provides an estimation error bound on $\norm{\paratwo - \parastar}_2$, to control the price difference $\abs[\big]{\pricetwo(\feat_t) - \pricestar(\feat_t)}$ for $t = \Termtwo + 1, \Termtwo + 2, \ldots, \Term$.

In particular, we establish that, with probability at least $1 - \const{} \, \Term^{-2}$, the following holds for all time steps $t = \Termtwo + 1, \Termtwo + 2, \ldots, \Term$ in Stage~3:
	\begin{align}
		\label{eq:featparatwo}
		\max \Big\{ \abs[\big]{ \feat_t^{\top} (\parainttwo - \paraintstar) }, \,
		\abs[\big]{ \feat_t^{\top} (\paraslptwo - \paraslpstar) } \Big\}
		& \; \leq \; \Const{} \cdot \sqrt{\bigg\{ \sum_{k=1}^{2\Dim} \frac{1}{\eigpop_k + \perturb^2} \bigg\} \frac{1}{\Termtwo}} \; \log\Term \, .
	\end{align}
	
Moreover, under the condition
\begin{align*}
	\Termtwo \; \geq \; \bigg\{ \sum_{k=1}^{2\Dim} \frac{1}{\eigpop_k + \perturb^2} \bigg\} \, \log^2 \Term \, ,
\end{align*}
we have
\begin{align}
	\label{eq:cond_featparaslptwo}
	\abs[\big]{\feat_t^{\top} (\paraslptwo - \paraslpstar)} \; \leq \; \frac{\featparalb}{2}
	\qquad \mbox{for $t = \Termtwo + 1, \Termtwo + 2, \ldots, \Term$} \, .
\end{align}
Using a similar calculation as in inequality~\eqref{eq:price_diff_stage2} from the analysis of Stage~2, we can derive the following bound for Stage 3:
\begin{align}
	\label{eq:diff_price_stage3}
	\abs[\big]{\pricetwo(\feat_t) - \pricestar(\feat_t)} 
	\; \leq \; \Const{} \cdot \Big\{ \abs[\big]{ \feat_t^{\top} (\parainttwo - \paraintstar) }
	+ \abs[\big]{ \feat_t^{\top} (\paraslptwo - \paraslpstar) } \Big\} \, .
\end{align}

By combining the bounds~\eqref{eq:featparatwo}~and~\eqref{eq:diff_price_stage3}, we have 
\begin{align}
		\abs[\big]{\pricetwo(\feat_t) - \pricestar(\feat_t)} \, \leq \; & \Const{} \cdot \sqrt{\bigg\{ \sum_{k=1}^{2\Dim} \frac{1}{\eigpop_k + \perturb^2} \bigg\} \frac{1}{\Termtwo}} \; \log\Term
\end{align}
By setting $C_1$ large enough such that $C/\sqrt{C_1}\le \bar\delta$, and plugging in the condition~\eqref{eq:cond_thm_proof_stage2} on $T_2$ , we obtain $\abs[\big]{\pricetwo(\feat_t) - \pricestar(\feat_t)}\le \bar\delta$. As $\pricestar(\feat_t)\in[\ell+\bar\delta,r-\bar\delta]$, it holds that $\pricetwo(\feat_t)\in[\ell,u]$, hence $p_t=\pricetwo(\feat_t)$, and we complete the proof of \Cref{lemma:thm_proof_stage3}.

\subsection{Proof of auxiliary results \yaqidone}
\label{sec:ub_cond}

In this section, we present the proofs of several auxiliary results that are essential for establishing \Cref{thm:ub_simple} and \Cref{thm:ub_main}.

\subsubsection{Proof of Lemma~\ref{lemma:exploration}, spectrum analysis of the covariance matrix $\Covparatworealpop$ \yaqidone}
\label{sec:proof:lemma:exploration}

From the expression \eqref{eq:opt_price} of the optimal pricing strategy $\pricestar$, it is straightforward to verify that
\begin{align*}
	\begin{bmatrix}
		\paraintstar  \\  2 \paraslpstar
	\end{bmatrix}^{\top}
	\Covparatworealpop
	\begin{bmatrix}
		\paraintstar  \\  2 \paraslpstar
	\end{bmatrix}
	& \; = \; \Exp_{\feat \sim \distr}
	\bigg[ \Big\{ \feat^{\top} \paraintstar - \frac{\feat^{\top} \paraintstar}{2 \, \feat^{\top} \paraslpstar} \, \feat^{\top} (2\paraslpstar) \Big\}^2 \bigg]
	\; = \; 0 \, .
\end{align*}
This implies that the vector $\begin{pmatrix} \paraintstar, \, 2 \paraslpstar \end{pmatrix} \in \Real^{2 \Dim}$ lies in the null space of the matrix $\Covparatworealpop$, confirming that $\rank(\Covparatworealpop) \leq 2 \Dim - 1$, as stated in the first part of \Cref{lemma:exploration}. \\

Next, we proceed to prove the remaining two claims in \Cref{lemma:exploration} using Assumptions~\ref{asp:basic-regularity} and~\ref{asp:exploration}.
Specifically, we consider any vectors $\vecu, \vecv \in \Real^{\Dim}$ such that the combined vector $\begin{pmatrix} \vecu, \, \vecv \end{pmatrix} \in \Real^{2\Dim}$ is a unit vector orthogonal to $\begin{pmatrix} \paraintstar, \, 2 \paraslpstar \end{pmatrix}$. This orthogonality condition can be expressed as:
\begin{align}
	\label{eq:uv_orth}
	0 \; = \;
	\begin{bmatrix}
		\vecu \\ \vecv
	\end{bmatrix}^{\top}
	\begin{bmatrix}
		\paraintstar  \\  2 \paraslpstar
	\end{bmatrix}
	\; = \; \vecu^{\top} \paraintstar + 2 \, \vecv^{\top} \paraslpstar \, .
\end{align}
Our goal is to establish a lower bound for the quadratic form
\begin{align*}
	\begin{bmatrix}
		\vecu  \\  \vecv
	\end{bmatrix}^{\top}
	\Covparatworealpop
	\begin{bmatrix}
		\vecu  \\  \vecv
	\end{bmatrix} \, .
\end{align*}
By leveraging the orthogonality condition \eqref{eq:uv_orth} and the given assumptions, we will show that this quadratic form is bounded below by a positive constant, thereby confirming the remaining claims in \Cref{lemma:exploration}. \\

Given the expression for the optimal pricing strategy $\pricestar$ in equation \eqref{eq:opt_price}, we have
\begin{align*}
	\begin{bmatrix}
		\vecu  \\  \vecv
	\end{bmatrix}^{\top}
	\Covparatworealpop
	\begin{bmatrix}
		\vecu  \\  \vecv
	\end{bmatrix}
	& \; = \; \Exp_{\feat \sim \distr}
	\bigg[ \Big\{ \feat^{\top} \vecu - \frac{\feat^{\top} \paraintstar}{2 \, \feat^{\top} \paraslpstar} \, \feat^{\top} \vecv \Big\}^2 \bigg]  \\
	& \; = \; \Exp_{\feat \sim \distr}
	\bigg[ \frac{1}{4 \, (\feat^{\top} \paraslpstar)^2} \big\{2 \, (\feat^{\top} \paraslpstar) (\feat^{\top} \vecu) - (\feat^{\top} \paraintstar) (\feat^{\top} \vecv) \big\}^2 \bigg] \, .
\end{align*}
Using the bound $\abs{\feat^{\top} \paraslpstar} \geq \featparalb$ from \Cref{asp:basic-regularity}, we obtain
\begin{align}
	\notag
	\begin{bmatrix}
		\vecu  \\  \vecv
	\end{bmatrix}^{\top}
	\Covparatworealpop
	\begin{bmatrix}
		\vecu  \\  \vecv
	\end{bmatrix}
	& \; \geq \; \frac{1}{4 \, \featparalb^2} \, \Exp_{\feat \sim \distr}
	\Big[ \big\{2 \, (\feat^{\top} \paraslpstar) (\feat^{\top} \vecu) - (\feat^{\top} \paraintstar) (\feat^{\top} \vecv) \big\}^2 \Big]  \\
	\label{eq:Sstareig}
	& \; = \; \frac{1}{4 \, \featparalb^2} \, \Exp_{\feat \sim \distr}
	\Big[ \big\{ \feat^{\top} \big( 2 \, \paraslpstar \vecu^{\top} - \paraintstar \vecv^{\top} \big) \, \feat \big\}^2 \Big] \, .
\end{align}
To simplify further, we introduce a symmetrized matrix
\begin{align*}
	\AMt \; \defn \; \paraslpstar \vecu^{\top} + \vecu (\paraslpstar)^{\top} - \tfrac{1}{2} \, \paraintstar \vecv^{\top} - \tfrac{1}{2} \, \vecv (\paraintstar)^{\top} \; \in \; \Sym^{\Dim} \, .
\end{align*}
We can reformulate the expression in \eqref{eq:Sstareig} as
\begin{align*}
	\begin{bmatrix}
		\vecu  \\  \vecv
	\end{bmatrix}^{\top}
	\Covparatworealpop
	\begin{bmatrix}
		\vecu  \\  \vecv
	\end{bmatrix}
	\; \geq \;
	\frac{1}{4 \, \featparalb^2} \, \Exp_{\feat \sim \distr}
	\big[ ( \feat^{\top} \AMt \, \feat )^2 \big] \, .
\end{align*}
Notice that $\AMt$ is a symmetric matrix with $\rank(\AMt) \leq 4$. Applying the anti-concentration condition from \Cref{asp:exploration}, we get
\begin{align}
	\label{eq:SstarAFnorm}
	\begin{bmatrix}
		\vecu  \\  \vecv
	\end{bmatrix}^{\top}
	\Covparatworealpop
	\begin{bmatrix}
		\vecu  \\  \vecv
	\end{bmatrix}
	& \; \geq \; \frac{\constcon}{4 \, \featparalb^2} \, \distrnorm{ \AMt }{F}^2 \, .
\end{align}

In the following, we establish that the Frobenius norm $\distrnorm{\AMt}{F}$ is bounded below by a constant, using \Cref{asp:exploration}.2. Through standard algebraic calculations, we derive
\begin{align}
	\notag
	\distrnorm{ \AMt }{F}^2
	& \; = \; \trace \Big\{ \big\{ \paraslpstar \vecu^{\top} + \vecu (\paraslpstar)^{\top} - \tfrac{1}{2} \, \paraintstar \vecv^{\top} - \tfrac{1}{2} \, \vecv (\paraintstar)^{\top} \big\}^2 \Big\}  \\
	\notag
	& \; = \;  2 \, \distrnorm{\paraslpstar}{2}^2 \, \distrnorm{\vecu}{2}^2 - 2 \, (\paraintstar)^{\top} \paraslpstar \vecu^{\top} \vecv + \frac{1}{2} \, \distrnorm{\paraintstar}{2}^2 \, \distrnorm{\vecv}{2}^2 \\
	\label{eq:AMtFnorm}
	& \qquad + 2 \, (\vecu^{\top} \paraslpstar)^2
	+ \frac{1}{2} \, (\vecv^{\top} \paraintstar)^2 -  2 \, \vecv^{\top} \paraslpstar \vecu^{\top} \paraintstar \, .
\end{align}
From equation~\eqref{eq:uv_orth}, we know that $\vecu^{\top} \paraintstar = -2 \vecv^{\top} \paraslpstar$, which implies
\begin{align*}
	- 2 \, \vecv^{\top} \paraslpstar \vecu^{\top} \paraintstar
	\; = \; (\vecu^{\top} \paraintstar)^2 \; \geq \; 0 \, .
\end{align*}
Therefore, equation~\eqref{eq:AMtFnorm} reduces to
\begin{align*}
	\distrnorm{ \AMt }{F}^2
	& \; \geq \; 2 \, \distrnorm{\paraslpstar}{2}^2 \, \distrnorm{\vecu}{2}^2 - 2 \, (\paraintstar)^{\top} \paraslpstar \vecu^{\top} \vecv + \frac{1}{2} \, \distrnorm{\paraintstar}{2}^2 \, \distrnorm{\vecv}{2}^2 \; \nfed \Gamma \, .
\end{align*}
Applying the non-collinearity condition from \Cref{asp:exploration} and the Cauchy-Schwarz inequality, we have
\begin{align*}
	(\paraintstar)^{\top} \paraslpstar \vecu^{\top} \vecv
	\; \leq \; (1 - \const{}) \, \distrnorm{\paraintstar}{2} \distrnorm{\paraslpstar}{2} \distrnorm{\vecu}{2} \distrnorm{\vecv}{2} \, .
\end{align*}
This yields
\begin{align*}
	\Gamma
	& \; \geq \; 2 \, \distrnorm{\paraslpstar}{2}^2 \, \distrnorm{\vecu}{2}^2 - 2 \, (1 - \const{}) \,  \distrnorm{\paraintstar}{2} \distrnorm{\paraslpstar}{2} \distrnorm{\vecu}{2} \distrnorm{\vecv}{2} + \frac{1}{2} \, \distrnorm{\paraintstar}{2}^2 \, \distrnorm{\vecv}{2}^2  \\
	& \; = \; \const{} \, \Big\{ 2 \, \distrnorm{\paraslpstar}{2}^2 \, \distrnorm{\vecu}{2}^2 +  \frac{1}{2} \, \distrnorm{\paraintstar}{2}^2 \, \distrnorm{\vecv}{2}^2 \Big\} + \frac{1 - \const{}}{2} \, \Big\{ 2 \, \distrnorm{\paraslpstar}{2} \distrnorm{\vecu}{2} - \distrnorm{\paraintstar}{2}  \distrnorm{\vecv}{2} \Big\}^2  \\
	& \; \geq \; \const{} \, \Big\{ 2 \, \distrnorm{\paraslpstar}{2}^2 \, \distrnorm{\vecu}{2}^2 +  \frac{1}{2} \, \distrnorm{\paraintstar}{2}^2 \, \distrnorm{\vecv}{2}^2 \Big\} \, .
\end{align*}
Since $\begin{pmatrix} \vecu, \, \vecv \end{pmatrix}$ is a unit vector, i.e., $\distrnorm{\vecu}{2}^2 + \distrnorm{\vecv}{2}^2 = 1$, we conclude
\begin{align*}
	\Gamma
	& \; \geq \; \const{} \, \Big\{ 2 \, \distrnorm{\paraslpstar}{2}^2 \, \distrnorm{\vecu}{2}^2 +  \frac{1}{2} \, \distrnorm{\paraintstar}{2}^2 \, \distrnorm{\vecv}{2}^2 \Big\}
	\; \geq \; \const{} \, \min \Big\{ 2 \, \distrnorm{\paraslpstar}{2}^2 , \, \frac{1}{2} \, \distrnorm{\paraintstar}{2}^2 \Big\} \, .
\end{align*}
Therefore, we find that 
\begin{align}
	\label{eq:AMtFnormlb}
	\distrnorm{ \AMt }{F}^2 \; \geq \; \const{} \, .
\end{align}

By combining the bounds from \eqref{eq:SstarAFnorm} and \eqref{eq:AMtFnormlb}, we conclude that
\begin{align*}
	\sup_{\begin{subarray}{l} (\vecu, \vecv) \\ \in \Sphere^{2\Dim-1} \cap \, (\paraintstar, 2\paraslpstar)^{\perp} \end{subarray}}
	\begin{bmatrix}
		\vecu  \\  \vecv
	\end{bmatrix}^{\top}
	\Covparatworealpop
	\begin{bmatrix}
		\vecu  \\  \vecv
	\end{bmatrix}
	\; \geq \; \const{} \, ,
\end{align*}
which implies that all the remaining $(2\Dim - 1)$ eigenvalues within the subspace orthogonal to $\begin{pmatrix} \paraintstar, \, 2\paraslpstar \end{pmatrix}$ are bounded below by a constant. Thus, we conclude that $\rank(\Covparatworealpop) = 2\Dim - 1$, completing the proof of claims 2 and 3 in \Cref{lemma:exploration}.


\subsubsection{Verification of conditions in Lemmas~\ref{lemma:paraone_err}~to~\ref{lemma:thm_proof_stage3} for the proof of Theorem~\ref{thm:ub_main} \yaqidone}
\label{sec:verification}

In this section, we confirm that the parameters chosen according to equations \eqref{eq:thm_main_perturb}, \eqref{eq:thm_main_Termone}, and \eqref{eq:thm_main_Termtwo} satisfy the conditions required for the validity of \Cref{lemma:paraone_err,lemma:thm_proof_stage2,lemma:paratwo_err,lemma:thm_proof_stage3}.
We will verify these conditions step by step. Note that, for simplicity, we may reuse the notation for constants $\Const{1}$ and $\Const{2}$, which may represent different values depending on the specific condition being checked.

\paragraph{Condition $\perturb \leq \Const{2}^{-1}$ in \eqref{eq:cond_paratwo_err} (from \Cref{lemma:paratwo_err})\,:}

To satisfy the condition $\perturb \leq \Const{2}^{-1}$ in equation \eqref{eq:cond_paratwo_err}, we can simply set $\Const{0}' \defn \Const{2}^{-1}$ in condition \eqref{eq:thm_main_perturb}. Our next goal is to demonstrate that, by choosing a sufficiently large constant \(\Const{1} > 0\) in the bound~\eqref{eq:ub_main_cond} (\mbox{$\Term \; \geq \; \Const{1} \cdot \Dim \, \log^2 \Term$}), the condition \eqref{eq:thm_main_perturb} has a valid solution for the perturbation level \(\perturb\).

To prove this, we establish that if
\begin{align}
	\label{eq:T_cond_paratwo_err}
	\Term \; \geq \; 2 \, \ConstCI^2 \, \eigpop_1 \, \Const{2}^6 \cdot \Dim \, \log^2 \Term
\end{align}
where $\eigpop_1$ denotes the largest eigenvalue of $\Covparatworealpop$
\footnote{The eigenvalue $\eigpop_1$ satisfies $\eigpop_1 \leq 2 \, \max\{1, \priceub^2\} \, \norm{\Covfeat}_2$ where $\priceub = \sup_{\feat \in \FeatSp} \pricestar(\feat)$ and $\Covfeat = \mathbb{E}_{\feat \sim \distr} [ \feat \, \feat^{\top} ]$.
Assumptions~\ref{asp:pure-online-second-moment-nondegenerate} and \ref{asp:basic-regularity} ensure that it can be regarded as a constant.}
, then setting $\perturb = \Const{2}^{-1}$ satisfies the critical inequality~\ref{eq:CI}.

Let us examine the left-hand side of the inequality \ref{eq:CI} under the choice $\perturb = \Const{2}^{-1}$:
\begin{align*}
	\mbox{Left-hand side of \ref{eq:CI}} \; = \sqrt{\frac{1}{2\Dim} \sum_{k=1}^{2\Dim} \min \Big\{ \frac{\Const{2}^{-2}}{\eigpop_k}, 1 \Big\} }
	\; \geq \sqrt{\frac{\Const{2}^{-2}}{\eigpop_1}} \, .
\end{align*}
Next, using inequality~\eqref{eq:T_cond_paratwo_err}, we can bound the right-hand side of \ref{eq:CI} as follows
\begin{align*}
	\mbox{Right-hand side of \ref{eq:CI}}
	& \; = \; \ConstCI \, \sqrt{\frac{2\Dim}{\Term}} \, \log \Term \, \cdot \, \frac{1}{\perturb^2}  \\
	& \; \leq \; \ConstCI \, \sqrt{\frac{2\Dim}{2 \, \ConstCI^2 \, \eigpop_1 \, \Const{2}^6 \cdot \Dim \, \log^2 \Term}} \, \log \Term \, \cdot \, \Const{2}^2
	\; =  \sqrt{\frac{\Const{2}^{-2}}{\eigpop_1}} \, .
\end{align*}
This confirms that $\perturb = \Const{2}^{-1}$ satisfies the critical inequality~\ref{eq:CI} when the condition~\eqref{eq:T_cond_paratwo_err} holds.
It further implies that the condition \eqref{eq:thm_main_perturb} is feasible for $\Const{0}' = \Const{2}^{-1}$, provided that $\Const{1}$ in condition~\eqref{eq:ub_main_cond} is chosen to be sufficiently large.

\paragraph{Condition $\Termone \ge \Const{1} \cdot \max\{\Dim, \, \log \Term\}$ in \eqref{eq:cond_paraone_err} (from \Cref{lemma:paraone_err}) \\
and condition $\Termone\ge \Const{1} \cdot \Dim \, \log^2 \Term$ in \eqref{eq:cond_thm_proof_stage2} (from \Cref{lemma:thm_proof_stage2}):}
To simplify our analysis, we first unify the two constants $\Const{1}$ in equations~\eqref{eq:cond_paraone_err} and \eqref{eq:cond_paraone_err} by selecting the larger value. Recall that, based on our choice of $\perturb$ in equation \eqref{eq:thm_main_perturb}, we have the bound $\perturb \leq \Const{0}’$.
In this section, we will show that by selecting a sufficiently large value for the constant $\ConstCI$ in the critical inequality \ref{eq:CI}, specifically such that
$\ConstCI \geq \frac{1}{2} \, {\Const{1}}/{(\Const{0}')^2}$,
the two lower bound conditions related to $\Termone$ are satisfied.

From our parameter choice for $\Termone$ in equation \eqref{eq:thm_main_Termone}, and under the condition $\EffDim \geq \EffDim(\perturb)$, we have
\begin{align*}
	\Termone & \; \geq \; \sqrt{\EffDim \, \Term} \, \log\Term \; \geq \; \sqrt{\EffDim(\perturb) \, \Term} \, \log\Term \, .
\end{align*}
The critical inequality~\ref{eq:CI} ensures that
\begin{align*}
	\Term \; \geq \; \sqrt{\EffDim(\perturb) \, \Term} \, \log\Term
	\; \geq \; \ConstCI \cdot 2 \Dim \log^2 \Term \cdot \frac{1}{\perturb^2} \, .
\end{align*}
Next, by applying the condition $\perturb \leq \Const{0}’$ from equation \eqref{eq:thm_main_perturb}, we can bound the expression
\begin{align*}
	\ConstCI \cdot 2 \Dim \log^2 \Term \cdot \frac{1}{\perturb^2}
	&  \; \geq \; \frac{2 \, \ConstCI}{(\Const{0}')^2} \cdot \Dim \log^2 \Term \, .
\end{align*}
Since we chose $\ConstCI$ such that $\ConstCI \geq \frac{1}{2} \, \Const{1} / (\Const{0}')^2$, we obtain
\begin{align*}
	\frac{2 \, \ConstCI}{(\Const{0}')^2} \cdot \Dim \log^2 \Term 
	& \; \geq \; \Const{1} \cdot \Dim \, \log^2 \Term  \, .
\end{align*}

By putting together the pieces, we have established that
\begin{align*}
\Termone \; \geq \;  \Const{1} \cdot \Dim \, \log^2 \Term  \; \geq \;  \Const{1} \cdot \max\{\Dim, \, \log \Term\} \, .
\end{align*}
Therefore, we have verified that the two lower bound conditions on $\Termone$ specified in equations~\eqref{eq:cond_paraone_err} and \eqref{eq:cond_thm_proof_stage2} are satisfied given that $\ConstCI \geq \frac{1}{2} \, \Const{1} / (\Const{0}')^2$.

\paragraph{Condition $\perturb \geq \Const{2} \cdot \sqrt{\Dim/ \Termone}  \, \log\Term$ in \eqref{eq:cond_thm_proof_stage2} (from \Cref{lemma:thm_proof_stage2}):}
We aim to show that the condition holds when we select $\ConstCI \geq \frac{1}{2} \, \Const{2}^2$ in the critical inequality \ref{eq:CI}.

Using the definition of $\Termone$ from equation \eqref{eq:thm_main_Termone}, we have
	\begin{align*}
		\mbox{Right-hand side (RHS)}
		\; = \; \Const{2} \cdot \sqrt{\Dim/ \Termone}  \, \log\Term
		\; \leq \; \Const{2} \cdot \sqrt{\frac{\Dim\, \log\Term}{{\sqrt{\EffDim \, \Term}}}}
		\; \leq \; \Const{2} \cdot \sqrt{\frac{\Dim\, \log\Term}{{\sqrt{\EffDim(\perturb) \, \Term}}}} \, .
	\end{align*}
By applying the critical inequality \ref{eq:CI}, we find that
	\begin{align*}
		\mbox{RHS} \; \leq \; \Const{2} \cdot \sqrt{\frac{\Dim\, \log\Term}{{\sqrt{\EffDim(\perturb) \, \Term}}}}
		\; \leq \; \frac{\Const{2}}{\sqrt{2 \ConstCI}} \cdot \perturb \, .
	\end{align*}
If we choose $\ConstCI \geq \frac{1}{2} \, \Const{2}^2$, then it is guaranteed that $\mbox{RHS} \leq \perturb$. This confirms that the condition $\perturb \geq \Const{2} \cdot \sqrt{\Dim/ \Termone}  \, \log\Term$ is satisfied when $\ConstCI \geq \frac{1}{2} \, \Const{2}^2$.

\paragraph{Condition $\Termtwo \geq \Const{1} \cdot \perturb^{-4} \max\{\Dim, \log \Term\}$ in \eqref{eq:cond_paratwo_err} (from \Cref{lemma:paratwo_err}):}
	We aim to show that the condition
	$ \Termtwo \geq \Const{1} \cdot \perturb^{-4} \max\{\Dim, \log \Term\} $
	holds when we select $\ConstCI \geq \sqrt{\Const{1} / 2}$.
	
	We start by observing that
	\begin{align*}
		\mbox{RHS}
		\; = \; \Const{1} \cdot \perturb^{-4} \max\{\Dim, \log \Term\}
		\; \leq \; \Const{1} \cdot \perturb^{-4} \Dim \, \log^2 \Term \, .
	\end{align*}
	From the critical inequality \ref{eq:CI}, we have
	\begin{align*}
		\perturb^{-4} \Dim \, \log^2 \Term
		\; \leq \; \frac{1}{2 \ConstCI^2} \cdot \frac{\EffDim(\perturb)}{2 \Dim} \, \Term
		\; \leq \; \frac{1}{2 \ConstCI^2} \cdot \frac{\EffDim}{2 \Dim} \, \Term \, .
	\end{align*}
	If we choose $\ConstCI \geq \sqrt{\Const{1} / 2}$, we obtain
	\begin{align*}
		\mbox{RHS}
		\; \leq \; \frac{\Const{1}}{2 \ConstCI^2} \cdot \frac{\EffDim}{2 \Dim} \, \Term
		\; \leq \; \frac{\EffDim}{2 \Dim} \, \Term
		\; \leq \; \Termtwo \, ,
	\end{align*}
	given our choice of $\Termtwo$ from Equation \eqref{eq:thm_main_Termtwo}.
	Therefore, we have verifyied the condition
	$ \Termtwo \geq \Const{1} \cdot \perturb^{-4} \max\{\Dim, \log \Term\} $
	under the assumption $\ConstCI \geq \sqrt{\Const{1} / 2}$.

\paragraph{Condition $\textstyle
	\Termtwo \ge \Const{1} \cdot \big\{ \sum_k (\eigpop_k + \perturb^2)^{-1} \big\} \log^2 \Term$ in \eqref{eq:cond_thm_proof_stage3} (from \Cref{lemma:thm_proof_stage3}):}
	We aim to show that if $\Term \geq ({ 2\Const{1}^2}/{\ConstCI^2}) \cdot 
	\Dim \, \log^2 \Term$, then the condition holds.
	
	We start by noting that
	\begin{align*}
		\sum_{k = 1}^{2\Dim} \frac{1}{\eigpop_k + \perturb^2} 
		\; \leq \; \frac{1}{\perturb^2} \sum_{k = 1}^{2\Dim} \min \Big\{ \frac{\perturb^2}{\eigpop_k}, 1 \Big\} 
		\; = \; \frac{\EffDim(\perturb)}{\perturb^2} \, .
	\end{align*}
	Using the bound above, the right-hand side (RHS) of the condition becomes
	\begin{align*}
		\mbox{RHS}
		& \; = \; \Const{1} \cdot \bigg\{ \sum_{k = 1}^{2\Dim} \frac{1}{\eigpop_k + \perturb^2} \bigg\} \log^2 \Term
		\; \leq \; \Const{1} \cdot \frac{\EffDim(\perturb) \, \log^2 \Term}{\perturb^2} \, .
	\end{align*}
	The critical inequality~\ref{eq:CI} implies
	\begin{align*}
		\mbox{RHS}
		\; \leq \; \frac{\Const{1}}{\ConstCI } \frac{\EffDim(\perturb) }{2 \Dim}  \sqrt{\EffDim(\perturb) \, \Term} \, \log \Term
		\; \leq \; \frac{\Const{1}}{\ConstCI } \frac{\EffDim(\perturb) }{2 \Dim} \sqrt{2 \Dim \, \Term} \, \log \Term \, .
	\end{align*}
	Under the assumption~$\Term \geq ({ 2\Const{1}^2}/{\ConstCI^2}) \cdot 
	\Dim \, \log^2 \Term$, we can simplify the bound to obtain
	\begin{align*}
		\mbox{RHS}
		\; \leq \; \frac{\EffDim(\perturb) }{2 \Dim} \, \Term \; \leq \; \Termtwo \, .
	\end{align*}
	Therefore, we have verified that $\textstyle
	\Termtwo \ge \Const{1} \cdot \big\{ \sum_k (\eigpop_k + \perturb^2)^{-1} \big\} \log^2 \Term$ and the condition is satisfied.

\subsubsection{Degenerate dimensions at the transition points \yaqidone}
\label{sec:turning_points}

In this part, we establish the following two claims in \Cref{sec:thm_main}:
\begin{itemize}
	\item For $\Term \asymp \Dim \, \log^2 \Term$, we have $\EffDim \asymp \Dim$.
	\item For $\Term \asymp \Dim^4 \, \log^2 \Term$, we find that $\EffDim \asymp 1$.
\end{itemize}
Below we address these two transition points in turn.

\paragraph{The transition point $\Term \asymp \Dim \, \log^2 \Term$:}

Suppose that
\begin{align*}
	\Term \; = \; \Const{} \cdot \Dim \, \log^2 \Term
\end{align*}
for some constant $\Const{} > 0$.

For this specific value of $\Term$, we can express the right-hand side of the critical inequality \ref{eq:CI} as
\begin{align*}
	\mbox{Right-hand side of \ref{eq:CI}}
	& \; = \; \ConstCI \, \sqrt{\frac{2\Dim}{\Const{} \cdot \Dim \, \log^2 \Term}} \, \log \Term \, \cdot \, \frac{1}{\perturb^2}
	\; =  \; \ConstCI \, \sqrt{\frac{2}{\Const{}}} \, \cdot \, \frac{1}{\perturb^2} \, .
\end{align*}
Since the left-hand side of \ref{eq:CI} satisfies $\Singular(\perturb) \leq 1$, any solution $\perturb$ to the critical inequality \ref{eq:CI} must satisfy
\begin{align*}
	\ConstCI \, \sqrt{\frac{2}{\Const{}}} \, \cdot \, \frac{1}{\perturb^2} \; \leq \; 1 \, .
\end{align*}
Rearranging the inequality, we obtain
\begin{align*}
	\perturb  \; \geq \; \sqrt[4]{\frac{2 \ConstCI ^2}{\Const{}}} \, .
\end{align*}

By definition, the effective dimension $\EffDim$ satisfies
\begin{align*}
	\EffDim \; \geq \; \EffDim(\perturb) \; \geq \; \frac{\perturb^2}{ \eigpop_1 } \cdot 2 \Dim \, .
\end{align*}
Substituting our bound for $\perturb$, we have
\begin{align*}
	\EffDim \; \geq \; \frac{2 \ConstCI}{\eigpop_1} \sqrt{\frac{2}{\Const{}}} \; \Dim \, .
\end{align*}
Together with the upper bound $\EffDim \leq 2 \Dim$, we conclude that $\EffDim \asymp \Dim$ when $\Term \asymp \Dim \, \log^2 \Term$.

\paragraph{The transition point $\Term \asymp \Dim^4 \, \log^2 \Term$:}

Suppose that
\begin{align*}
	\Term \; = \; \Const{} \cdot \Dim^4 \, \log^2 \Term \, .
\end{align*}

Notice that the left-hand side of \ref{eq:CI} has a lower bound
\begin{align*}
	\mbox{Left-hand side of \ref{eq:CI}} \; = \; \Singular(\perturb) \; \geq \; \frac{\perturb}{\eigpop_1} \, .
\end{align*}
For the right-hand side of \ref{eq:CI}, we have
\begin{align*}
	\mbox{Right-hand side of \ref{eq:CI}}
	& \; = \; \ConstCI \, \sqrt{\frac{2\Dim}{\Const{} \cdot \Dim^4 \, \log^2 \Term}} \, \log \Term \, \cdot \, \frac{1}{\perturb^2}
	\; =  \; \ConstCI \, \sqrt{\frac{2}{\Const{}}} \, \cdot \, \frac{1}{\perturb^2 \sqrt{\Dim^3}} \, .
\end{align*}
From the above, we see that there exists a solution for $\perturb$ satisfying
\begin{align}
	\label{eq:perturb_ub}
	\perturb \; \leq \; \sqrt[6]{\frac{2}{\Const{}}} \, \cdot \, \frac{\sqrt[3]{\ConstCI \, \eigpop_1}}{\sqrt{\Dim}} \, .
\end{align}

We now revoke \Cref{lemma:exploration}, which is developed based on \Cref{asp:exploration}, and derive
\begin{align}
	\label{eq:EffDim_ub}
	1 \; \leq \; \EffDim(\perturb) \; \leq \; \frac{\perturb^2}{ \eigpop_{2 \Dim - 1}} \cdot (2 \Dim - 1) + 1 \, ,
\end{align}
where $\eigpop_{2 \Dim - 1} > 0$ is a constant. 
Combining the bounds~\eqref{eq:perturb_ub}~and~\eqref{eq:EffDim_ub}, we obtain
\begin{align*}
	1 \leq \EffDim(\perturb) \; \leq \; 2 \sqrt[3]{\frac{2 \, \ConstCI^2 \, \eigpop_1^2}{\Const{}}} + 1 \, .
\end{align*}
Therefore, we conclude that $\EffDim(\perturb) \asymp 1$, meaning that we can always pick a degenerate dimension $\EffDim$ of constant order in the regime of $\Term \; \asymp \; \Dim^4 \, \log^2 \Term$.

	\section{Proof of the minimax lower bound: Theorem~\ref{thm:pure-online-lower-bound}}
\label{sec:proof-lower-bound}

In this section, we present the proof of lower bound results using van Trees inequality as the main tool. 
\subsection{Proof of \eqref{equ:lower-bound-1}}
The intended result is equivalent to stating that for any fixed policy $\pi$ and fixed $\feat_{1:T}$, 
\begin{align*}
\sup_{\btheta\in \mathfrak{M}}\Exp_{\btheta}^{\pi}\Big[\sum_{t=1}^{\Term} \Big(r^*({\feat}_t,\btheta)- r({\feat}_t,\price_t,\btheta)\Big)\Bigm|{\feat_{1:T}}\Big]\gtrsim \sqrt{\Term}.
\end{align*}

Recall that we abbreviate $\mathfrak{M}_{b_1,b_2,
\rho_1,\rho_2}$ as $\mathfrak{M}$.
Due to the quadratic structure of the reward function, and similar to \eqref{eq:regret_ub}, we have that 
\begin{align*}
\sup_{\btheta\in \mathfrak{M}}\Exp_{\btheta}^{\pi}\Big[\sum_{t=1}^{\Term} \Big(r^*({\feat}_t,\btheta)- r({\feat}_t,\price_t,\btheta)\Big)\Bigm|{\feat_{1:T}}\Big]&\gtrsim \sup_{\btheta\in \mathfrak{M}}\Exp_{\btheta}^{\pi}\Big[\sum_{t=1}^{\Term}\left(p_t-\pricestar(\feat_t)\right)^2 \Bigm| \feat_{1:T}\Big].
\end{align*}

For any prior $\lambda$ supported on $\mathfrak{M}$, we have \begin{align*}\sup_{\btheta\in \mathfrak{M}}\Exp_{\btheta}^{\pi}\Big[\sum_{t=1}^{\Term}\left(p_t-\pricestar(\feat_t)\right)^2 \Bigm| \feat_{1:T}\Big]\ge\Exp_\lambda\Exp_{\btheta}^{\pi}\Big[\sum_{t=1}^{\Term}\left(p_t-\pricestar(\feat_t)\right)^2 \Bigm| \feat_{1:T}\Big].\end{align*}

It then suffices to prove 
\begin{align}
\label{equ:lower-bound-suffice}
\Exp_\lambda\Exp_{\btheta}^{\pi}\Big[\sum_{t=1}^{\Term}\left(p_t-\pricestar(\feat_t)\right)^2 \Bigm| \feat_{1:T}\Big]\gtrsim \sqrt{T}.  
\end{align} for some chosen prior $\lambda$.

\paragraph{Proof of \eqref{equ:lower-bound-suffice}}
We specify the choices of parameters in order to apply van Trees inequality. Fix a $t\in [T]$.
\begin{enumerate}
\item \emph{ Choice of prior $\lambda$:} Take \begin{align}
\label{equ:def-delta-lower-bound}\delta=\min\{\frac{b_1}{2dC_f},c_\delta\},\end{align} where $c_\delta$ is a constant that will be defined later. 
Pick the prior to be supported at a cube centered at $\overline{\para}$, i.e., $\operatorname{supp}(\lambda)=\prod_{i=1}^{2d}[\overline{\theta_i}-\delta,\overline{\theta_i}+\delta]$. 

More specifically, define 
\[\lambda(\btheta)=\prod_{i=1}^{2d} f_i(\theta_i),\] where $f_i(\theta_i)=\frac{1}{\delta}\cos^2(\frac{\pi(\theta_i-\overline{\theta_i})}{2\delta})$. We state the following Claim \ref{claim:lower-bound-support-inclusion} showing this is a valid prior.

\begin{claim}
\label{claim:lower-bound-support-inclusion}
$\operatorname{Supp}(\lambda)\subseteq \mathfrak{M}$.
\end{claim}
\item \emph{Choice of targets:}
$\psi_n=p_t$, $\psi(\btheta)=\pricestar(\feat_t)$.
\item \emph{Choice of $C$:} $C(\btheta)=\begin{bmatrix}
    \balpha\\2\bbeta
\end{bmatrix}$.

\end{enumerate}
Note that $p_t$ can be viewed as a measurable function of $D_{1:t-1}$. Hence we can apply Lemma \ref{lemma:van-trees} and arrive that
\begin{align}
\label{equ:van-trees-application}\Exp_{\lambda}\Exp_{\btheta}^{\pi}\left[(\price_t-\pricestar({\feat}_t))^2\Bigm|\feat_{1:T}\right]\ge \frac{(\Exp_{\lambda}[C(\btheta)^{\top} \frac{\partial \pricestar({\feat}_t,\btheta)}{\partial \theta}])^2}{\widetilde\calI(\lambda)+\Exp_{\lambda}[C(\btheta)^{\top}\calI_{t-1}({\feat}_{1:t-1},\btheta)C(\btheta)]}.\end{align}
In \eqref{equ:van-trees-application}, we use $\calI_{t-1}({\feat}_{1:t-1},\btheta)$ in place of the information matrix $\calI(\btheta)$ defined in \eqref{equ:van-trees-information definition}, to indicate the dependence on $t$. And $\widetilde\calI(\lambda)$ is defined in \eqref{equ:van-trees-prior-information definition}. Concretely, let the joint density function of the demands before time $t$ be $f(D_{1:t-1},\btheta)$, we define 
\begin{align}
\calI_{t-1}({\feat}_{1:t-1},\btheta)=\Exp_{\btheta}\left[\frac{\partial f(D_{1:t-1},\btheta)}{\partial \btheta}\left(\frac{\partial f(D_{1:t-1},\btheta)}{\partial \btheta}\right)^{\top}\right]
\end{align}
and
\begin{align}
\widetilde \calI(\lambda)=\int_{\mathfrak{M}}\left(\sum_{k=1}^{2d} \frac{\partial}{\partial \theta_k}\left(C_k(\btheta)\lambda(\btheta)\right)\right)^2\frac{1}{\lambda(\btheta)}d\btheta.\end{align}

We make the following claims.
\begin{claim}
\label{claim:information}
$\widetilde\calI(\lambda)=\calO(\Dim^2+\frac{\Dim}{\delta^2})$.
\end{claim}

\begin{claim}
\label{claim-3}
$(\Exp_{\lambda}[C(\btheta)^{\top} \frac{\partial \pricestar({\feat}_t,\btheta)}{\partial \btheta}])^2=\Omega(1)$.
\end{claim}

\begin{claim}
\label{claim-4}
$\Exp_{\lambda}[C(\btheta)^{\top}\calI_t({\feat}_{1:t},\btheta)C(\btheta)]\lesssim\sum_{s=1}^t \Exp_{\lambda}\Exp_{\btheta}^{\pi}\left[(p_s-\pricestar({\feat}_s))^2\Bigm|\feat_{1:T}\right]$.
\end{claim}
We defer the proofs of all these claims until the end and first complete the proof of \eqref{equ:lower-bound-1}.
Based on these claims, \eqref{equ:van-trees-application} then simplifies to 
\begin{align*}
\Exp_{\lambda}\Exp_{\btheta}^{\pi}\left[(\price_t-\pricestar({\feat}_t))^2\Bigm|\feat_{1:T}\right]&\gtrsim \frac{1}{\Dim^2+\frac{\Dim}{\delta^2}+\sum_{s=1}^{t-1} \Exp_{\lambda}\Exp_{\btheta}^{\pi}\left[(\price_s-\pricestar({\feat}_s))^2\Bigm|\feat_{1:T}\right]}\\&\gtrsim \frac{1}{\Dim^2+\frac{\Dim}{\delta^2}+\sum_{s=1}^{t} \Exp_{\lambda}\Exp_{\btheta}^{\pi}\left[(\price_s-\pricestar({\feat}_s))^2\Bigm|\feat_{1:T}\right]}.
\end{align*}
Furthermore, with the choice of $\delta$ in \eqref{equ:def-delta-lower-bound} and $t\le T$, we have
\begin{align*}
\Exp_{\lambda}\Exp_{\btheta}^{\pi}\left[(\price_t-\pricestar({\feat}_t))^2\Bigm|\feat_{1:T}\right]\gtrsim \frac{1}{\Dim^3+\sum_{s=1}^T \Exp_{\lambda}\Exp_{\btheta}^{\pi}\left[(\price_s-\pricestar({\feat}_s))^2\Bigm|\feat_{1:T}\right]}.
\end{align*}
Summing the last expression over $t=1,2,\cdots, T$, we obtain
\begin{align}
\label{equ:upper-bound-last-step}
\sum_{t=1}^T\Exp_{\lambda}\Exp_{\btheta}^{\pi}\left[(\price_t-\pricestar({\feat}_t))^2\Bigm|\feat_{1:T}\right]\gtrsim \frac{T}{\Dim^3+\sum_{s=1}^T \Exp_{\lambda}\Exp_{\btheta}^{\pi}\left[(\price_s-\pricestar({\feat}_s))^2\Bigm|\feat_{1:T}\right]}.
\end{align}
This completes the proof of \eqref{equ:lower-bound-suffice}, where we used $T\ge d^6$. The proof of \eqref{equ:lower-bound-1} follows from this.\qed

\paragraph{Proof of Claim \ref{claim:information}}
To start with, we bound $\|\overline{\para}\|_2^2$. Specifically, we have 
\[b_2^2\ge \Exp_\distr[(\bx^{\top}\overline{\bbeta})^2]=\overline{\bbeta}\!\,^{\top} \bSigma \, \overline{\bbeta}\ge c_{\bSigma}\|\overline{\bbeta}\|_2^2,\]
which implies $\|\overline{\bbeta}\|_2^2\le \frac{b_2^2}{c_{\bSigma}}$. Similarly, it follows that $\|\overline{\balpha}\|_2^2\le \frac{b_2^2}{c_{\bSigma}}$, Combining the two yields \[\|\overline{\para}\|_2^2\le \frac{2b_2^2}{c_{\bSigma}}.\]
We can then directly bound $\widetilde\calI(\lambda)$ as follows,
\begin{align*}
\widetilde I(\lambda)&=\int_{\operatorname{supp}(\lambda)}\left(\sum_{i=1}^\Dim \frac{\frac{\Dim}{\Dim\alpha_i}\Big[\alpha_i \frac{1}{\delta}\cos^2(\frac{\pi(\alpha_i-\overline{\alpha_i})}{2\delta})\Big]}{\frac{1}{\delta}\cos^2(\frac{\pi(\alpha_i-\overline{\alpha_i})}{2\delta})}+\sum_{i=1}^\Dim \frac{\frac{\Dim}{\Dim\beta_i}\Big[2\beta_i \frac{1}{\delta}\cos^2(\frac{\pi(\beta_i-\overline{\beta_i})}{2\delta})\Big]}{\frac{1}{\delta}\cos^2(\frac{\pi(\beta_i-\overline{\beta_i})}{2\delta})}\right)^2 \lambda(\btheta)\Dim\btheta\\
&=\int_{\operatorname{supp}(\lambda)}\left(3d-\sum_{i=1}^\Dim \frac{\alpha_i}{2\delta}\sin(\frac{\pi(\alpha_i-\overline{\alpha_i})}{2\delta})-\sum_{i=1}^\Dim\frac{\beta_i}{\delta}\sin(\frac{\pi(\beta_i-\overline{\beta_i})}{2\delta})\right)^2\lambda(\btheta)\Dim\btheta\\
&\le 2\left(9d^2+\frac{\Dim}{\delta^2}\sum_{i=1}^\Dim [(|\overline{\alpha_i}|+\delta)^2+(|\overline{\beta_i}|+\delta)^2]\right)\\
&\le 2\left(9d^2+\frac{2d}{\delta^2}\sum_{i=1}^\Dim [(\overline{\alpha_i})^2+(\overline{\beta_i})^2]+4d^2\right)\\&\le C(\Dim^2+\frac{\Dim}{\delta^2}),
\end{align*}
where the first inequality uses Cauchy-Schwartz Inequality, and in the last inequality, it suffices to choose the constant $C=\max\{26,2\|\overline{\para}\|_2^2\}\le \max\{26,\frac{4b_2^2}{c_{\bSigma}}\}$. 
\qed

\paragraph{Proof of Claim \ref{claim:lower-bound-support-inclusion}}
From $\overline{\para}\in\mathfrak{M}_{2b_1,b_2/2,
2\rho_1,\widetilde \rho_2}$, we have
$-{\feat}_t^{\top} \overline{\bbeta},{\feat}_t^{\top}\overline{\balpha}\in [2b_1,b_2/2]$. 
Recall that $\|\btheta-\overline{\para}\|_{\infty}\le \delta$.

\begin{enumerate}
\item[(1)] For $\delta\le\frac{b_1}{2dC_f}$, we have for any $\btheta\in \operatorname{supp}(\lambda)$,  
\[-{\feat}_t^{\top} \bbeta=-{\feat}_t^{\top} \overline{\bbeta}+{\feat}_t^{\top} (\overline{\bbeta}-\bbeta)\ge 2b_1-\sqrt{\Dim}\delta\sqrt{2d}\ge b_1\] where we used $\|{\feat}\|\le C_f\sqrt{\Dim}.$
Similarly, we can show that ${\feat}_t^{\top}\balpha\ge b_1$.

\item[(2)] On the other hand, 
\[-{\feat}_t^{\top} \bbeta=-{\feat}_t^{\top} \overline{\bbeta}+{\feat}_t^{\top} (\overline{\bbeta}-\bbeta)\le b_2/2+\sqrt{\Dim}\delta\sqrt{2d}\le b_2.\] 
Similarly, we can show that ${\feat}_t^{\top}\balpha\le b_2$.

\item[(3)] Moreover, we have 
$\|\balpha\|^2\ge \frac{2}{3}\|\overline{\balpha}\|^2-2d\delta^2\ge \frac{16}{3}\rho_1-2d\delta^2$, $\|\bbeta\|^2\ge \frac{2}{3}\|\overline{\bbeta}\|^2-2d\delta^2\ge \frac{4}{3}\rho_1-2d\delta^2$. For $\delta\le \sqrt{\frac{\rho_1}{3d}}$, we have for any $\btheta\in \mathfrak{M}$, \[\|\balpha\|^2\ge 4\rho_1 ,\quad\|\bbeta\|^2\ge \rho_1.\] 

\item[(4)] Last but not least, it holds that $\|\balpha\|\ge \|\overline{\balpha}\|-\delta\sqrt{d}$ by Minkowski inequality, and 
\begin{align*}|\balpha^{\top}\bbeta|&=|\overline{\balpha}^{\top}\overline{\bbeta}+(\balpha-\overline{\balpha})^{\top}\overline{\bbeta}+(\bbeta-\overline{\bbeta})^{\top}\overline{\balpha}+(\bbeta-\overline{\bbeta})^{\top}(\balpha-\overline{\balpha})|\\
&\le |\overline{\balpha}^{\top}\overline{\bbeta}|+\delta\|\overline{\bbeta}\|_1+\delta\|\overline{\balpha}\|_1+d\|\bbeta-\overline{\bbeta}\|_{\infty}\|\balpha-\overline{\balpha}\|_{\infty}\\
&\le |\overline{\balpha}^{\top}\overline{\bbeta}|+\delta\sqrt{d}\|\overline{\bbeta}\|+\delta\sqrt{d}\|\overline{\balpha}\|+d\delta^2.
\end{align*}
It follows that
\begin{align*}\frac{|\balpha^{\top}\bbeta|}{\|\balpha\|\|\bbeta\|}\le \frac{|\overline{\balpha}^{\top}\overline{\bbeta}|+\delta\sqrt{d}\|\overline{\bbeta}\|+\delta\sqrt{d}\|\overline{\balpha}\|+d\delta^2}{(\|\overline{\balpha}\|-\delta\sqrt{d})(\|\overline{\bbeta}\|-\delta\sqrt{d})}\le 
\frac{\widetilde \rho_2\|\overline{\balpha}\|\|\overline{\bbeta}\|+\delta\sqrt{d}\|\overline{\balpha}\|+\delta\sqrt{d}\|\overline{\bbeta}\|+d\delta^2}{(\|\overline{\balpha}\|-\delta\sqrt{d})(\|\overline{\bbeta}\|-\delta\sqrt{d})}.
\end{align*}
Note that $\rho_1,\rho_2$ are constants and $\rho_2<1$. It's straightforward to show for the constant $c_\delta=\min\{\rho_1/2,\rho_2(1-\sqrt{\frac{2\widetilde \rho_2}{\widetilde \rho_2+\rho_2}}),\frac{\rho_1(\rho_2-\widetilde \rho_2)}{16},\frac{\rho_1\sqrt{\rho_2-\widetilde \rho_2}}{4}\}$, when $\delta<\frac{c_\delta}{\sqrt{d}}$, we have that \begin{align*}\frac{|\balpha^{\top}\bbeta|}{\|\balpha\|\|\bbeta\|}\le \frac{\widetilde \rho_2}{(1-\frac{c_\delta}{\rho_2})^2}+\frac{4c_\delta}{\rho_1}+\frac{4c_\delta^2}{\rho_1^2}
\le \rho_2.\end{align*}
\end{enumerate}
Therefore, combining (1)--(4), we have $\operatorname{supp}(\lambda)\subseteq \mathfrak{M}$.

\paragraph{Proof of Claim \ref{claim-3}}
The proof directly follows from 
\[(\Exp_{\lambda}[C(\btheta)^{\top} \frac{\partial \pricestar({\feat}_t,\btheta)}{\partial \btheta}])^2=\Exp_{\lambda}(\frac{{\feat}_t^{\top}\balpha}{2{\feat}_t^{\top}\bbeta})^2\ge \frac{b_1^2}{4b_2^2},\]
where the inequality is as a result of Claim \ref{claim:lower-bound-support-inclusion} and the definition of $\mathfrak{M}$.

\paragraph{Proof of Claim \ref{claim-4}}
Since $\noise_i$ are i.i.d. $\calN(0,\sigma^2)$, it is straightforward to show  $\calI_t({\feat}_{1:t},\btheta)=\frac{1}{\sigma^4}\sum_{s=1}^t\Exp_{\btheta}^{\pi}\begin{bmatrix}
{\feat}_s\\{\feat}_s p_s
\end{bmatrix}\begin{bmatrix}
{\feat}_s\\{\feat}_s p_s
\end{bmatrix}^{\top}$. The proof then directly follows from
\begin{align*}\Exp_{\lambda}[C(\btheta)^{\top}\calI_t({\feat}_{1:t},\btheta)C(\btheta)]&=\frac{1}{\sigma^4}\sum_{s=1}^t \Exp_{\lambda}\Exp_{\btheta}^{\pi}({\feat}_s^{\top}\balpha+2p_s{\feat}_s^{\top}\bbeta )^2\\
&=\frac{1}{\sigma^4}\sum_{s=1}^t \Exp_{\lambda}\Exp_{\btheta}^{\pi}(-2\pricestar({\feat}_s){\feat}_s^{\top}\bbeta+2p_s{\feat}_s^{\top}\bbeta )^2\\
&\le\frac{4b_2^2}{\sigma^4}\sum_{s=1}^t \Exp_{\lambda}\Exp_{\btheta}^{\pi}(p_s-\pricestar({\feat}_s))^2.
\end{align*}

\subsection{Proof of \eqref{equ:lower-bound-2}}
Take expectation over ${\feat}_{1:t}$ on both sides of \eqref{equ:upper-bound-last-step}, and applying Jensen's Inequality, we obtain
\begin{align*}
\sum_{t=1}^{\Term}\Exp_{\lambda}\Exp_{\btheta}^{\pi}(\price_t-\pricestar({\feat}_t))^2\gtrsim\frac{\Term}{\Dim^3+\sum_{t=1}^{\Term} \Exp_{\lambda}\Exp_{\btheta}^{\pi}(\price_t-\pricestar({\feat}_t))^2}.
\end{align*}
Hence, when $T>d^6$, it follows that $\inf_\pi\sup_{\btheta\in \mathfrak{M}}R^{\pi}_{\btheta}(\Term)\gtrsim \inf_\pi\sum_{t=1}^{\Term}\Exp_{\lambda}\Exp_{\btheta}^{\pi}(\price_t-\pricestar({\feat}_t))^2\gtrsim \sqrt{\Term}$.\qed

	\section{Technical Lemmas}

In this section, we present several technical lemmas utilized in the main proof.

\subsection{Analysis of linear regression with sub-Gaussian noise \yaqidone}
\label{sec:ls}

We consider a least-squares regression problem with a fixed design. The data takes the form $\{ (\bx_i, y_i) \}_{i=1}^\numobs$, where $\bx_i \in \Real^{\Dim}$ are the covariates and $y_i \in \Real$ are the responses. The responses are generated according to the model
\begin{align*}
	y_i = \bx_i^{\top} \bbeta^{\star} + \noise_i,
\end{align*}
where $\bbeta^{\star} \in \Real^{\Dim}$ is the true parameter vector, and the noise terms ${\noise_j}{j=1}^\numobs$ are i.i.d. sub-Gaussian random variables with zero mean and variance proxy $\sigma^2$.

We define the design matrix and response vector as follows:
\begin{align*}
	\bX \defn
	\begin{bmatrix}
		\bx_1 & \bx_2 & \cdots & \bx_\numobs
	\end{bmatrix}^{\top}
	\in \Real^{\numobs \times \Dim}
	\qquad 
	\by \defn 
	\begin{bmatrix}
		y_1 & y_2 & \cdots & y_\numobs
	\end{bmatrix}^{\top}
	\in \Real^{\Dim}
	\, .
\end{align*}
The empirical covariance matrix of the design is given by
\begin{align*}
	\bSigmahat \; \defn \; \frac{1}{\numobs} \, \bX^{\top}  \bX
	= \; \frac{1}{\numobs} \sum_{i=1}^\numobs \, \bx_i \, \bx_i^{\top}
	\in \Real^{\Dim \times \Dim} \, .
\end{align*}
The least-squares estimate $\bbetahat \in \Real^{\Dim}$ of the parameter $\bbeta^{\star}$ is given by
$\bbetahat \defn \bSigmahat^{-1} \big( \numobs^{-1} \bX^{\top} \by \big)$.

The following lemma provides a high-probability bound on the estimation error $\distrnorm{\bbetahat - \bbeta^{\star}}{2}$.

\begin{lemma}[Linear regression with sub-Gaussian noise]
	\label{lemma:ls}
	Consider the linear regression problem described above. For any $z \geq 1$, there exists a universal constant $\Const{} > 0$ such that the estimate $\bbetahat$ satisfies
	\begin{align}
		\label{eq:ls}
		\Prob\Big( \, \distrnorm[\big]{\bbetahat - \bbeta^{\star}}{2}^2
		\; \leq \; \Const{} \cdot (\sigma^2/\numobs) \, \trace(\bSigmahat^{-1}) \, z \, \Big)
		\; \geq \; 1 - \exp(-z) \, .
	\end{align}
\end{lemma}

Below we provide the proof of \Cref{lemma:ls}.

\begin{proof}[Proof of \Cref{lemma:ls}]
	
	The proof of \Cref{lemma:ls} is straightforward.
	We define a vector
	{$\bnoise \defn (\noise_1, \noise_2, \ldots, \noise_\numobs)^{\top} \! \in \Real^{\numobs}$}.
	It is evident that
	\begin{align*}
		\bbetahat - \bbeta^{\star}
		= \bSigmahat^{-1} \big( \numobs^{-1} \bX^{\top} \bnoise \big) \, .
	\end{align*}
	Therefore, its squared $L^2$-norm is given by
	\begin{align*}
		\distrnorm[\big]{\bbetahat - \bbeta^{\star}}{2}^2
		\; = \; \bnoise^{\top} \big( \numobs^{-2} \bX \bSigmahat^{-2} \bX^{\top} \big) \, \bnoise \, .
	\end{align*}

We apply the following result, \Cref{lemma:variant-hanson-wright}, which follows directly from Lemma~E.1 in \cite{wang2023pseudo}.

\begin{lemma}
\label{lemma:variant-hanson-wright}
Suppose $\bepsilon\in \RR^{d}$ is a zero-mean random vector with $\|\bepsilon\|_{\psi_2}\le \sigma$ where $\sigma > 0$ is a constant. There exists a universal $C>0$ such that for any positive semi-definite matrix $\bSigma\in\Real^{d\times d}$ and $z\ge 1$, we have 
\[\Prob\left(\bepsilon^{\top} \bSigma \, \bepsilon\le C \cdot \sigma^2 \, \trace(\bSigma) \, z\right) \; \ge \; 1-\exp(-z) \, .\]
\end{lemma}

It follows directly from \Cref{lemma:variant-hanson-wright} that
\begin{align*}
	\Prob\Big( \, \distrnorm[\big]{\bbetahat - \bbeta^{\star}}{2}^2
	\; \leq \; \Const{} \cdot \sigma^2 \, \trace\big( \numobs^{-2} \bX \bSigmahat^{-2} \bX^{\top} \big) \, z \, \Big)
	\; \geq \; 1 - \exp(-z) \, ,
\end{align*}
which reduces to inequality~\eqref{eq:ls} by noticing that
$\trace\big( \numobs^{-2} \bX \bSigmahat^{-2} \bX^{\top} \big) = \numobs^{-1}\trace(\bSigmahat^{-1})$.

\end{proof}

\subsection{Multivariate van Trees inequality}

The lower bound utilizes the Multivariate van Trees inequality\citep{gill1995applications}, which is stated below for completeness.
\begin{lemma}[Multivariate van Trees]
\label{lemma:van-trees}
Consider estimating a real-valued function $\psi(\btheta)$ with parameter $\btheta\in \RR^s$. Suppose we are given $n$ observations $X_1,\cdots,X_n$ drawn from a joint distribution with joint probability density function $f(\bx,\btheta)$. The prior probability density function of $\btheta$ is denoted by $\lambda(\btheta)$, which is supported on the compact set $\Theta\subseteq \RR^s$. And $C(\btheta)\in \RR^s$. If $f,\lambda,C$ satisfy certain regularity conditions (see Assumptions in Section 4 of \cite{gill1995applications}), and in particular, $\lambda(\btheta)$ is positive in the interior of $\Theta$ and zero on its boundary, then for any estimator $\psi_n$ based on $X_1,\cdots,X_n$, we have 
\[\Exp_{\lambda}\Exp_{\btheta}(\psi_n-\psi(\btheta))^2\ge \frac{(\Exp_{\lambda}[C(\btheta)^{\top} \frac{\partial \psi}{\partial \btheta}])^2}{\widetilde\calI(\lambda)+\Exp_{\lambda}[C(\btheta)^{\top}\calI(\btheta)C(\btheta)]}\]
where 
\begin{align}
\label{equ:van-trees-information definition}
\calI(\btheta)=\Exp_{\btheta}\left[\frac{\partial f(X,\btheta)}{\partial \btheta}\left(\frac{\partial f(X,\btheta)}{\partial \btheta}\right)^{\top}\right]
\end{align}
and
\begin{align}
\label{equ:van-trees-prior-information definition}
\widetilde \calI(\lambda)=\int_{\Theta}\left(\sum_{k=1}^s \frac{\partial}{\partial \theta_k}\left(C_k(\btheta)\lambda(\btheta)\right)\right)^2\frac{1}{\lambda(\btheta)}d\btheta.\end{align}
\end{lemma}

\end{document}